\documentclass[11pt, a4paper, logo, copyright, nonumbering]{mll_style}

\usepackage{amsmath,amsfonts,bm}

\def\eqref#1{equation~\ref{#1}}

\def\1{\bm{1}}

\DeclareMathAlphabet{\mathsfit}{\encodingdefault}{\sfdefault}{m}{sl}
\SetMathAlphabet{\mathsfit}{bold}{\encodingdefault}{\sfdefault}{bx}{n}

\usepackage{xspace}  %
\usepackage{listings}
\usepackage[most]{tcolorbox}
\newcommand{\numqa}{$21,154$\xspace}
\newcommand{\numimg}{$3,268$\xspace}
\newcommand{\numgroup}{$976$\xspace}
\newcommand{\name}{$\textsc{MindCube}$\xspace}
\newcommand{\namesub}{$\textsc{MindCube-Tiny}$\xspace}
\newcommand{\modelname}{Qwen2.5-VL-3B-Instruct\xspace}
\newcommand{\iclr}[1]{#1}
\definecolor{deepgreen}{RGB}{0,100,0}
\newcommand{\up}{\textcolor{deepgreen}{$\uparrow$}}
\newcommand{\down}{\textcolor{red!60}{$\downarrow$}}

\definecolor{mygreen}{RGB}{0,150,0}
\definecolor{boxbackground}{HTML}{F0F7FF}  %
\definecolor{boxborder}{HTML}{D0D9E5}      %
\definecolor{accentblue}{HTML}{4A86E8}     %
\definecolor{lightblue}{HTML}{EEF3FF}  %
\definecolor{bordergray}{HTML}{CCCCCC}  %
\definecolor{headerblue}{HTML}{2C5AA0}  %

\definecolor{lavenderframe}{HTML}{E6E6FA}  %
\definecolor{lighterlav}{HTML}{F5F5FF}  %
\definecolor{codegray}{rgb}{0.5,0.5,0.5}  %
\definecolor{codepurple}{HTML}{483D8B}  %
\definecolor{backcolour}{HTML}{F5F5FF}  %

\definecolor{jsonkey}{RGB}{44, 130, 201}     %
\definecolor{jsonstring}{RGB}{255, 140, 0}   %
\definecolor{jsonnumber}{RGB}{34, 139, 34}   %

\lstdefinelanguage{json}{
    basicstyle=\ttfamily\small,
    numbers=left,
    numberstyle=\tiny\color{gray},
    stepnumber=1,
    numbersep=5pt,
    showstringspaces=false,
    breaklines=true,
    frame=none,
    backgroundcolor=\color{gray!5},
    literate=
     *{:}{{{\color{jsonkey}:}}}{1}
      {,}{{{\color{jsonkey},}}}{1}
      {"}{{{\color{jsonstring}"}}}{1}
      {[}{{{\color{jsonkey}[}}}{1}
      {]}{{{\color{jsonkey}]}}}{1}
      {0}{{{\color{jsonnumber}0}}}{1}
      {1}{{{\color{jsonnumber}1}}}{1}
      {2}{{{\color{jsonnumber}2}}}{1}
      {3}{{{\color{jsonnumber}3}}}{1}
      {4}{{{\color{jsonnumber}4}}}{1}
      {5}{{{\color{jsonnumber}5}}}{1}
      {6}{{{\color{jsonnumber}6}}}{1}
      {7}{{{\color{jsonnumber}7}}}{1}
      {8}{{{\color{jsonnumber}8}}}{1}
      {9}{{{\color{jsonnumber}9}}}{1}
}

\newtcblisting{jsonbox}{
  listing engine=listings,
  colback=gray!3!white,
  colframe=gray!75!black,
  boxrule=0.4mm,
  arc=2mm,
  outer arc=2mm,
  breakable,
  enhanced,
  listing only,
  listing options={language=json}
}

\newtcolorbox{promptbox}[2][]{ %
    enhanced,
    breakable,
    boxsep=5pt,
    left=9pt,
    right=7pt,
    top=5pt,
    bottom=5pt,
    colback=boxbackground,
    colframe=boxborder,
    boxrule=0.5pt,
    arc=4pt,
    frame hidden, %
    borderline west={3pt}{0pt}{accentblue},
    shadow={0.5pt}{0.5pt}{1.5pt}{black!10},
    fontupper=\normalsize,
    title=#2, %
    colbacktitle=accentblue, %
    coltitle=white,         %
    fonttitle={\fontsize{9}{11}\selectfont\bfseries}, %
    attach boxed title to top left={yshift=-2.5mm, xshift=3.2mm},
    boxed title style={
        enhanced,
        left=3pt,
        right=3pt,
        top=1pt,    %
        bottom=1pt, %
        boxsep=2pt,
        arc=3pt,
        boxrule=0pt,
        colback=accentblue,
    },
    #1 %
}

\newtcolorbox{notitlepromptbox}[1][]{
    enhanced,
    breakable,
    boxsep=5pt,          %
    left=9pt,            %
    right=7pt,           %
    top=5pt,             %
    bottom=5pt,          %
    colback=boxbackground,
    colframe=boxborder,
    boxrule=0.5pt,
    arc=4pt,             %
    frame hidden,
    borderline west={3pt}{0pt}{accentblue},  %
    shadow={0.5pt}{0.5pt}{1.5pt}{black!10},  %
    notitle,
    fontupper=\normalsize,    %
    #1
}

\newtcolorbox{onebox}[2][]{
    enhanced, 
    center title,
    left*=0pt, right*=0pt,
    boxsep=2pt, left=5pt, right=5pt,
    skin first=enhanced,
    skin middle=enhanced,
    skin last=enhanced,
    colframe = mybluetitle!90,
  colback  = mybluetitle!10,
    fonttitle=\bfseries\rmfamily\fontfamily{phv}\selectfont,
    title={\footnotesize\strut{#2}  \refstepcounter{subsubsection} \addcontentsline{toc}{subsubsection}{\string\numberline{\thesubsubsection}#2}
    },
    #1
    }

\usepackage{natbib}
\usepackage{makecell}
\usepackage{subcaption}
\usepackage{xspace}
\usepackage{adjustbox}
\usepackage{array}
\usepackage{float}
\usepackage{dblfloatfix}
\usepackage{enumitem}
\usepackage{setspace}
\usepackage{multirow}
\usepackage{longtable}
\usepackage{xltabular}
\usepackage{threeparttable}
\usepackage{siunitx}
\usepackage{pdflscape}

\usepackage{mathtools}
\usepackage{bm}
\usepackage{dsfont}

\usepackage{tikz}
\usepackage{pgfplots}
\pgfplotsset{compat=1.18}
\usepackage{algorithm}
\usepackage{algpseudocode}
\usepackage{listings}

\usepackage{cleveref}

\usepackage{todonotes}
\usepackage{fontawesome5}
\usepackage{CJKutf8}
\usepackage{lipsum}
\usepackage{wrapfig}

\usepackage[toc,page,header]{appendix}
\usepackage{minitoc}

\renewcommand \partname{}

\definecolor{cvprblue}{HTML}{6B71A9}
\definecolor{tableblue}{RGB}{201,226,239}

\addto\extrasenglish{
}

\makeatletter
\def\@BTrule[#1]{%
  \ifx\longtable\undefined
    \let\@BTswitch\@BTnormal
  \else\ifx\hline\LT@hline
    \nobreak
    \let\@BTswitch\@BLTrule
  \else
     \let\@BTswitch\@BTnormal
  \fi\fi
  \global\@thisrulewidth=#1\relax
  \ifnum\@thisruleclass=\tw@\vskip\@aboverulesep\else
  \ifnum\@lastruleclass=\z@\vskip\@aboverulesep\else
  \ifnum\@lastruleclass=\@ne\vskip\doublerulesep\fi\fi\fi
  \@BTswitch}
\makeatother

\reportnumber{001}

\title{\centering Spatial Mental Modeling from Limited Views}

\date{}

\author[*]{
Qineng Wang$^{1*}$, Baiqiao Yin$^{1, 3*}$, Pingyue Zhang$^1$, Jianshu Zhang$^1$, Kangrui Wang$^{1}$, Zihan Wang$^1$, Jieyu Zhang$^4$, Keshigeyan Chandrasegaran$^2$, Han Liu$^1$, Ranjay Krishna$^4$, Saining Xie$^3$, Li Fei-Fei$^{2\dag}$, Jiajun Wu$^{2\dag}$, Manling Li$^{1\dag}$\\
{\small $^*$Equal Contribution in Alphabetical Order; $^{\dag}$Equal Advising}
\\
\small $^1$Northwestern University~~~
$^2$Stanford University~~~
$^3$New York University~~~
$^4$University of Washington\\
\vspace{-10pt}
}

\begin{abstract}
Can Vision-Language Models (VLMs) imagine the full scene from just a few views, like humans do? Humans form \textit{spatial mental models} naturally, internal representations of \textit{unseen space}, to reason about layout, perspective, and motion. 
Our \name benchmark with \numqa questions across \numimg images exposes this critical gap, where existing VLMs exhibit near-random performance. 
Using \name, we systematically evaluate how well VLMs build robust spatial mental models through representing positions (cognitive mapping), orientations (perspective-taking), and dynamics (mental simulation for ``what-if'' movements).
We then explore three approaches to help approximate spatial mental models in VLMs, focusing on incorporating unseen intermediate views, natural language reasoning chains, and cognitive maps.
The significant improvement comes from a synergistic approach, ``map-then-reason'', that jointly trains the model to first generate a cognitive map and then reason upon it. %
By training models to reason over these internal maps, we boosted accuracy from $37.8\%$ to $57.8\%$ ($+20.0\%$). Adding reinforcement learning pushed performance even further to $61.3\%$ ($+23.5\%$).
Our key insight is that such scaffolding of spatial mental models, actively constructing and utilizing internal structured spatial representations with flexible reasoning processes, significantly improves understanding of unobservable space.

\vspace{8pt}

\textbf{Website}: \href{https://mll-lab-nu.github.io/mind-cube}{https://mll-lab-nu.github.io/mind-cube}\\~
\textbf{Code}: \href{https://github.com/mll-lab-nu/MindCube}{https://github.com/mll-lab-nu/MindCube}\\~
\textbf{Dataset}: \href{https://huggingface.co/datasets/MLL-Lab/MindCube}{https://huggingface.co/datasets/MLL-Lab/MindCube}\\~
\textbf{Checkpoints}: \href{https://huggingface.co/MLL-Lab/models}{https://huggingface.co/MLL-Lab/models}

\end{abstract}

\begin{document}
\begin{CJK*}{UTF8}{gbsn}

\doparttoc
\faketableofcontents

\maketitle

\section{Introduction}

\begin{figure}[tbp]
    \centering
    \vspace*{-0.5em}
    \includegraphics[width=\linewidth]{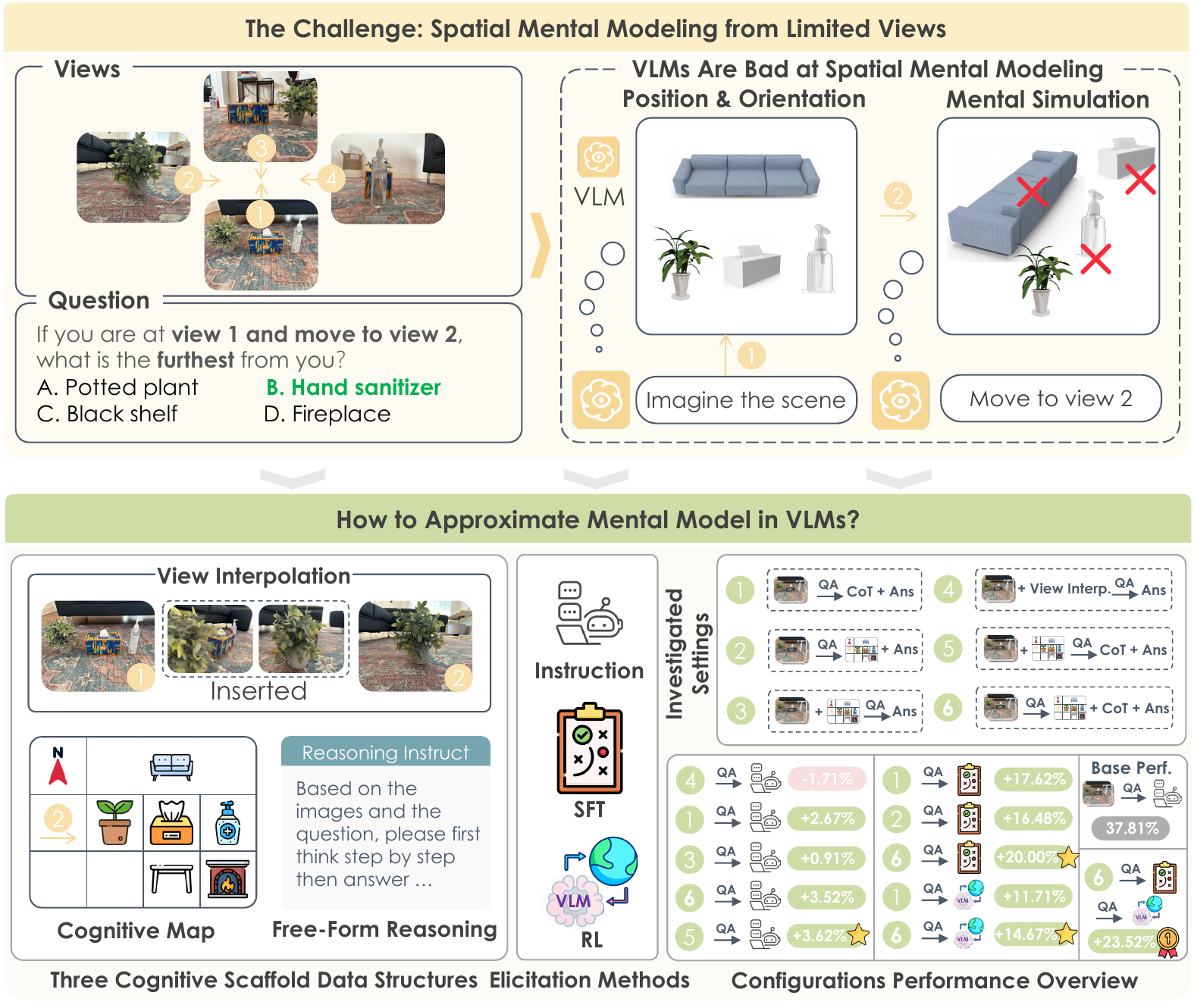}
    \caption{\textbf{Top}: VLMs cannot maintain a coherent mental model when evaluating on the \name benchmark. \textbf{Bottom}: We study how we can help build spatial mental models through external (scaling of views, cognitive map input) and internal strategies (fine-tuning, cognitive map elicitation). We find joint cognitive map and reasoning setting yields the highest gain ($+23.52\%$). }
    \label{fig:teaser_figure}
\end{figure}

For Vision-Language Models (VLMs)~\citep{gpt4o_blog} to move beyond passive perception~\citep{li2023blip2bootstrappinglanguageimagepretraining} to interact with partially observable environments~\citep{yang2024thinkingspacemultimodallarge}, it is fundamental to reason about unseen spatial relationships from limited views.
Consider how effortlessly a human can infer the layout of a room or the hidden objects behind furniture, all by integrating information from several egocentric observations. For example, given the second viewpoint in Figure~\ref{fig:teaser_figure}, human can easily infer the unseen objects behind the ``\textit{plant}'' are the ``\textit{tissue box}'' and the ``\textit{hand sanitizer}'',  including their position, pose, and their relationship with objects that are not simultaneously visible. 
We humans build and update a mental model of our surroundings, even when objects are out of sight. This is enabled by a core cognitive function referred to as \textbf{spatial mental model}~\citep{johnson1980mental, johnson1983mental}: an internal representation of the environment that allows for consistent understanding and inference about space, independent of the current viewpoint. 
VLMs, despite their impressive progress, struggle to synthesize spatial information from limited views, maintain spatial consistency across views, and reason about objects not directly visible~\citep{ma20253dsrbenchcomprehensive3dspatial}.

\begin{figure}[tbp]
    \centering
    \includegraphics[width=\linewidth]{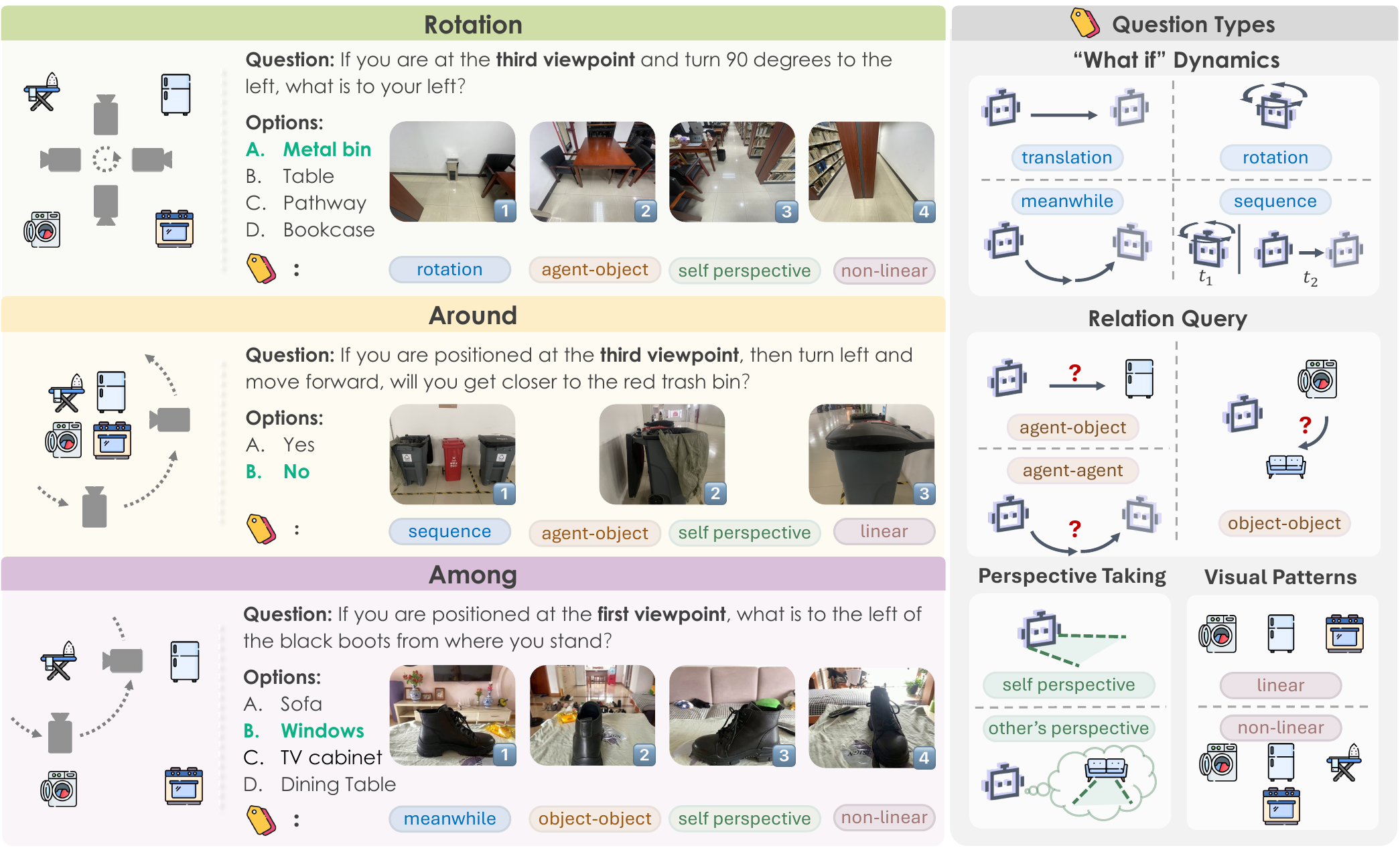}
    \caption{\name taxonomy and examples. Left: Three camera movement patterns (\textsc{Rotation}, \textsc{Around}, \textsc{Among}) with corresponding spatial QA examples. \textbf{Right}: Four-dimensional taxonomy categorizing \name questions types.}
    \label{fig:examples}
\end{figure}

This gap calls for specialized evaluation settings, which must include: (a) reasoning with partial observations where objects are occluded or out of view (such as ``\textit{hand sanitizer}'' in the second viewpoint in Figure~\ref{fig:teaser_figure}), (b) maintaining cross-view consistency across shifting viewpoints (such as through anchor objects ``\textit{plant}''), and (c) mental simulation to infer hidden spatial relationships (such as ``\textit{what if turning left and moving forward}''). 
To fill this gap, we introduce \name, featuring \numqa questions and \numimg images, organized into \numgroup multi-view groups through various types of viewpoint transformations (i.e., \textsc{Rotation}, \textsc{Among}, \textsc{Around} in Figure~\ref{fig:examples}). 
We annotate questions with a focus on objects that are not visible in the current query view. 
As shown in Figure~\ref{fig:examples}, we systematically design question types requiring 
``what-if'' mental simulations from the given view (such as ``\textit{what if turning to left}''), perspective taking (such as ``\textit{what if taking the sofa's perspective}''), complex relation reasoning queries (referencing either the agent or other objects).

Our extensive evaluations of $17$ state-of-the-art VLMs on \name reveal that both open-weight and closed-source models perform only marginally better than random guessing. This poor performance motivates a central question: \textbf{How can we facilitate spatial mental models to reason effectively from partial observations?}

Inspired by spatial cognition~\citep{ramakrishnan2025doesspatialcognitionemerge, lee2025perspectiveawarereasoningvisionlanguagemodels, zha2025enablellm3dcapacity} operating through \textit{visual imagery}, \textit{linguistic reasoning}, or \textit{explicit cognitive maps}, to build consistent spatial awareness across different views, we investigate three approaches to determine whether intermediate representations can assist approximating spatial mental models in VLMs.
\textbf{View Interpolation} enhances the input by providing additional views and thereby offering more information using recorded video, which unexpectedly is not helpful, highlighting the importance of reasoning directly from \textit{limited} views. 
\textbf{Free-form Natural Language Reasoning} verbalizes the mental simulation process, achieving performance gains ($+2.7\%$). 
\textbf{Structured Cognitive Map} simulates global spatial memory from an allocentric (bird's-eye) perspective with orientation and view augmentation. 
Interestingly, providing ground truth cognitive maps directly to answer questions will not yield strong improvements ($-5.81\%$), only actively engaging reasoning with a map achieves strong improvements ($+3.62\%$). 
Despite the effectiveness of reasoning over maps, building accurate spatial mental models exhibit a significant bottleneck attributed to VLMs' intrinsic ability, evidenced by low Isomorphic Rates ($<10\%$) with ground truth maps during generation.
Recognizing this limitation, we train VLMs by constructing $10,000$ reasoning chains and ground truth cognitive maps, investigating how to effectively guide spatial mental models toward achieving accuracy.
While SFT on free-form reasoning chains proved more effective with a gain of $+2.8\%$, guiding models to first build cognitive maps and then perform free-form reasoning over them achieved the best performance, resulting in a total gain of $+5.1\%$, proving scaffolding spatial mental models via actively constructing and utilizing internal structured spatial representations with flexible reasoning processes is highly effective. We also use Reinforcement Learning (RL) to further boost post-SFT performance, guiding models to think in terms of building and reasoning over cognitive maps by injecting structured thinking before RL training, using our SFT model. 
This approach leads to a significant improvement, raising task accuracy from a baseline of $37.8\%$ to $61.3\%$. Our empirical evidence substantiates a critical finding: \textbf{autonomously generating and leveraging internal mental representations help VLMs exhibit superior performance in spatial reasoning tasks, as compared to conventional approaches such as view interpolation or externally-supplied maps.}

\section{\name Benchmark and Evaluation}

\subsection{\name Benchmark}

\begin{table}[tbp]
\centering
\caption{Left: \name data statistics. The number next to the setting (\textsc{Rotation}, \textsc{Among}, \textsc{Around}) means the total QA pairs. Numbers next to each dataset (e.g., Arkitscenes) mean QA pairs/image groups. For example, ``865/53'' for Arkitscenes in \textsc{Rotation} means 865 QA pairs and 53 image groups from it. Right: Performance of VLMs on \name.  \textcolor{black}{\colorbox{blue!7}{Dark blue}} indicates the best result among all models and \textcolor{black}{\colorbox{blue!4}{light blue}} indicates the second best result among all models.} \label{tab:mindcube_stats_eval_results}
\vspace{-0.8em}
\begin{minipage}{0.25\textwidth}
\centering

    \includegraphics[width=\textwidth]{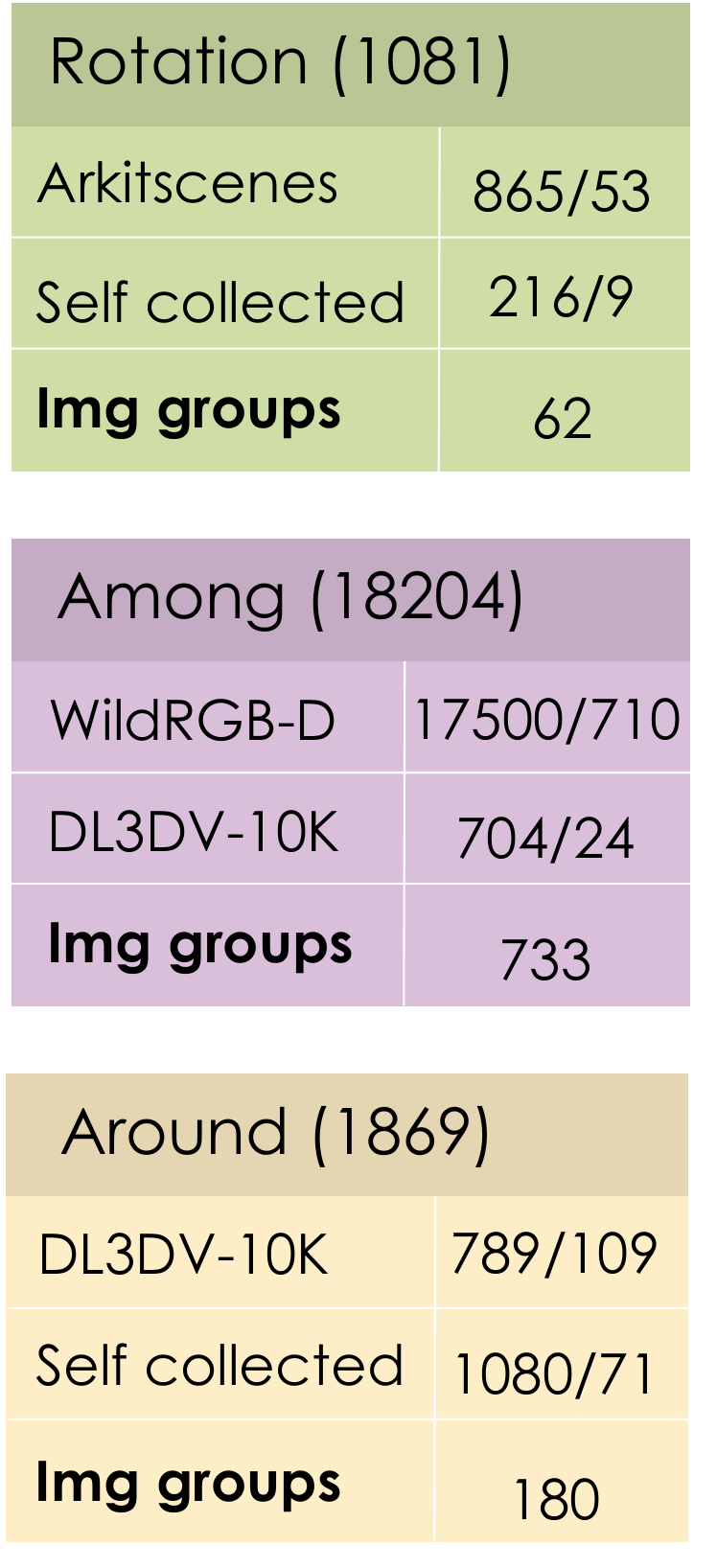} 

\end{minipage}
\hfill
\begin{minipage}{0.73\textwidth}
\centering
\resizebox{\linewidth}{!}{%
  \begin{tabular}{lcccccc}
 \textbf{Method}      & \cellcolor[HTML]{E8E8E8} \textbf{Overall}  & \cellcolor[HTML]{CFDDA6} \textbf{Rotation} & \cellcolor[HTML]{DABFDA} \textbf{Among} & \cellcolor[HTML]{FDEEC7} \textbf{Around} \\
    \hline
\rowcolor{tableblue}
\textit{Baseline} &&&&\\
  \it \textcolor{gray}{Random (chance)} & \it \textcolor{gray}{32.35}  & \it \textcolor{gray}{36.36} & \it \textcolor{gray}{32.29} & \it \textcolor{gray}{30.66} \\
     \it \textcolor{gray}{Random (frequency)} & \it \textcolor{gray}{33.02}  & \it \textcolor{gray}{38.30} & \it \textcolor{gray}{32.66} & \it \textcolor{gray}{35.79} \\
    \hline
\rowcolor{tableblue}
\textit{Open-Weight Multi Image Models} & \rule{0pt}{1em} & \rule{0pt}{1em} & \rule{0pt}{1em} & \rule{0pt}{1em} \\
     LLaVA-Onevision-7B~\cite{onevision}     &  47.43  & 36.45 & \cellcolor{blue!4}48.42 & 44.09 \\
     LLaVA-Video-Qwen-7B~\cite{llava-video}   &  41.96  & 35.71 & 43.55 & 30.12 \\
     LongVA-7B~\cite{longva}     & 29.46  &  35.89 & 29.55 & 24.88 \\
     mPLUG-Owl3-7B-241101~\cite{mplug-owl3}   & 44.85  & 37.84 & 47.11 & 26.91  \\
     InternVL3-8B~\cite{zhu2025internvl3exploringadvancedtraining}        & 37.50 & 26.00 & 42.03 & 36.00 \\
     Qwen2.5-VL-7B-Instruct~\cite{Qwen2.5-VL}         & 29.26  & 38.76 & 29.50 & 21.35  \\
     Qwen2.5-VL-3B-Instruct~\cite{Qwen2.5-VL}         & 33.21  & 37.37 & 33.26 & 30.34  \\
     DeepSeek-VL2-Small~\cite{lu2024deepseekvlrealworldvisionlanguageunderstanding}    &  \cellcolor{blue!7}47.62  & 37.00 & \cellcolor{blue!7}50.38 & 26.91 \\

     Gemma-3-12B-it~\cite{team2025gemma} & 46.67 & 38.39 & 48.38 & 34.63\\
     Mantis-8B (SigLip)~\cite{mantis} &  41.05 & 37.65 & 40.23 & \cellcolor{blue!7}50.99  \\
    \hline

\rowcolor{tableblue}
\textit{Proprietary Models} & \rule{0pt}{1em} & \rule{0pt}{1em} & \rule{0pt}{1em} & \rule{0pt}{1em} \\
     GPT-5-2025-08-07~\cite{OpenAI_GPT5_2025}  &  \cellcolor{blue!4}47.59 & \cellcolor{blue!7}93.33 & 34.17 & 41.63 \\
     Gemini-2.5-pro-2025-06~\cite{geminiteam2025geminifamilyhighlycapable}  &  47.05 & \cellcolor{blue!4}85.50 & 25.95 & 38.40 \\
     Claude-4-Sonnet-20250514~\cite{anthropic_claude4_2025}    &  44.75  & 48.42 & 44.21 & \cellcolor{blue!4}47.62 \\
    \hline
    
\rowcolor{tableblue}
\textit{Spatial Models}&&&& \\
     RoboBrain~\cite{ji2025robobrain} & 37.38 & 35.80 & 38.28 & 29.53 \\
     SpaceMantis~\cite{space-mantis}    & 22.81  & 37.65 & 21.26 & 29.32 \\ 
     Spatial-MLLM~\cite{wu2025spatial} & 32.06 & 38.39 & 20.92 & 32.82 \\
     Space-Qwen~\cite{space-mantis}    & 33.28  & 38.02 & 33.71 & 26.32 \\
  \end{tabular}
 }
\end{minipage}

\vspace*{-1.5em}
\end{table}

\noindent\textbf{Overview.}~
We introduce \name, a benchmark for evaluating VLMs' spatial reasoning under partial observations and dynamic viewpoints.
\name features multi-view orthogonal images paired with spatial reasoning questions, enabling fine-grained analysis of spatial mental modeling performance. It targets key challenges such as maintaining object consistency across views and reasoning about occluded or invisible elements.

\noindent\textbf{Settings.}~
\name incorporates three distinct settings—\textbf{Rotation}, \textbf{Around} and \textbf{Among}  (visualized in left of Figure~\ref{fig:examples}). In the \textbf{Rotation} setting, the challenge lies in interpreting multiple orthogonal views from a static and rotational observation point, requiring models to form a holistic understanding of the environment despite only incremental visibility shifts. The \textbf{Around} setting leverages occlusion to force VLMs to maintain object permanence even with partial visibility and to convert lateral (left-right) relations in frontal views into depth (front-back) cues in side views. The \textbf{Among} setting  maintain spatial consistency and overcome visibility constraints as views are captured around a central object with adjacent ones, each view showing the central object positioned before one surrounding element. VLMs need to share information across views, deducing the overall spatial arrangement and relationships even when not all elements are visible simultaneously. Table~\ref{tab:mindcube_stats_eval_results} (left) summarizes the benchmark's overall data distribution. Details on benchmark design about settings and taxonomies and curation are provided in the Appendix~\ref{app_sec:mindcube}, \ref{app_sec:mindcube_eval}  and ~\ref{app_sec:taxonomy_def}.

\noindent\textbf{Dataset Curation.}~\label{sec:mindcube_curation}
The \name dataset was created through a pipeline: We first selected multi-view image groups matching our taxonomy's movement patterns (Figure~\ref{fig:examples}) and spatial criteria. These were then annotated with key spatial information. Finally, we algorithmically generated taxonomy-aligned questions with targeted distractors. Details are included in the Appendix~\ref{app_sec:mindcube_collection}.

\subsection{Evaluation on \name}
We evaluate VLMs’ spatial mental modeling ablities on \name using a diverse set of models (Table~\ref{tab:mindcube_stats_eval_results}, right; setup details in the Appendix~\ref{app_sec:mindcube_eval}). Results reveal a striking performance gap: the best model, DeepSeek-VL2-Small, achieves only 47.62\% accuracy, well above chance but far from human-level~\ref{app_sec:human_eval}. While some models show strength in specific areas—notably GPT-5 in \textsc{Rotation} (93.33\%) and Mantis-8B (SigLip) in \textsc{Around} (50.99\%)—no single model excels across all categories. We also observe that proprietary models generally outperform the open-source ones. Spatial fine-tuning also yielded varied outcomes without consistently reaching top performance. Overall, neither multi-image input nor spatial fine-tuning reliably improves spatial reasoning, raising a key question:
\textbf{How can we help VLMs develop or approximate these crucial spatial reasoning capabilities?}

\section{Which Scaffolds Best Guide Spatial Mental Modeling?}
To address the identified gap, we first evaluate whether structured data forms can scaffold spatial reasoning in frozen VLMs by approximating spatial mental models under limited views.

\begin{figure}[htbp]
\centering
\includegraphics[width=\linewidth]{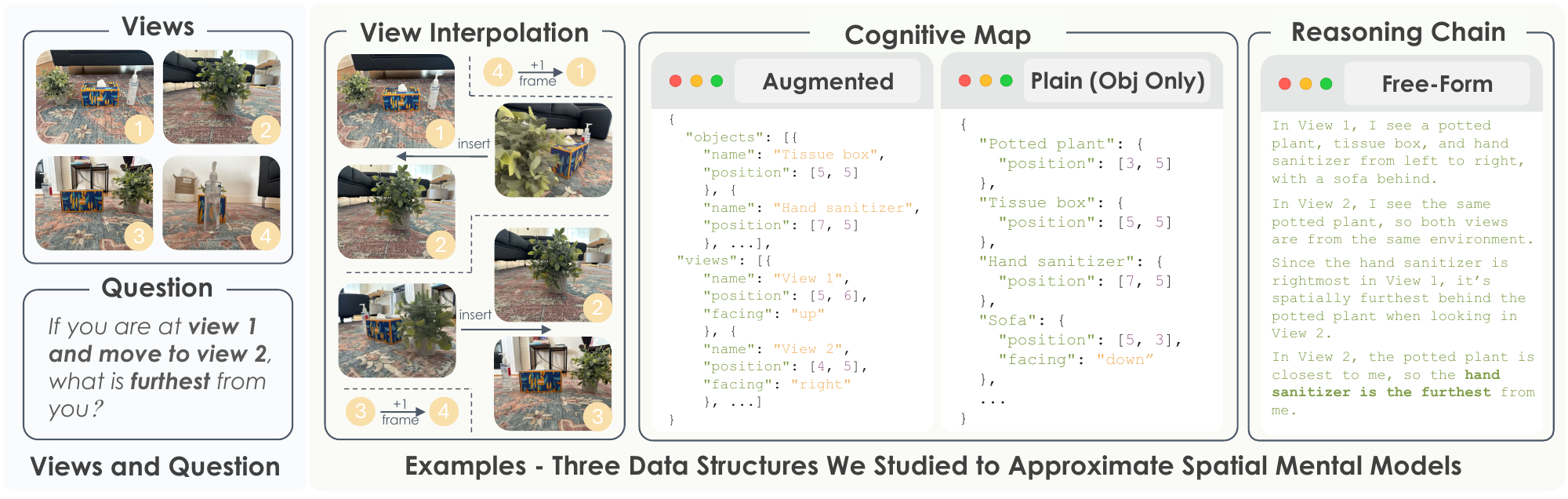}
\caption{Grounded examples of our three data structures that approximate spatial mental models.}
\label{fig:data_examples}
\vspace*{-2em}
\end{figure}

\subsection{Data Structures as Cognitive Scaffolds for Spatial Mental Models} \label{sec:scaffolds}

We investigate whether certain data structures can act as cognitive scaffolds that help form spatial mental models in VLMs from limited visual observations. In cognitive science, spatial mental models are internal representations encoding the relative configuration of objects and viewpoints. Rather than metric-precise maps, they are schematic, manipulable constructs that support reasoning across fragmented observations and unseen perspectives \citep{johnson1983mental, tversky1993cognitive, tversky1994spatial, tversky2003structures}. For instance, humans can mentally simulate turning or infer what lies behind them, suggesting that such representations are flexible, incomplete, yet functionally effective. Drawing on this literature, we define three data structures below (detailed introduction can be found in Appendix~\ref{app_sec:data_structure}), each targeting distinct cognitive properties (integration, transformation, inference) of spatial mental models, with grounded examples in Figure~\ref{fig:data_examples}:
\begin{enumerate}[leftmargin=12pt, topsep=-1pt, itemsep=0pt]
    \item \textbf{View Interpolation}. Interpolating between sparse views introduces perceptual continuity, echoing the process of \textit{mental animation} \citep{hegarty1992mental} and supporting internal transformation such as imagined rotation. This structure scaffolds the dynamic updating capability of spatial mental models. Figure~\ref{fig:data_examples} shows a one-frame inserting example that replaces the original question images.
    \item \textbf{Augmented Cognitive Map}. A cognitive map is a 2D schematic representation of object layouts in space. Such maps resemble Tversky's \textit{cognitive collages}~\citep{tversky1993cognitive}, and they capture locally coherent but fragmented structures. Recent studies \citep{yang2024thinkingspacemultimodallarge, yeh2025seeing} on VLM-based spatial intelligence typically adopt a \textit{plain} form that only encodes object positions in a top-down view. We propose an \textit{augmented} variant that incorporates discrete views, with both objects and views annotated by position and orientation, thereby approaching the relational consistency of \textit{spatial mental models}.

    \item \textbf{Free Form Reasoning}. Open-ended, step-by-step natural language reasoning offers a \textit{procedural approximation} of how spatial models are constructed and queried. While less rigid than map-like structures, such reasoning reflects the inferential function of spatial mental models, especially under ambiguous or incomplete observations \citep{tversky1994spatial}. 
\end{enumerate}

\subsection{Experiment Setup} \label{sec:prompt_exp_setup}

We conduct controlled experiments with fixed input formats to test whether structured scaffolds can help without retraining. Each condition introduces a different structure to support internal modeling.

\begin{table}[htbp]
  \centering
  \footnotesize
  \caption{
    Input--output configurations used in all experiments.
    The suffix ``-In'' means the cognitive map is given to the model as input, whereas ``-Out'' means the cognitive map is predicted as an intermediate output before answering.
    ``Aug'' indicates maps with object and camera annotations; ``Plain'' indicates maps without these augmentations.
    VI = View Interpolation, CGMap = Cognitive Map, FFR = Free-form reasoning.
    Figure~\ref{fig:data_examples} shows visual examples of input structures.
  }
  \label{tab:configs}
  \setlength{\tabcolsep}{4pt}
  \vspace{-0.75em}
  \begin{tabularx}{\linewidth}{@{}l X X@{}}
    \toprule
    \textbf{Name} &
    \textbf{What the model receives (input)} &
    \textbf{What the model produces (output)} \\
    \midrule
    \texttt{Raw-QA} &
      Raw views + question text &
      Direct answer \\
    \texttt{VI-1} &
      Raw views + \textbf{1} interpolated view + question text &
      Direct answer \\
    \texttt{VI-2} &
      Raw views + \textbf{2} interpolated views + question text &
      Direct answer \\
    \texttt{FFR} &
      Raw views + question text &
      Free-form reasoning $\rightarrow$ answer \\
    \texttt{Aug-CGMap-In} &
      Augmented cognitive map (objects + camera) + question text &
      Direct answer \\
    \texttt{Aug-CGMap-Out} &
      Raw views + question text &
      Augmented cognitive map $\rightarrow$ answer \\
    \texttt{Plain-CGMap-Out} &
      Raw views + question text &
      Plain cognitive map $\rightarrow$ answer \\
    \texttt{Aug-CGMap-FFR-Out} &
      Raw views + question text &
      Augmented cognitive map + free-form reasoning $\rightarrow$ answer \\
    \texttt{Plain-CGMap-FFR-Out} &
      Raw views + question text &
      Plain cognitive map + free-form reasoning $\rightarrow$ answer \\
    \texttt{CGMap-In-FFR-Out} &
      Augmented cognitive map (objects + camera) + question text &
      Free-form reasoning $\rightarrow$ answer \\
    \bottomrule
  \end{tabularx}
\end{table}

\noindent\textbf{Configurations and Evaluation Metrics.}~ \label{sec:prompt_task_configuration}
Each experiment is defined by two orthogonal axes:
\emph{Input Structure} (what spatial evidence VLMs receive) and
\emph{Output Format} (the required response type).
As the experimental foundation of this paper, we begin with the ten possible configurations listed in Table~\ref{tab:configs}, from which we investigate a representative subset. Specifically, our grounded cognitive maps are generated using the object arrangements annotation described in Section~\ref{sec:mindcube_curation}, and examples for all configurations are provided in the Appendix~\ref{app_sec:example_configuration}. In the frozen VLMs evaluation setup, we exclude the \texttt{Aug-CGMap-Out} and \texttt{Plain-CGMap-Out} settings, as VLMs tend to conflate map generation with reasoning, even when instructed otherwise. Beyond evaluating task performance using QA accuracy, we also introduce two well-defined graph metrics for generated cognitive maps: (1) \textit{Overall Similarity}, a weighted score combining directional and facing consistency; and (2) \textit{Isomorphic Rate}, measuring whether all pairwise object relations match the ground truth under optimal alignment. Full definitions are provided in the Appendix~\ref{app_sec:evaluation_metrics}.

\noindent\textbf{Model and Evaluation Data}~ We conduct all experiments using \textit{Qwen2.5-VL-3B-Instruct}~\citep{Qwen2.5-VL} with all evaluations performed on \namesub, a diagnostic subset sampled from \name, containing 1,050 questions in total. Detailed statistics are: 600 from \textsc{Among}, 250 from \textsc{Around}, and 200 from \textsc{Rotation}.  \vspace{-0.8em}

\subsection{Do Scaffolds Improve Spatial Mental Modeling Without Training?} \label{sec:prompt_analysis}
We evaluate how well the seven input configurations defined in Table~\ref{tab:configs} support spatial mental modeling in VLMs under limited views, without any model updates. Results are shown in Table~\ref{tab:prompt_results_main} (left).

\noindent\textbf{How far can structure alone go?} We begin with the baseline: raw input views and direct answering (\texttt{Raw-QA}), which achieves 37.81\% accuracy. Adding interpolated views, which we hope to simulate smoother perceptual transitions, leads to no meaningful gain ($\uparrow 0.09\%$). \iclr{We include a further analysis on \texttt{VI} in Appendix~\ref{app:vi_frozen}.} Similarly, providing a pre-computed augmented cognitive map as direct input (\texttt{Aug-CGMap-In}) severely degrades performance to 32.00\%. In contrast, enabling free-form reasoning (\texttt{FFR}) alone or combined with other settings provides a substantial boost to 41.33\%. These results suggest: \textit{structure alone, whether visual or spatial, is not enough}. Without engaging reasoning, VLMs struggle to leverage even well-formed spatial cues to improve spatial mental models.
\setcounter{table}{2}
\begin{table}[htbp]
\centering
\caption{Left: QA accuracy (\%) of \textit{\modelname} on the \namesub benchmark under different configs for frozen VLMs. Right: Graph metrics for two cog map output settings.} \label{tab:prompt_results_main}
\begin{minipage}{0.61\textwidth}
\resizebox{\linewidth}{!}{%
\begin{tabular}{lcccc}
\toprule
\raisebox{-0.5ex}{\textbf{Config.}} & \raisebox{-0.5ex}{\textbf{Overall}}
  & \raisebox{-0.5ex}{\textbf{Rotation}}
  & \raisebox{-0.5ex}{\textbf{Among}}
  & \raisebox{-0.5ex}{\textbf{Around}} \\[1ex]
\midrule
\texttt{Raw-QA}           & 37.81 & 34.00 & 36.00 & 45.20 \\ 
\midrule
\texttt{VI-1}         & 37.90\up & 35.50 & 37.33 & 41.20 \\
\texttt{VI-2}         & 37.81$-$ & 35.50 & 36.50 & 42.80 \\
\midrule
\texttt{Aug-CGMap-In}         & 32.00\down   & 35.00 & 30.50 & {33.20} \\ 
\texttt{FFR}                     & 40.48\up   & 32.00 & 36.00 & 58.00 \\
\texttt{Aug-CGMap-FFR-Out}            & 40.57\up   & 21.00 & \textbf{43.00} & 50.40 \\
\texttt{Plain-CGMap-FFR-Out}           & {41.33}\up & {25.00} & 39.67 & \textbf{58.40} \\
\texttt{CGMap-In-FFR-Out}      & \textbf{41.43}\up & \textbf{37.00} & 41.67 & 44.40 \\
\bottomrule
\end{tabular}
}
\end{minipage}
\hfill
\begin{minipage}{0.36\textwidth}
\centering
\includegraphics[width=\textwidth]{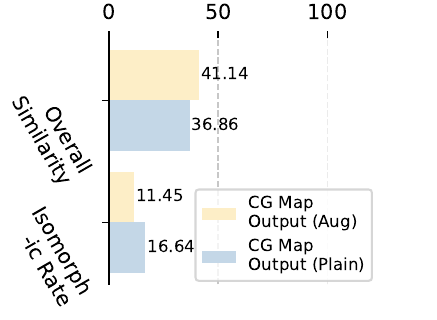} 
\end{minipage}
\end{table}

\noindent\textbf{Can we prompt the model to think spatially?} The answer appears to be yes. Prompting the model to generate a cognitive map (\texttt{Aug-CGMap-FFR-Out}, \texttt{Plain-CGMap-FFR-Out}) before answering leads to further improvements over free-form reasoning alone (\texttt{FFR}) from 40.48\% to 41.43\%. This suggests that generating a map may encourage the model to first form a global understanding of the scene, which in turn supports more structured reasoning.
Both map forms have a great format-following ability, yet fail to generate accurate maps. Overall, augmented maps perform worse. In Table~\ref{tab:prompt_results_main} (Right), despite generating syntactically valid maps for both formats, similarity to grounded maps is low ($<$ 50\%), reflecting limited mapping ability. Notably, both augmented and plain maps have low isomorphism rates (0.10\%, 7.43\%). The reason that the isomorphic rate for augmented map setting is nearly zero is likely because the added view-level details increase generation errors. Detailed case examples can be found in the Appendix~\ref{app_sec:prompt}.

\begin{promptbox}{\faLightbulb\ Key Takeaways: Scaffolding Spatial Mental Models in \textit{Frozen VLMs}}
\it
\begin{itemize}[leftmargin=*, after=\vspace{-0.5em}, itemsep=1pt]
\item Explicit reasoning is crucial for improving performance.
\item Reasoning acts as a necessary mechanism to ground spatial structure in frozen settings.
\item Passive structures (like maps as input) alone and visual continuity offer little benefit.
\end{itemize}
\end{promptbox}

\section{Can We Train for the Emergence of Spatial Mental Models via VLMs’ Use of Scaffolds?}
So far, prompting frozen VLMs with external scaffolds, such as interpolated views or cognitive maps, has yielded limited gains. These techniques fail to tackle the core limitation: VLMs do not form internal spatial representations or reason through space effectively. To go further, we want to know: Can supervised fine-tuning (SFT) and Reinforcement learning (RL) teach VLMs to build and leverage spatial mental models from within?

\subsection{Designing a Robust Experimental Framework} \label{sec:sft_exp_config}
To ensure consistency and comparability, we inherit experimental configurations detailed in Sections \ref{sec:scaffolds} and \ref{sec:prompt_exp_setup}. Specifically, we retain: (1) the two effective data scaffolds—Cognitive Maps (Object-only / Object + Camera) and Free-Form Reasoning, (2) the base model \textit{\modelname}, (3) the evaluation benchmark \namesub, and (4) all established evaluation metrics. View interpolation is excluded due to its limited performance gains in earlier validations.

\noindent\textbf{SFT Task Configurations.}
~Drawing on insights from Section~\ref{sec:prompt_analysis}, we use selected configurations from Table~\ref{tab:configs} to evaluate the incremental impact of cognitive map generation and free-form reasoning in SFT. These include baseline QA without explicit reasoning (\texttt{Raw-QA}), reasoning guided by generated maps only (\texttt{Plain-CGMap-Out}, \texttt{Aug-CGMap-Out}), reasoning-augmented prompts (\texttt{FFR}), and a fully integrated setup that asks VLMs to generate both maps and reasoning (\texttt{Aug-CGMap-FFR-Out} and \texttt{Plain-CGMap-FFR-Out}).

\noindent\textbf{RL Task Configurations and Reward Design.} We employ the VAGEN framework~\citep{wang2025vagen} for VLM policy optimization, using Group Relative Policy Optimization (GRPO) \citep{shao2024deepseekmath} as our core algorithm. We evaluate RL variants along two axes: the output format (\texttt{FFR}-only vs. \texttt{CGMap-FFR}) and the initialization (from scratch vs. from the best SFT checkpoint), yielding six configurations in total (Table~\ref{tab:combined_results}). Detailed settings can be found in the Appendix~\ref{app_sec:rl_exp_setup}.

\noindent\textbf{Grounded Cognitive Maps and Free-Form Reasoning Chain.}
~Grounded cognitive maps are not only used as the input in the \texttt{Aug-CGMap-In} and \texttt{CGMap-In-FFR-Out} setting for the frozen VLMs in the Section~\ref{sec:prompt_task_configuration}, but also as the training and comparison data. We curate such grounded cognitive maps through a template-based method, where we always select the front image in our annotation as the ``up'' direction.
We also manually constructed grounded reasoning chains using detailed image annotations and structured question templates, ensuring logical coherence and clear grounding in observable spatial relations (see an example in Figure~\ref{fig:data_examples}). The detailed grounded cognitive maps and reasoning data generation pipelines are shown in the Appendix~\ref{app_sec:sft_cogmap_gen} and \ref{app_sec:sft_ffr_gen}. We also evaluate the effect of removing viewpoint descriptors from the question text in Appendix~\ref{app:view_prior}, confirming that the map-then-reason advantage holds even without textual directional cues.

\begin{figure}[tbp]
  \vspace*{-0.5em}

  \centering
  \includegraphics[width=\linewidth]{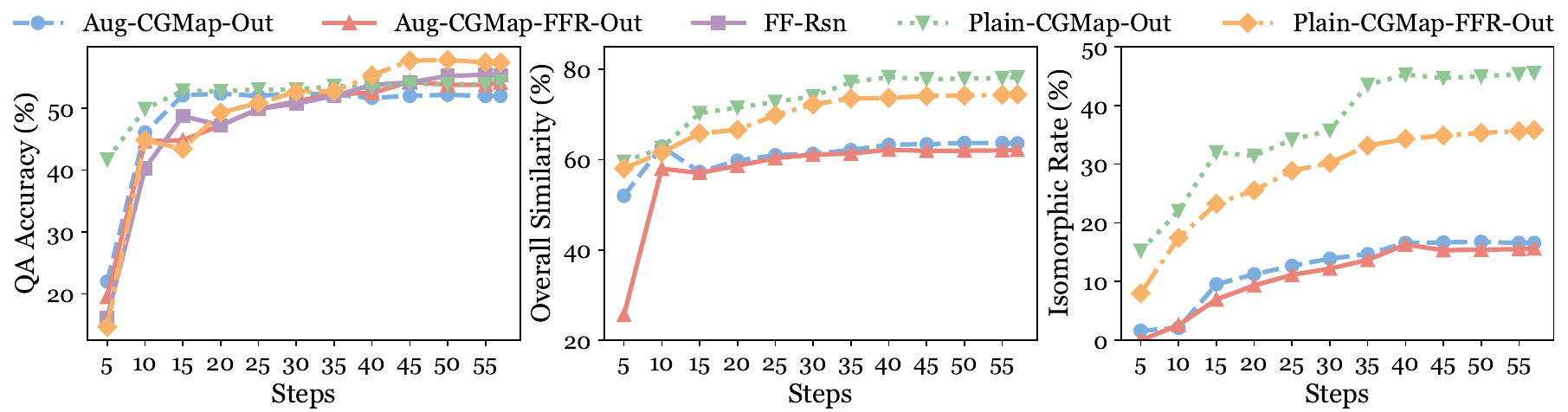}
  \caption{SFT per 5 step training performance on task accuracy and graph metrics.}
  \label{fig:sft-whole}
\end{figure}
\subsection{Do the Emergence of Spatial Mental Models Truly Benefit from Explicit Training?}
We explore several SFT configurations (results shown in Table~\ref{tab:combined_results}), guided by a series of core questions. Fine-tuning directly on raw QA pairs, without spatial supervision, raises accuracy from 37.81\% to 52.67\%. This suggests VLMs can absorb some spatial cues from QA data alone. We use this setup as the baseline for evaluating methods that explicitly incorporate spatial structures. Primary modifications in SFT phase include adjusted training hyperparameters (detailed in the Appendix~\ref{app_sec:sft_exp_setup}) and the input-output configurations.
\begin{table}[htbp]
    \centering
    \caption{QA accuracy (\%) and cognitive map generation quality of \textit{\modelname} under both SFT and RL on \namesub. \texttt{FFR} refers to free-form reasoning. Bolded means the best within that training category (SFT or RL).}
    \scriptsize
    \vspace*{-0.5em}
    \resizebox{\linewidth}{!}{%
    \begin{tabular}{>{\centering\arraybackslash}p{2pt}lcccccc}
    \hline

    & \multirow{2}{*}{\rule{0pt}{2.5ex}\textbf{Config.}} &
    \multicolumn{4}{c}{\textbf{\namesub QA Accuracy (\%)}} &
    \multicolumn{2}{c}{\textbf{Generated Cognitive Map (\%)}} \\
    \cline{3-6}\cline{7-8}
    & & \makecell{\rule{0pt}{2.5ex}\textbf{Overall}} & \makecell{\rule{0pt}{2.5ex}\textbf{Rotation}} & \makecell{\rule{0pt}{2.5ex}\textbf{Among}} & \makecell{\rule{0pt}{2.5ex}\textbf{Around}} & \makecell{\rule{0pt}{2.5ex}\textbf{Overall Sim.}} & \makecell{\rule{0pt}{2.5ex}\textbf{Isom. Rate}} \\
    \hline
    \multirow{6}{*}{\makebox[5pt][c]{\rotatebox[origin=c]{90}{\scriptsize\textbf{SFT}}}} & \texttt{Raw-QA} & 52.67 & 34.50 & 52.50 & 67.60 & -- & -- \\
    & \texttt{FFR} & 55.43\up & 36.00 & 57.17 & 66.80 & -- & -- \\
    & \texttt{Aug-CGMap-Out} & 52.48 & 30.00 & 52.17 & 71.20 & 61.28 & 13.90 \\
    & \texttt{Plain-CGMap-Out} & 54.29\up & 32.00 & 53.67 & \textbf{73.60} & \textbf{78.18} & \textbf{45.52} \\
    & \texttt{Aug-CGMap-FFR-Out} & 54.29\up & \textbf{41.50} & 52.33 & 69.20 & 61.92 & 15.33 \\
    & \texttt{Plain-CGMap-FFR-Out} & \textbf{57.81}\up & 36.50 & \textbf{61.17} & 66.80 & 74.18 & 35.33 \\
    \hline
    \multirow{6}{*}{\makebox[5pt][c]{\rotatebox[origin=c]{90}{\scriptsize\textbf{RL}}}} &
    \texttt{RL-FFR (from scratch)} & 49.52 & 26.50 & 51.50 & 63.20 & -- & -- \\
    & \texttt{RL-Aug-CGMap-FFR-Out (from scratch)} & 52.48 & \textbf{36.00} & 51.50 & 68.00 & 55.71 & 0.00 \\
    & \texttt{RL-Plain-CGMap-FFR-Out (from scratch)} & 50.86 & 34.00 & 50.50 & 65.20 & 29.59 & 6.67 \\
    & \texttt{RL-FFR (from SFT)} & 59.14 & 31.50 & 66.00 & 64.80 & -- & -- \\
    & \texttt{RL-Aug-CGMap-FFR-Out (from SFT)} & 60.86 & \textbf{36.00} & 66.00 & \textbf{68.40} & 62.48 & 16.95 \\
    & \texttt{RL-Plain-CGMap-FFR-Out (from SFT)} & \textbf{61.33} & 29.50 & \textbf{69.17} & 68.00 & \textbf{73.36} & \textbf{35.33} \\
    \hline
    \end{tabular}
    }
    \label{tab:combined_results}
\end{table}

\noindent\textbf{Can structured approximations of mental models alone meaningfully improve performance?} As shown in Table\ref{tab:combined_results}, supervised fine-tuning on explicit cognitive maps, either \textit{Augmented} or \textit{Plain}, leads to substantial improvements in graph structure quality. However, the effect on end-task accuracy remains limited. \texttt{Aug-CGMap-Out} (52.48\%) shows no improvement over \texttt{Raw-QA} (52.67\%), while \texttt{Plain-CGMap-Out} (54.29\%) offers only a modest gain. \texttt{FFR} alone yields a moderate gain (55.43\%), yet still falls short of the joint approach. This means that a scaffold alone is not sufficient to automatically translate into performance gains.

\noindent\textbf{Generating both cognitive maps and free-form reasoning is the most effective approximation.}
Among all configurations, the combination of generating a plain map and then reasoning (\texttt{Plain-CGMap-FFR-Out}) yields performance gain ($\uparrow$5.14\% compared to \texttt{Raw QA-SFT}), surpassing models that rely on only map generation or reasoning alone. This suggests a synergy between structured spatial modeling and natural language inference. The training dynamics reveal a crucial trade-off that explains this synergy. As shown in Figure~\ref{fig:sft-whole} (b, c), models trained solely on map generation (\texttt{Plain-CGMap-Out}) learn the target structure very rapidly, quickly reaching high similarity and isomorphism. However, their QA accuracy soon plateaus (Figure~\ref{fig:sft-whole}a), suggesting the model learns the structure without fully grasping its functional utility. In contrast, the top-performing \texttt{Plain-CGMap-FFR-Out} model learns the map structure more slowly and never reaches the same level of structural perfection. Yet, its QA accuracy continues to increase and surpass all other configurations. This suggests that the joint pressure of the reasoning task forces the model not just to replicate a structure, but to build a functionally effective spatial representation, which can lead to improvement for overall spatial understanding despite being imperfect.

\begin{promptbox}{\faLightbulb\ Key Takeaways: Explicit Training for the Emergence of Spatial Mental Models}
\it
\begin{itemize}[leftmargin=*, after=\vspace{-0.5em}, itemsep=1pt]
\item Joint cogmap and reasoning setting yields optimal performance through synergistic effects.
\item Neither map generation nor reasoning alone matches the performance of the joint approach.
\end{itemize}
\end{promptbox}
\vspace{-0.5em}

\subsection{Can Reinforcement Learning Further Refine Spatial Mental Models?}
While SFT establishes a strong baseline for spatial mental modeling, emerging evidence from models like DeepSeek R1~\citep{guo2025deepseek} suggests reinforcement learning (RL) can offer additional gains by optimizing behavior through outcome-driven feedback. We ask: Can reward-guided refinement help VLMs build sharper spatial models and reason more effectively?

RL lets a model \emph{feel} the consequences of its spatial thoughts through reward, but does that feedback alone forge a genuine “mental map”, or must we first teach the model what a map looks like? Table \ref{tab:combined_results} summarizes key settings and answers this question in two parts.

\noindent\textbf{RL in a vacuum is not enough.}
Training from scratch with sparse rewards provides insufficient guidance for building robust spatial representations. When asked to produce free-form reasoning (\texttt{RL-FFR (from scratch)}), the model achieves only 49.52\% overall accuracy. This result, while an improvement over initial baselines, confirms that task-level rewards alone are too unstructured to effectively teach spatial abstraction.

\noindent\textbf{Structured outputs provide modest benefits when learned from scratch.}
Introducing a cognitive map structure provides only marginal improvement (\texttt{RL-Aug-CGMap-FFR-Out}: 52.48\%, \texttt{RL-Plain-CGMap-FFR-Out}: 50.86\%). In both cases, the model fails to learn meaningful geometry, with low similarity scores and near-zero isomorphism rates. This suggests that without a prior concept of a "good" map, RL struggles to exploit the provided structural format, even if it can learn to fill it out validly.

\noindent\textbf{RL performs better when it trains from SFT checkpoint.}
The most substantial improvements occur when warm-starting RL from an optimal SFT checkpoint. All three from-SFT configurations significantly outperform their from-scratch counterparts, with \texttt{RL-Plain-CGMap-FFR-Out (from SFT)} achieving the highest accuracy of 61.33\% ($\uparrow$3.52\% over the best SFT model, $\uparrow$8.85\% over the best RL-from-scratch). Notably, even \texttt{RL-FFR (from SFT)} reaches 59.14\%, confirming that SFT initialization is critical. However, the map-then-reason configurations consistently outperform FFR-only, reinforcing the advantage of structured spatial scaffolding. The \texttt{Plain-CGMap} variant continues to produce geometrically superior maps (35.33\% vs. 16.95\% isomorphism rate), suggesting that simpler map formats allow RL to better preserve spatial structure. These results indicate that RL's primary role is polishing and refining the strong priors learned during SFT, and raising the performance ceiling beyond what SFT alone can achieve.

\begin{promptbox}{\faLightbulb\ Key Takeaways: Reinforcement Learning for the Emergence of Spatial Mental Models}
\it
\begin{itemize}[leftmargin=*, after=\vspace{-0.5em}, itemsep=1pt]
\item Combining cognitive maps with reasoning consistently improves all learning outcomes.
\item Starting from scratch, RL provides only marginal gains for spatial reasoning; its true power is unlocked when building upon a strong SFT foundation.
\end{itemize}
\end{promptbox}

\subsection{Effect of Object Presentation Order in Cognitive Map Supervision}
\label{sec:object_order}

When constructing cognitive map supervision, the order in which objects are listed in the textual map description is a design choice that may influence learning dynamics. We investigate two settings: (1) \textbf{Fixed Spatial Order}, where objects follow a consistent spatial convention (e.g., clockwise from the camera's viewpoint), and (2) \textbf{Randomized Order}, where the object sequence is shuffled independently for each training example. We examine this factor for both SFT and RL, since the SFT checkpoint also serves as the initialization for RL training. All primary results reported in Table~\ref{tab:combined_results} use randomized order.

\begin{table}[htbp]
    \centering
    \caption{SFT and RL results under fixed spatial object order. The map-then-reason approach (\texttt{Plain-CGMap-FFR-Out}) remains the best-performing configuration, consistent with the randomized-order results in Table~\ref{tab:combined_results}.}
    \scriptsize
    \vspace*{-0.5em}
    \resizebox{\linewidth}{!}{%
    \begin{tabular}{>{\centering\arraybackslash}p{2pt}lcccccc}
    \hline
    & \rule{0pt}{2.5ex}\textbf{Config.} & \makecell{\textbf{Overall}} & \makecell{\textbf{Rotation}} & \makecell{\textbf{Among}} & \makecell{\textbf{Around}} & \makecell{\textbf{Overall Sim.}} & \makecell{\textbf{Isom. Rate}} \\
    \hline
    \multirow{6}{*}{\makebox[5pt][c]{\rotatebox[origin=c]{90}{\scriptsize\textbf{SFT}}}}
    & \texttt{Raw-QA} & 52.28 & 34.50 & 52.50 & 66.00 & -- & -- \\
    & \texttt{FFR} & 53.52\up & 36.00 & 54.67 & 64.80 & -- & -- \\
    & \texttt{Aug-CGMap-Out} & 54.19\up & 35.50 & 53.17 & \textbf{71.60} & 74.30 & 43.24 \\
    & \texttt{Plain-CGMap-Out} & 54.38\up & 35.50 & 53.50 & \textbf{71.60} & \textbf{91.73} & \textbf{89.05} \\
    & \texttt{Aug-CGMap-FFR-Out} & 55.24\up & \textbf{49.50} & 52.50 & 66.40 & 75.27 & 46.00 \\
    & \texttt{Plain-CGMap-FFR-Out} & \textbf{60.76}\up & 47.50 & \textbf{62.33} & 67.60 & 88.79 & 73.81 \\
    \hline
    \multirow{2}{*}{\makebox[5pt][c]{\rotatebox[origin=c]{90}{\scriptsize\textbf{RL}}}}
    & \texttt{RL-Aug-CGMap-FFR-Out (from SFT)} & \textbf{70.67} & \textbf{53.00} & 76.83 & \textbf{70.00} & 85.53 & 58.86 \\
    & \texttt{RL-Plain-CGMap-FFR-Out (from SFT)} & \textbf{70.67} & 48.00 & \textbf{79.17} & 68.40 & \textbf{85.79} & \textbf{71.52} \\
    \hline
    \end{tabular}
    }
    \label{tab:object_order}
\end{table}

\begin{figure}[tbp]
  \centering
  \includegraphics[width=\linewidth]{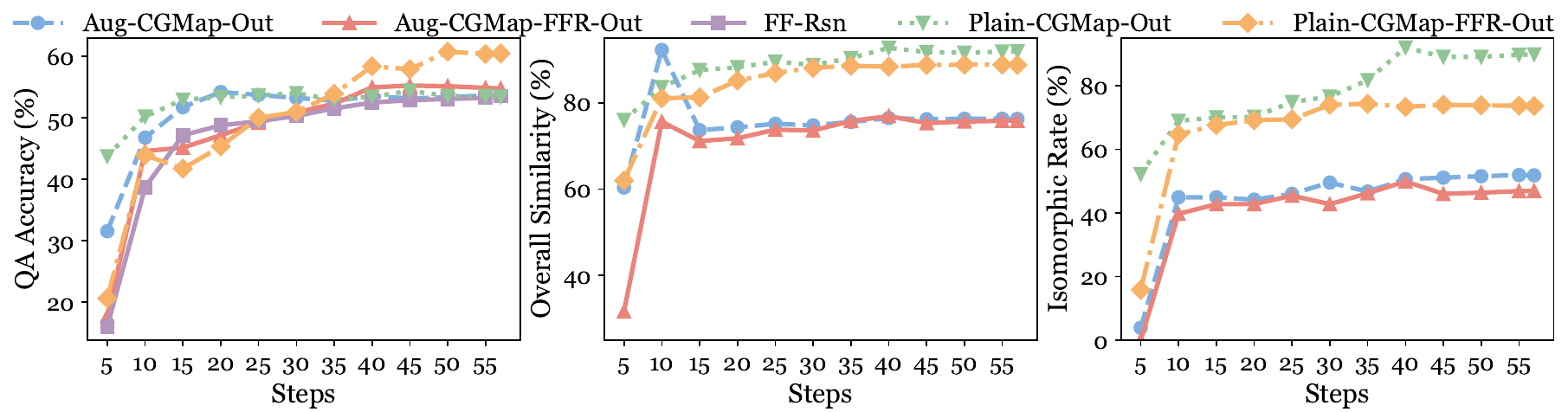}
  \caption{SFT training dynamics under fixed spatial object order. Compared to randomized order (Figure~\ref{fig:sft-whole}), the overall learning trends are consistent: \texttt{Plain-CGMap-FFR-Out} achieves the highest QA accuracy despite not producing the most structurally perfect maps.}
  \label{fig:sft-fixed-order}
\end{figure}

\noindent\textbf{Both settings exhibit consistent trends.} As shown in Table~\ref{tab:object_order} and Table~\ref{tab:combined_results}, the relative ranking of configurations is preserved regardless of object order: \texttt{Plain-CGMap-FFR-Out} consistently achieves the highest QA accuracy in SFT, and RL from SFT continues to yield the strongest overall results. The training dynamics (Figure~\ref{fig:sft-whole} vs. Figure~\ref{fig:sft-fixed-order}) further confirm that in both settings, models trained solely on map generation learn the target structure rapidly but plateau in QA accuracy, while the joint map-and-reasoning model learns maps more slowly yet continues to improve on the end task.

\noindent\textbf{Randomized order better evaluates genuine spatial understanding.} Under fixed spatial order, models achieve notably higher map reconstruction quality (e.g., 89.05\% vs. 45.52\% isomorphism rate for \texttt{Plain-CGMap-Out}), as the deterministic sequence provides a predictable pattern that the model can exploit without necessarily developing robust internal spatial representations. In contrast, randomized order forces the model to construct its own spatial understanding from scratch for each example, without relying on ordering shortcuts. We therefore adopt randomized order as our primary setting (Table~\ref{tab:combined_results}), as it more faithfully reflects the model's ability to build genuine internal representations---the central goal of our investigation. We present the fixed-order results here for completeness, noting that the core conclusions hold under both settings

\vspace*{-0.3em}
\section{Related Works}

\textbf{Spatial Cognition.} 
Spatial cognition encompasses skills like mental rotation, spatial visualization, and object assembly, essential for perceiving and manipulating spatial relationships in both 2D and 3D environments~\citep{xu2025definingevaluatingvisuallanguage, zha2025enablellm3dcapacity, wang2025sitespatialintelligencethorough}. At the core of these abilities are Spatial Mental Models (SMMs)~\citep{johnson1980mental, johnson1983mental}, which are internal representations that allow for consistent understanding of space.
Recently, much effort has been dedicated to evaluating spatial cognition in VLMs~\citep{zhan2025open3dvqabenchmarkcomprehensivespatial, ma20253dsrbenchcomprehensive3dspatial, lee2025perspectiveawarereasoningvisionlanguagemodels, zhang2025sphereunveilingspatialblind}. Moreover, some methods are proposed to enhance spatial understanding, such as coordinate-aware prompting~\citep{cai2024spatialbot}, CoT reasoning~\citep{ma2025spatialreasonerexplicitgeneralizable3d, liu2025spatialcotadvancingspatialreasoning}, explicit spatial representation alignment~\citep{cheng2024spatialrgptgroundedspatialreasoning, space-mantis}, and an RL-based approach~\citep{pan2025metaspatial}. However, existing benchmarks~\citep{lee2025perspectiveawarereasoningvisionlanguagemodels, zhan2025open3dvqabenchmarkcomprehensivespatial, chen2025spatialreasoninghardvlms, qi2025semanticsrediscoveringspatialawareness, zhang2025sphereunveilingspatialblind, ma20253dsrbenchcomprehensive3dspatial, ramakrishnan2025doesspatialcognitionemerge, tang2025sparklemasteringbasicspatial, fu2024blink,yang2024thinkingspacemultimodallarge,zhang2024task} and approaches often neglect the mental-level spatial reasoning that underpins human cognition, leaving a gap between machine and human capabilities.
To bridge this gap, a new approach is needed that trains VLMs to reason about space not only through visual data but also through mental-level spatial reasoning, aligning more closely with human spatial cognition.

\noindent\textbf{Multi Views understanding.} 
Multiview spatial understanding leverages multiple viewpoints to reconstruct 3D structures and overcome single-view limitations. Efficient techniques optimize view processing, while reconstruction methods~\citep{wang2025vggt,liu2025citygaussianv2,fu2025maskgaussian,qu2025drag},  view synthesis methods~\citep{sun2018multiview,zhang2024viewcrafter,sargent2023zeronvs} and multiview equivariant learning~\citep{you2024multiviewequivarianceimproves3d} enhance geometric consistency. Topological representations like \cite{zhang2024multiviewscenegraph} encode object relations for holistic reasoning, while frameworks such as \cite{hong20233dconceptlearningreasoning} advance open-vocabulary concept learning from multiview data via neural fields and vision-language fusion.  LMMs augmented with multiview inputs~\citep{daxberger2025mmspatialexploring3dspatial,wu2025spatial,fan2025vlm3rvisionlanguagemodelsaugmented,zheng2025learningvideos3dworld,lee2025perspectiveawarereasoningvisionlanguagemodels,zhao2025embodiedrcollaborativeframeworkactivating,xu2025multispatialmllmmultiframespatialunderstanding} demonstrate marked improvements in spatial tasks like geometric understanding and perspective taking. Yet, they struggle with multiview consistency understanding due to fragmented reasoning and 2D-to-3D projection ambiguities, leaving a gap for robust spatial AI.

\vspace*{-1em}
\section{Conclusion}
\vspace*{-1em}
We introduced \name to study how VLMs can approximate spatial mental models from limited views, a core cognitive ability for reasoning in partially observable environments. Moving beyond benchmarking, we explored \textit{how} internal representations can be scaffolded through structured data and reasoning. Our key finding is that \textit{constructing and reasoning over self-generated cognitive maps}, rather than relying on view interpolation or externally provided maps, yields the most effective approximation of spatial mental models across all elicitation methods (input-output configurations, supervised fine-tuning, and reinforcement learning). Initializing RL from a well-trained SFT checkpoint further optimizes the process, further improving spatial reasoning performance.
\section*{Ethics Statement}
The \name benchmark was developed using a combination of publicly available, anonymized datasets (ArkitScenes, WildRGB-D, DL3DV-10K) and self-collected imagery. For our self-collected data, care was taken to capture indoor and outdoor scenes without including personally identifiable information (PII) or sensitive content. All human annotators involved in the data curation and evaluation phases were compensated at rates significantly exceeding their local minimum wage.

We acknowledge several limitations and ethical considerations. The datasets used, while diverse, may not fully represent the vast range of global environments, potentially introducing geographic or cultural biases into the model's spatial understanding. Furthermore, the training, fine-tuning, and evaluation of the large-scale Vision-Language Models discussed in this paper carry a significant computational and environmental cost. While our research is intended to advance the scientific understanding of AI cognition, we recognize that technologies enhancing spatial reasoning in machines could have dual-use applications.

\section*{Reproducibility Statement}
To ensure the reproducibility of our findings, we have included our complete codebase for data processing, model training, and evaluation in the supplementary materials as a .zip archive. Furthermore, the full \name benchmark, encompassing all of our training data, test data, annotations, and evaluation protocols, will be released in a public repository to facilitate further research and verification by the community.

\subsubsection*{Acknowledgments}
This work is in part supported by the Stanford Institute for Human-Centered AI (HAI), ONR N00014-23-1-2355, ONR MURI N00014-22-1-2740, ONR MURI N00014-21-1-2801, and DSO National Laboratories Agreement DSOCO25017.

{
\small
\bibliographystyle{unsrt}
\bibliography{iclr2026_conference}

@misc{wu2025reinforcingspatialreasoningvisionlanguage,
      title={Reinforcing Spatial Reasoning in Vision-Language Models with Interwoven Thinking and Visual Drawing}, 
      author={Junfei Wu and Jian Guan and Kaituo Feng and Qiang Liu and Shu Wu and Liang Wang and Wei Wu and Tieniu Tan},
      year={2025},
      eprint={2506.09965},
      archivePrefix={arXiv},
      primaryClass={cs.CV},
      url={https://arxiv.org/abs/2506.09965}, 
}

@misc{geminiteam2025geminifamilyhighlycapable,
      title={Gemini: A Family of Highly Capable Multimodal Models}, 
      author={Gemini Team},
      year={2025},
      eprint={2312.11805},
      archivePrefix={arXiv},
      primaryClass={cs.CL},
      url={https://arxiv.org/abs/2312.11805}, 
}

@misc{zhu2025internvl3exploringadvancedtraining,
      title={InternVL3: Exploring Advanced Training and Test-Time Recipes for Open-Source Multimodal Models}, 
      author={Jinguo Zhu and Weiyun Wang and Zhe Chen and Zhaoyang Liu and Shenglong Ye and Lixin Gu and Hao Tian and Yuchen Duan and Weijie Su and Jie Shao and Zhangwei Gao and Erfei Cui and Xuehui Wang and Yue Cao and Yangzhou Liu and Xingguang Wei and Hongjie Zhang and Haomin Wang and Weiye Xu and Hao Li and Jiahao Wang and Nianchen Deng and Songze Li and Yinan He and Tan Jiang and Jiapeng Luo and Yi Wang and Conghui He and Botian Shi and Xingcheng Zhang and Wenqi Shao and Junjun He and Yingtong Xiong and Wenwen Qu and Peng Sun and Penglong Jiao and Han Lv and Lijun Wu and Kaipeng Zhang and Huipeng Deng and Jiaye Ge and Kai Chen and Limin Wang and Min Dou and Lewei Lu and Xizhou Zhu and Tong Lu and Dahua Lin and Yu Qiao and Jifeng Dai and Wenhai Wang},
      year={2025},
      eprint={2504.10479},
      archivePrefix={arXiv},
      primaryClass={cs.CV},
      url={https://arxiv.org/abs/2504.10479}, 
}

@techreport{OpenAI_GPT5_2025,
  author = {{OpenAI}},
  title = {{GPT-5 System Card}},
  institution = {{OpenAI}},
  year = {2025},
  month = {aug},
  note = {Accessed: 2025-08-10},
  url = {https://cdn.openai.com/gpt-5-system-card.pdf}
}

@misc{lu2024deepseekvlrealworldvisionlanguageunderstanding,
      title={DeepSeek-VL: Towards Real-World Vision-Language Understanding}, 
      author={Haoyu Lu and Wen Liu and Bo Zhang and Bingxuan Wang and Kai Dong and Bo Liu and Jingxiang Sun and Tongzheng Ren and Zhuoshu Li and Hao Yang and Yaofeng Sun and Chengqi Deng and Hanwei Xu and Zhenda Xie and Chong Ruan},
      year={2024},
      eprint={2403.05525},
      archivePrefix={arXiv},
      primaryClass={cs.AI},
      url={https://arxiv.org/abs/2403.05525}, 
}

@misc{cheng2024spatialrgptgroundedspatialreasoning,
      title={SpatialRGPT: Grounded Spatial Reasoning in Vision Language Models}, 
      author={An-Chieh Cheng and Hongxu Yin and Yang Fu and Qiushan Guo and Ruihan Yang and Jan Kautz and Xiaolong Wang and Sifei Liu},
      year={2024},
      eprint={2406.01584},
      archivePrefix={arXiv},
      primaryClass={cs.CV},
      url={https://arxiv.org/abs/2406.01584}, 
}

@inproceedings{zhang2024task,
  title={Task Me Anything},
  author={Zhang, Jieyu and Huang, Weikai and Ma, Zixian and Michel, Oscar and He, Dong and Gupta, Tanmay and Ma, Wei-Chiu and Farhadi, Ali and Kembhavi, Aniruddha and Krishna, Ranjay},
  booktitle={Thirty-Eighth Annual Conference on Neural Information Processing Systems Datasets and Benchmarks Track},
  year={2024}
}

@misc{hong20233dconceptlearningreasoning,
      title={3D Concept Learning and Reasoning from Multi-View Images}, 
      author={Yining Hong and Chunru Lin and Yilun Du and Zhenfang Chen and Joshua B. Tenenbaum and Chuang Gan},
      year={2023},
      eprint={2303.11327},
      archivePrefix={arXiv},
      primaryClass={cs.CV},
      url={https://arxiv.org/abs/2303.11327}, 
}

@article{cai2024spatialbot,
  title={SpatialBot: Precise Spatial Understanding with Vision Language Models},
  author={Cai, Wenxiao and Ponomarenko, Yaroslav and Yuan, Jianhao and Li, Xiaoqi and Yang, Wankou and Dong, Hao and Zhao, Bo},
  journal={arXiv preprint arXiv:2406.13642},
  year={2024}
}

@misc{ling2023dl3dv10klargescalescenedataset,
      title={DL3DV-10K: A Large-Scale Scene Dataset for Deep Learning-based 3D Vision}, 
      author={Lu Ling and Yichen Sheng and Zhi Tu and Wentian Zhao and Cheng Xin and Kun Wan and Lantao Yu and Qianyu Guo and Zixun Yu and Yawen Lu and Xuanmao Li and Xingpeng Sun and Rohan Ashok and Aniruddha Mukherjee and Hao Kang and Xiangrui Kong and Gang Hua and Tianyi Zhang and Bedrich Benes and Aniket Bera},
      year={2023},
      eprint={2312.16256},
      archivePrefix={arXiv},
      primaryClass={cs.CV},
      url={https://arxiv.org/abs/2312.16256}, 
}

@misc{gpt4o_blog,
    author = {OpenAI},
    title = {Hello GPT-4o},
    year = {2024},
    month = {05},
    howpublished = {Blog},
    url = {https://openai.com/index/hello-gpt-4o/},
    note = {Accessed: November 22, 2024}
}

@article{johnson1980mental,
  title={Mental models in cognitive science},
  author={Johnson-Laird, Philip N},
  journal={Cognitive science},
  volume={4},
  number={1},
  pages={71--115},
  year={1980},
  publisher={Elsevier}
}

@book{johnson1983mental,
  title={Mental models: Towards a cognitive science of language, inference, and consciousness},
  author={Johnson-Laird, Philip Nicholas},
  number={6},
  year={1983},
  publisher={Harvard University Press}
}

@article{onevision,
  title={Llava-onevision: Easy visual task transfer},
  author={Li, Bo and Zhang, Yuanhan and Guo, Dong and Zhang, Renrui and Li, Feng and Zhang, Hao and Zhang, Kaichen and Zhang, Peiyuan and Li, Yanwei and Liu, Ziwei and others},
  journal={arXiv preprint arXiv:2408.03326},
  year={2024}
}

@article{llava-video,
  title={Video instruction tuning with synthetic data},
  author={Zhang, Yuanhan and Wu, Jinming and Li, Wei and Li, Bo and Ma, Zejun and Liu, Ziwei and Li, Chunyuan},
  journal={arXiv preprint arXiv:2410.02713},
  year={2024}
}

@article{mplug-owl3,
  title={mplug-owl3: Towards long image-sequence understanding in multi-modal large language models},
  author={Ye, Jiabo and Xu, Haiyang and Liu, Haowei and Hu, Anwen and Yan, Ming and Qian, Qi and Zhang, Ji and Huang, Fei and Zhou, Jingren},
  journal={arXiv preprint arXiv:2408.04840},
  year={2024}
}

@inproceedings{chen2024internvl,
title={Internvl: Scaling up vision foundation models and aligning for generic visual-linguistic tasks},
author={Chen, Zhe and Wu, Jiannan and Wang, Wenhai and Su, Weijie and Chen, Guo and Xing, Sen and Zhong, Muyan and Zhang, Qinglong and Zhu, Xizhou and Lu, Lewei and others},
booktitle={Proceedings of the IEEE/CVF Conference on Computer Vision and Pattern Recognition},
pages={24185--24198},
year={2024}
}

@article{Qwen2.5-VL,
  title={Qwen2.5-VL Technical Report},
  author={Bai, Shuai and Chen, Keqin and Liu, Xuejing and Wang, Jialin and Ge, Wenbin and Song, Sibo and Dang, Kai and Wang, Peng and Wang, Shijie and Tang, Jun and Zhong, Humen and Zhu, Yuanzhi and Yang, Mingkun and Li, Zhaohai and Wan, Jianqiang and Wang, Pengfei and Ding, Wei and Fu, Zheren and Xu, Yiheng and Ye, Jiabo and Zhang, Xi and Xie, Tianbao and Cheng, Zesen and Zhang, Hang and Yang, Zhibo and Xu, Haiyang and Lin, Junyang},
  journal={arXiv preprint arXiv:2502.13923},
  year={2025}
}

@article{longva,
  title={Long context transfer from language to vision},
  author={Zhang, Peiyuan and Zhang, Kaichen and Li, Bo and Zeng, Guangtao and Yang, Jingkang and Zhang, Yuanhan and Wang, Ziyue and Tan, Haoran and Li, Chunyuan and Liu, Ziwei},
  journal={arXiv preprint arXiv:2406.16852},
  year={2024}
}

@article{mantis,
  title={MANTIS: Interleaved Multi-Image Instruction Tuning},
  author={Dongfu Jiang and Xuan He and Huaye Zeng and Cong Wei and Max W.F. Ku and Qian Liu and Wenhu Chen},
  journal={Transactions on Machine Learning Research},
  year={2024},
  volume={2024},
  url={https://openreview.net/forum?id=skLtdUVaJa}
}

@article{ji2025robobrain,
  title={Robobrain: A unified brain model for robotic manipulation from abstract to concrete},
  author={Ji, Yuheng and Tan, Huajie and Shi, Jiayu and Hao, Xiaoshuai and Zhang, Yuan and Zhang, Hengyuan and Wang, Pengwei and Zhao, Mengdi and Mu, Yao and An, Pengju and others},
  journal={arXiv preprint arXiv:2502.21257},
  year={2025}
}

@article{space-mantis,
  title = {SpatialVLM: Endowing Vision-Language Models with Spatial Reasoning Capabilities},
  author = {Chen, Boyuan and Xu, Zhuo and Kirmani, Sean and Ichter, Brian and Driess, Danny and Florence, Pete and Sadigh, Dorsa and Guibas, Leonidas and Xia, Fei},
  journal = {arXiv preprint arXiv:2401.12168},
  year = {2024},
  url = {https://arxiv.org/abs/2401.12168},
}

@inproceedings{tversky1993cognitive,
  title={Cognitive maps, cognitive collages, and spatial mental models},
  author={Tversky, Barbara},
  booktitle={European conference on spatial information theory},
  pages={14--24},
  year={1993},
  organization={Springer}
}

@article{tversky2003structures,
  title={Structures of mental spaces: How people think about space},
  author={Tversky, Barbara},
  journal={Environment and behavior},
  volume={35},
  number={1},
  pages={66--80},
  year={2003},
  publisher={Sage Publications}
}

@article{tversky1994spatial,
  title={Spatial mental models from descriptions},
  author={Tversky, Barbara and Franklin, Nancy and Taylor, Holly A and Bryant, David J},
  journal={Journal of the American society for information science},
  volume={45},
  number={9},
  pages={656--668},
  year={1994},
  publisher={Wiley Online Library}
}

@article{hegarty1992mental,
  title={Mental animation: Inferring motion from static displays of mechanical systems.},
  author={Hegarty, Mary},
  journal={Journal of experimental psychology: learning, memory, and cognition},
  volume={18},
  number={5},
  pages={1084},
  year={1992},
  publisher={American Psychological Association}
}

@article{zhou2025stable,
  title={STABLE VIRTUAL CAMERA: Generative View Synthesis with Diffusion Models},
  author={Zhou, Jensen Jinghao and Gao, Hang and Voleti, Vikram and Vasishta, Aaryaman and Yao, Chun-Han and Boss, Mark and Torr, Philip and Rupprecht, Christian and Jampani, Varun},
  journal={arXiv preprint arXiv:2503.14489},
  year={2025}
}

@article{guo2025deepseek,
  title={Deepseek-r1: Incentivizing reasoning capability in llms via reinforcement learning},
  author={Guo, Daya and Yang, Dejian and Zhang, Haowei and Song, Junxiao and Zhang, Ruoyu and Xu, Runxin and Zhu, Qihao and Ma, Shirong and Wang, Peiyi and Bi, Xiao and others},
  journal={arXiv preprint arXiv:2501.12948},
  year={2025}
}

@article{shao2024deepseekmath,
  title={Deepseekmath: Pushing the limits of mathematical reasoning in open language models},
  author={Shao, Zhihong and Wang, Peiyi and Zhu, Qihao and Xu, Runxin and Song, Junxiao and Bi, Xiao and Zhang, Haowei and Zhang, Mingchuan and Li, YK and Wu, Y and others},
  journal={arXiv preprint arXiv:2402.03300},
  year={2024}
}

@article{yeh2025seeing,
  title={Seeing from Another Perspective: Evaluating Multi-View Understanding in MLLMs},
  author={Yeh, Chun-Hsiao and Wang, Chenyu and Tong, Shengbang and Cheng, Ta-Ying and Wang, Rouyu and Chu, Tianzhe and Zhai, Yuexiang and Chen, Yubei and Gao, Shenghua and Ma, Yi},
  journal={arXiv preprint arXiv:2504.15280},
  year={2025}
}

@article{fu2024blink,
          title={BLINK: Multimodal Large Language Models Can See but Not Perceive},
          author={Fu, Xingyu and Hu, Yushi and Li, Bangzheng and Feng, Yu and Wang, Haoyu and Lin, Xudong and Roth, Dan and Smith, Noah A and Ma, Wei-Chiu and Krishna, Ranjay},
          journal={arXiv preprint arXiv:2404.12390},
          year={2024}
        }

@misc{wang2025vagen,
  title={Reinforcing Visual State Reasoning for Multi-Turn VLM Agents},
  author={Kangrui Wang* and Pingyue Zhang* and Zihan Wang* and Yaning Gao* and Linjie Li* and Qineng Wang and Hanyang Chen and Chi Wan and Yiping Lu and Zhengyuan Yang and Lijuan Wang and Ranjay Krishna and Jiajun Wu and Li Fei-Fei and Yejin Choi and Manling Li},
  year={2025},
  url={https://github.com/RAGEN-AI/VAGEN}
}

@misc{xu2025definingevaluatingvisuallanguage,
      title={Defining and Evaluating Visual Language Models' Basic Spatial Abilities: A Perspective from Psychometrics}, 
      author={Wenrui Xu and Dalin Lyu and Weihang Wang and Jie Feng and Chen Gao and Yong Li},
      year={2025},
      eprint={2502.11859},
      archivePrefix={arXiv},
      primaryClass={cs.CV},
      url={https://arxiv.org/abs/2502.11859}, 
}

@misc{lee2025perspectiveawarereasoningvisionlanguagemodels,
      title={Perspective-Aware Reasoning in Vision-Language Models via Mental Imagery Simulation}, 
      author={Phillip Y. Lee and Jihyeon Je and Chanho Park and Mikaela Angelina Uy and Leonidas Guibas and Minhyuk Sung},
      year={2025},
      eprint={2504.17207},
      archivePrefix={arXiv},
      primaryClass={cs.CV},
      url={https://arxiv.org/abs/2504.17207}, 
}

@misc{zha2025enablellm3dcapacity,
      title={How to Enable LLM with 3D Capacity? A Survey of Spatial Reasoning in LLM}, 
      author={Jirong Zha and Yuxuan Fan and Xiao Yang and Chen Gao and Xinlei Chen},
      year={2025},
      eprint={2504.05786},
      archivePrefix={arXiv},
      primaryClass={cs.CV},
      url={https://arxiv.org/abs/2504.05786}, 
}

@misc{wang2025sitespatialintelligencethorough,
      title={SITE: towards Spatial Intelligence Thorough Evaluation}, 
      author={Wenqi Wang and Reuben Tan and Pengyue Zhu and Jianwei Yang and Zhengyuan Yang and Lijuan Wang and Andrey Kolobov and Jianfeng Gao and Boqing Gong},
      year={2025},
      eprint={2505.05456},
      archivePrefix={arXiv},
      primaryClass={cs.CV},
      url={https://arxiv.org/abs/2505.05456}, 
}

@misc{zhan2025open3dvqabenchmarkcomprehensivespatial,
      title={Open3DVQA: A Benchmark for Comprehensive Spatial Reasoning with Multimodal Large Language Model in Open Space}, 
      author={Weichen Zhan and Zile Zhou and Zhiheng Zheng and Chen Gao and Jinqiang Cui and Yong Li and Xinlei Chen and Xiao-Ping Zhang},
      year={2025},
      eprint={2503.11094},
      archivePrefix={arXiv},
      primaryClass={cs.CV},
      url={https://arxiv.org/abs/2503.11094}, 
}

@misc{chen2025spatialreasoninghardvlms,
      title={Why Is Spatial Reasoning Hard for VLMs? An Attention Mechanism Perspective on Focus Areas}, 
      author={Shiqi Chen and Tongyao Zhu and Ruochen Zhou and Jinghan Zhang and Siyang Gao and Juan Carlos Niebles and Mor Geva and Junxian He and Jiajun Wu and Manling Li},
      year={2025},
      eprint={2503.01773},
      archivePrefix={arXiv},
      primaryClass={cs.CL},
      url={https://arxiv.org/abs/2503.01773}, 
}

@misc{qi2025semanticsrediscoveringspatialawareness,
      title={Beyond Semantics: Rediscovering Spatial Awareness in Vision-Language Models}, 
      author={Jianing Qi and Jiawei Liu and Hao Tang and Zhigang Zhu},
      year={2025},
      eprint={2503.17349},
      archivePrefix={arXiv},
      primaryClass={cs.CV},
      url={https://arxiv.org/abs/2503.17349}, 
}

@misc{zhang2025sphereunveilingspatialblind,
      title={SPHERE: Unveiling Spatial Blind Spots in Vision-Language Models Through Hierarchical Evaluation}, 
      author={Wenyu Zhang and Wei En Ng and Lixin Ma and Yuwen Wang and Jungqi Zhao and Allison Koenecke and Boyang Li and Lu Wang},
      year={2025},
      eprint={2412.12693},
      archivePrefix={arXiv},
      primaryClass={cs.CV},
      url={https://arxiv.org/abs/2412.12693}, 
}

@misc{ma20253dsrbenchcomprehensive3dspatial,
      title={3DSRBench: A Comprehensive 3D Spatial Reasoning Benchmark}, 
      author={Wufei Ma and Haoyu Chen and Guofeng Zhang and Yu-Cheng Chou and Celso M de Melo and Alan Yuille},
      year={2025},
      eprint={2412.07825},
      archivePrefix={arXiv},
      primaryClass={cs.CV},
      url={https://arxiv.org/abs/2412.07825}, 
}

@misc{ramakrishnan2025doesspatialcognitionemerge,
      title={Does Spatial Cognition Emerge in Frontier Models?}, 
      author={Santhosh Kumar Ramakrishnan and Erik Wijmans and Philipp Kraehenbuehl and Vladlen Koltun},
      year={2025},
      eprint={2410.06468},
      archivePrefix={arXiv},
      primaryClass={cs.AI},
      url={https://arxiv.org/abs/2410.06468}, 
}

@misc{tang2025sparklemasteringbasicspatial,
      title={Sparkle: Mastering Basic Spatial Capabilities in Vision Language Models Elicits Generalization to Spatial Reasoning}, 
      author={Yihong Tang and Ao Qu and Zhaokai Wang and Dingyi Zhuang and Zhaofeng Wu and Wei Ma and Shenhao Wang and Yunhan Zheng and Zhan Zhao and Jinhua Zhao},
      year={2025},
      eprint={2410.16162},
      archivePrefix={arXiv},
      primaryClass={cs.CV},
      url={https://arxiv.org/abs/2410.16162}, 
}

@misc{daxberger2025mmspatialexploring3dspatial,
      title={MM-Spatial: Exploring 3D Spatial Understanding in Multimodal LLMs}, 
      author={Erik Daxberger and Nina Wenzel and David Griffiths and Haiming Gang and Justin Lazarow and Gefen Kohavi and Kai Kang and Marcin Eichner and Yinfei Yang and Afshin Dehghan and Peter Grasch},
      year={2025},
      eprint={2503.13111},
      archivePrefix={arXiv},
      primaryClass={cs.CV},
      url={https://arxiv.org/abs/2503.13111}, 
}

@misc{ma2025spatialreasonerexplicitgeneralizable3d,
      title={SpatialReasoner: Towards Explicit and Generalizable 3D Spatial Reasoning}, 
      author={Wufei Ma and Yu-Cheng Chou and Qihao Liu and Xingrui Wang and Celso de Melo and Jieneng Chen and Jianwen Xie and Alan Yuille},
      year={2025},
      eprint={2504.20024},
      archivePrefix={arXiv},
      primaryClass={cs.CV},
      url={https://arxiv.org/abs/2504.20024}, 
}

@misc{liu2025spatialcotadvancingspatialreasoning,
      title={SpatialCoT: Advancing Spatial Reasoning through Coordinate Alignment and Chain-of-Thought for Embodied Task Planning}, 
      author={Yuecheng Liu and Dafeng Chi and Shiguang Wu and Zhanguang Zhang and Yaochen Hu and Lingfeng Zhang and Yingxue Zhang and Shuang Wu and Tongtong Cao and Guowei Huang and Helong Huang and Guangjian Tian and Weichao Qiu and Xingyue Quan and Jianye Hao and Yuzheng Zhuang},
      year={2025},
      eprint={2501.10074},
      archivePrefix={arXiv},
      primaryClass={cs.RO},
      url={https://arxiv.org/abs/2501.10074}, 
}

@misc{yang2024thinkingspacemultimodallarge,
      title={Thinking in Space: How Multimodal Large Language Models See, Remember, and Recall Spaces}, 
      author={Jihan Yang and Shusheng Yang and Anjali W. Gupta and Rilyn Han and Li Fei-Fei and Saining Xie},
      year={2024},
      eprint={2412.14171},
      archivePrefix={arXiv},
      primaryClass={cs.CV},
      url={https://arxiv.org/abs/2412.14171}, 
}

@misc{li2023blip2bootstrappinglanguageimagepretraining,
      title={BLIP-2: Bootstrapping Language-Image Pre-training with Frozen Image Encoders and Large Language Models}, 
      author={Junnan Li and Dongxu Li and Silvio Savarese and Steven Hoi},
      year={2023},
      eprint={2301.12597},
      archivePrefix={arXiv},
      primaryClass={cs.CV},
      url={https://arxiv.org/abs/2301.12597}, 
}

@article{pan2025metaspatial,
  title={MetaSpatial: Reinforcing 3D Spatial Reasoning in VLMs for the Metaverse},
  author={Pan, Zhenyu and Liu, Han},
  journal={arXiv preprint arXiv:2503.18470},
  year={2025}
}

@inproceedings{xia2024rgbd,
  title={RGBD objects in the wild: scaling real-world 3D object learning from RGB-D videos},
  author={Xia, Hongchi and Fu, Yang and Liu, Sifei and Wang, Xiaolong},
  booktitle={Proceedings of the IEEE/CVF Conference on Computer Vision and Pattern Recognition},
  pages={22378--22389},
  year={2024}
}

@inproceedings{baruch1arkitscenes,
  title={ARKitScenes: A Diverse Real-World Dataset For 3D Indoor Scene Understanding Using Mobile RGB-D Data},
  author={Baruch, Gilad and Chen, Zhuoyuan and Dehghan, Afshin and Feigin, Yuri and Fu, Peter and Gebauer, Thomas and Kurz, Daniel and Dimry, Tal and Joffe, Brandon and Schwartz, Arik and others},
  booktitle={Thirty-fifth Conference on Neural Information Processing Systems Datasets and Benchmarks Track (Round 1)},
  year={2021}
}

@article{wei2022chain,
  title={Chain-of-thought prompting elicits reasoning in large language models},
  author={Wei, Jason and Wang, Xuezhi and Schuurmans, Dale and Bosma, Maarten and Xia, Fei and Chi, Ed and Le, Quoc V and Zhou, Denny and others},
  journal={Advances in neural information processing systems},
  volume={35},
  pages={24824--24837},
  year={2022}
}

@article{li2024embodied,
  title={Embodied agent interface: Benchmarking llms for embodied decision making},
  author={Li, Manling and Zhao, Shiyu and Wang, Qineng and Wang, Kangrui and Zhou, Yu and Srivastava, Sanjana and Gokmen, Cem and Lee, Tony and Li, Erran Li and Zhang, Ruohan and others},
  journal={Advances in Neural Information Processing Systems},
  volume={37},
  pages={100428--100534},
  year={2024}
}

@article{yang2025embodiedbench,
  title={EmbodiedBench: Comprehensive Benchmarking Multi-modal Large Language Models for Vision-Driven Embodied Agents},
  author={Yang, Rui and Chen, Hanyang and Zhang, Junyu and Zhao, Mark and Qian, Cheng and Wang, Kangrui and Wang, Qineng and Koripella, Teja Venkat and Movahedi, Marziyeh and Li, Manling and others},
  journal={arXiv preprint arXiv:2502.09560},
  year={2025}
}

@article{wang2024rethinking,
  title={Rethinking the bounds of LLM reasoning: Are multi-agent discussions the key?},
  author={Wang, Qineng and Wang, Zihao and Su, Ying and Tong, Hanghang and Song, Yangqiu},
  journal={arXiv preprint arXiv:2402.18272},
  year={2024}
}

@inproceedings{liang2023code,
  title={Code as policies: Language model programs for embodied control},
  author={Liang, Jacky and Huang, Wenlong and Xia, Fei and Xu, Peng and Hausman, Karol and Ichter, Brian and Florence, Pete and Zeng, Andy},
  booktitle={2023 IEEE International Conference on Robotics and Automation (ICRA)},
  pages={9493--9500},
  year={2023},
  organization={IEEE}
}

@inproceedings{driess2023palm,
  title={PaLM-E: an embodied multimodal language model},
  author={Driess, Danny and Xia, Fei and Sajjadi, Mehdi SM and Lynch, Corey and Chowdhery, Aakanksha and Ichter, Brian and Wahid, Ayzaan and Tompson, Jonathan and Vuong, Quan and Yu, Tianhe and others},
  booktitle={Proceedings of the 40th International Conference on Machine Learning},
  pages={8469--8488},
  year={2023}
}

@article{tang2025uad,
  title={UAD: Unsupervised Affordance Distillation for Generalization in Robotic Manipulation},
  author={Tang, Yihe and Huang, Wenlong and Wang, Yingke and Li, Chengshu and Yuan, Roy and Zhang, Ruohan and Wu, Jiajun and Fei-Fei, Li},
  journal={arXiv preprint arXiv:2506.09284},
  year={2025}
}

@article{huang2023voxposer,
  title={Voxposer: Composable 3d value maps for robotic manipulation with language models},
  author={Huang, Wenlong and Wang, Chen and Zhang, Ruohan and Li, Yunzhu and Wu, Jiajun and Fei-Fei, Li},
  journal={arXiv preprint arXiv:2307.05973},
  year={2023}
}

@article{huang2024rekep,
  title={Rekep: Spatio-temporal reasoning of relational keypoint constraints for robotic manipulation},
  author={Huang, Wenlong and Wang, Chen and Li, Yunzhu and Zhang, Ruohan and Fei-Fei, Li},
  journal={arXiv preprint arXiv:2409.01652},
  year={2024}
}

@inproceedings{yao2023react,
  title={React: Synergizing reasoning and acting in language models},
  author={Yao, Shunyu and Zhao, Jeffrey and Yu, Dian and Du, Nan and Shafran, Izhak and Narasimhan, Karthik and Cao, Yuan},
  booktitle={International Conference on Learning Representations (ICLR)},
  year={2023}
}

@article{team2025gemma,
  title={Gemma 3 technical report},
  author={Team, Gemma and Kamath, Aishwarya and Ferret, Johan and Pathak, Shreya and Vieillard, Nino and Merhej, Ramona and Perrin, Sarah and Matejovicova, Tatiana and Ram{\'e}, Alexandre and Rivi{\`e}re, Morgane and others},
  journal={arXiv preprint arXiv:2503.19786},
  year={2025}
}

@article{wu2025spatial,
  title={Spatial-mllm: Boosting mllm capabilities in visual-based spatial intelligence},
  author={Wu, Diankun and Liu, Fangfu and Hung, Yi-Hsin and Duan, Yueqi},
  journal={arXiv preprint arXiv:2505.23747},
  year={2025}
}

@misc{anthropic_claude4_2025,
  title        = {Claude 4 Sonnet System Card},
  author       = {{Anthropic}},
  year         = {2025},
  month        = May,
  url          = {https://www-cdn.anthropic.com/6be99a52cb68eb70eb9572b4cafad13df32ed995.pdf},
  note         = {Version 20250514, accessed 2025-06-23}
}

@misc{you2024multiviewequivarianceimproves3d,
      title={Multiview Equivariance Improves 3D Correspondence Understanding with Minimal Feature Finetuning}, 
      author={Yang You and Yixin Li and Congyue Deng and Yue Wang and Leonidas Guibas},
      year={2024},
      eprint={2411.19458},
      archivePrefix={arXiv},
      primaryClass={cs.CV},
      url={https://arxiv.org/abs/2411.19458}, 
}

@misc{zhang2024multiviewscenegraph,
      title={Multiview Scene Graph}, 
      author={Juexiao Zhang and Gao Zhu and Sihang Li and Xinhao Liu and Haorui Song and Xinran Tang and Chen Feng},
      year={2024},
      eprint={2410.11187},
      archivePrefix={arXiv},
      primaryClass={cs.CV},
      url={https://arxiv.org/abs/2410.11187}, 
}

@misc{fan2025vlm3rvisionlanguagemodelsaugmented,
      title={VLM-3R: Vision-Language Models Augmented with Instruction-Aligned 3D Reconstruction}, 
      author={Zhiwen Fan and Jian Zhang and Renjie Li and Junge Zhang and Runjin Chen and Hezhen Hu and Kevin Wang and Huaizhi Qu and Dilin Wang and Zhicheng Yan and Hongyu Xu and Justin Theiss and Tianlong Chen and Jiachen Li and Zhengzhong Tu and Zhangyang Wang and Rakesh Ranjan},
      year={2025},
      eprint={2505.20279},
      archivePrefix={arXiv},
      primaryClass={cs.CV},
      url={https://arxiv.org/abs/2505.20279}, 
}

@misc{zheng2025learningvideos3dworld,
      title={Learning from Videos for 3D World: Enhancing MLLMs with 3D Vision Geometry Priors}, 
      author={Duo Zheng and Shijia Huang and Yanyang Li and Liwei Wang},
      year={2025},
      eprint={2505.24625},
      archivePrefix={arXiv},
      primaryClass={cs.CV},
      url={https://arxiv.org/abs/2505.24625}, 
}

@misc{zhao2025embodiedrcollaborativeframeworkactivating,
      title={Embodied-R: Collaborative Framework for Activating Embodied Spatial Reasoning in Foundation Models via Reinforcement Learning}, 
      author={Baining Zhao and Ziyou Wang and Jianjie Fang and Chen Gao and Fanhang Man and Jinqiang Cui and Xin Wang and Xinlei Chen and Yong Li and Wenwu Zhu},
      year={2025},
      eprint={2504.12680},
      archivePrefix={arXiv},
      primaryClass={cs.AI},
      url={https://arxiv.org/abs/2504.12680}, 
}

@misc{xu2025multispatialmllmmultiframespatialunderstanding,
      title={Multi-SpatialMLLM: Multi-Frame Spatial Understanding with Multi-Modal Large Language Models}, 
      author={Runsen Xu and Weiyao Wang and Hao Tang and Xingyu Chen and Xiaodong Wang and Fu-Jen Chu and Dahua Lin and Matt Feiszli and Kevin J. Liang},
      year={2025},
      eprint={2505.17015},
      archivePrefix={arXiv},
      primaryClass={cs.CV},
      url={https://arxiv.org/abs/2505.17015}, 
}

@article{sargent2023zeronvs,
  title={ZeroNVS: Zero-Shot Novel View Synthesis from a Single Real Image},
  author={Sargent, Kyle and Li, Zizhang and Shah, Tanmay and Herrmann, Charles and Yu, Hong-Xing and Zhang, Yunzhi and Chan, Eric Ryan and Lagun, Dmitry and Fei-Fei, Li and Sun, Deqing and Wu, Jiajun},
  journal={arXiv:2310.17994},
  year={2023}
}

@inproceedings{zhang2024viewcrafter,
  title={ViewCrafter: Taming Video Diffusion Models for High-fidelity Novel View Synthesis},
  author={Zhang, Yuxuan and Yang, Yifan and Zhang, Jing and Wang, Yifang and Zhang, Yijun and Yang, Ming-Hsuan},
  booktitle={ECCV},
  year={2024}
}

@inproceedings{sun2018multiview,
  title={Multi-view to Novel view: Synthesizing Novel Views with Self-Learned Confidence},
  author={Sun, Shao-Hua and Huh, Minyoung and Liao, Yuan-Hong and Zhang, Ning and Lim, Joseph J},
  booktitle={ECCV},
  year={2018}
}

@inproceedings{wang2025vggt,
  title     = {VGGT: Visual Geometry Grounded Transformer for Universal 3D Reconstruction},
  author    = {Wang, Jianyuan and others},
  booktitle = {CVPR},
  year      = {2025},
}

@inproceedings{liu2025citygaussianv2,  
  title     = {CityGaussianV2: Efficient and Geometrically Accurate Reconstruction for Large-Scale Scenes},  
  author    = {Liu, Deku and Zhang, Yihan and Chen, Zhe and others},  
  booktitle = {ICLR},  
  year      = {2025},  
  url       = {https://arxiv.org/pdf/2411.00771},  
 
}

@inproceedings{fu2025maskgaussian,  
  title     = {MaskGaussian: Differentiable Mask Pruning for Efficient 3D Gaussian Rendering},  
  author    = {Fu, Chuanyu and Chen, Guanying and others},  
  booktitle = {CVPR},  
  year      = {2025},  
  url       = {https://arxiv.org/abs/2506.02751},  

}

@inproceedings{qu2025drag,  
  title     = {Drag Your Gaussian: Effective Drag-Based Editing with Score Distillation for 3D Gaussian Splatting},  
  author    = {Qu, Yansong and Wang, Jie and others},  
  booktitle = {SIGGRAPH Asia},  
  year      = {2025},  
  url       = {https://arxiv.org/pdf/2501.18672},  
 
}
}

\appendix
\newpage
\renewcommand{\partname}{}
\part{Appendix} %
\parttoc %
\clearpage

\section{The Use of Large Language Models}
We used large language models (LLMs), including Google's Gemini~2.5~Pro and OpenAI's GPT-5, as auxiliary tools to assist with writing, editing, and conducting the literature review for this manuscript. All content was critically reviewed, fact-checked, and revised by the human authors to ensure its scientific validity and originality. The authors are fully responsible for all statements and conclusions presented in this paper. Specifically, we use LLMs for polishing our wording and writing, and we use LLMs to retrieve several related works.

\section{\name Benchmark}\label{app_sec:mindcube}
\subsection{Details for Data Collection and Annotation} \label{app_sec:mindcube_collection}
\setcounter{table}{0}
\setcounter{figure}{0}

\begin{figure*}[ht]
  \centering
  \includegraphics[width=1\linewidth]{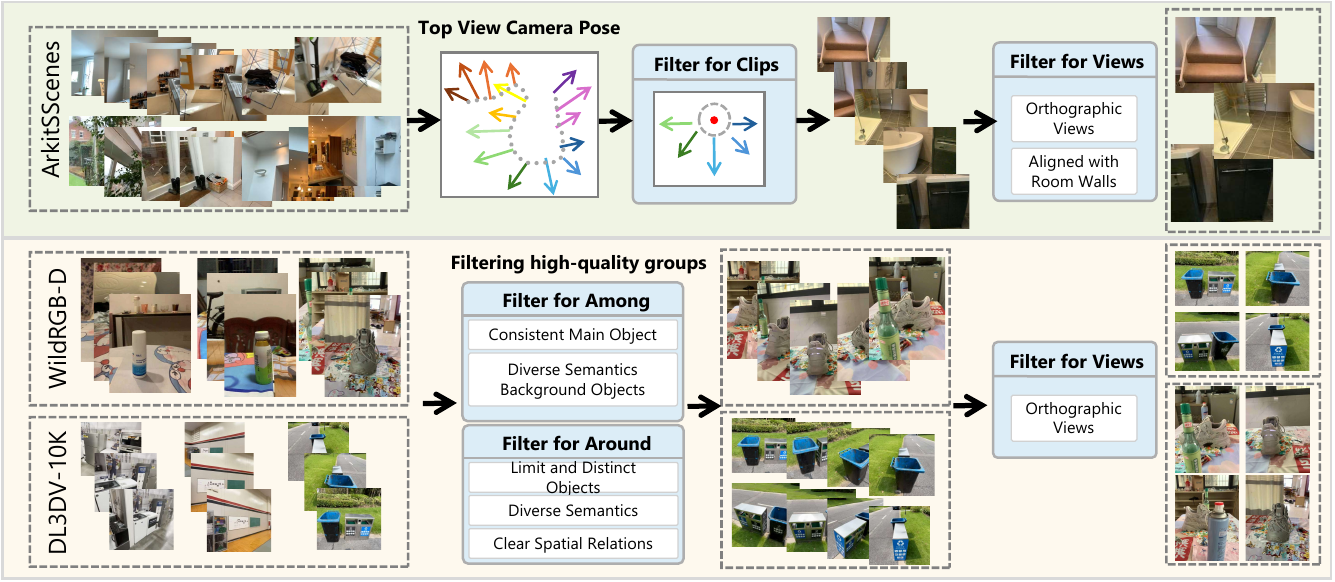}
  \caption{\name Bench construction pipeline.}
  \label{fig:pipeline}
\end{figure*}

\textbf{Image Collection and Selection.} Our \name benchmark comprises \numimg images (2,302 indoor/outdoor images from publicly released dataset and 400 self-collected images), where we implement a comprehensive image selection methodology encompassing four distinct view dynamics, incorporating various data sources and processing procedures, as shown in Fig.\ref{fig:pipeline}.

 For rotation view dynamics, we implement a three-stage filtering strategy to extract meaningful camera trajectories and key frames from ArkitScenes~\cite{baruch1arkitscenes} dataset.

In the first stage, we analyze the top-down view of camera poses within each scene to identify two types of trajectories: linear paths and small rotational arcs. A linear trajectory is characterized by consistently oriented cameras exhibiting significant displacement perpendicular to their viewing direction. A rotational arc trajectory is identified when three to four camera positions demonstrate approximately 90-degree relative orientation changes while being distributed along an approximate circular arc.
\begin{figure}[htbp]
    \centering
    \begin{subfigure}[ht]{0.4\textwidth}
        \centering
        \includegraphics[width=\textwidth]{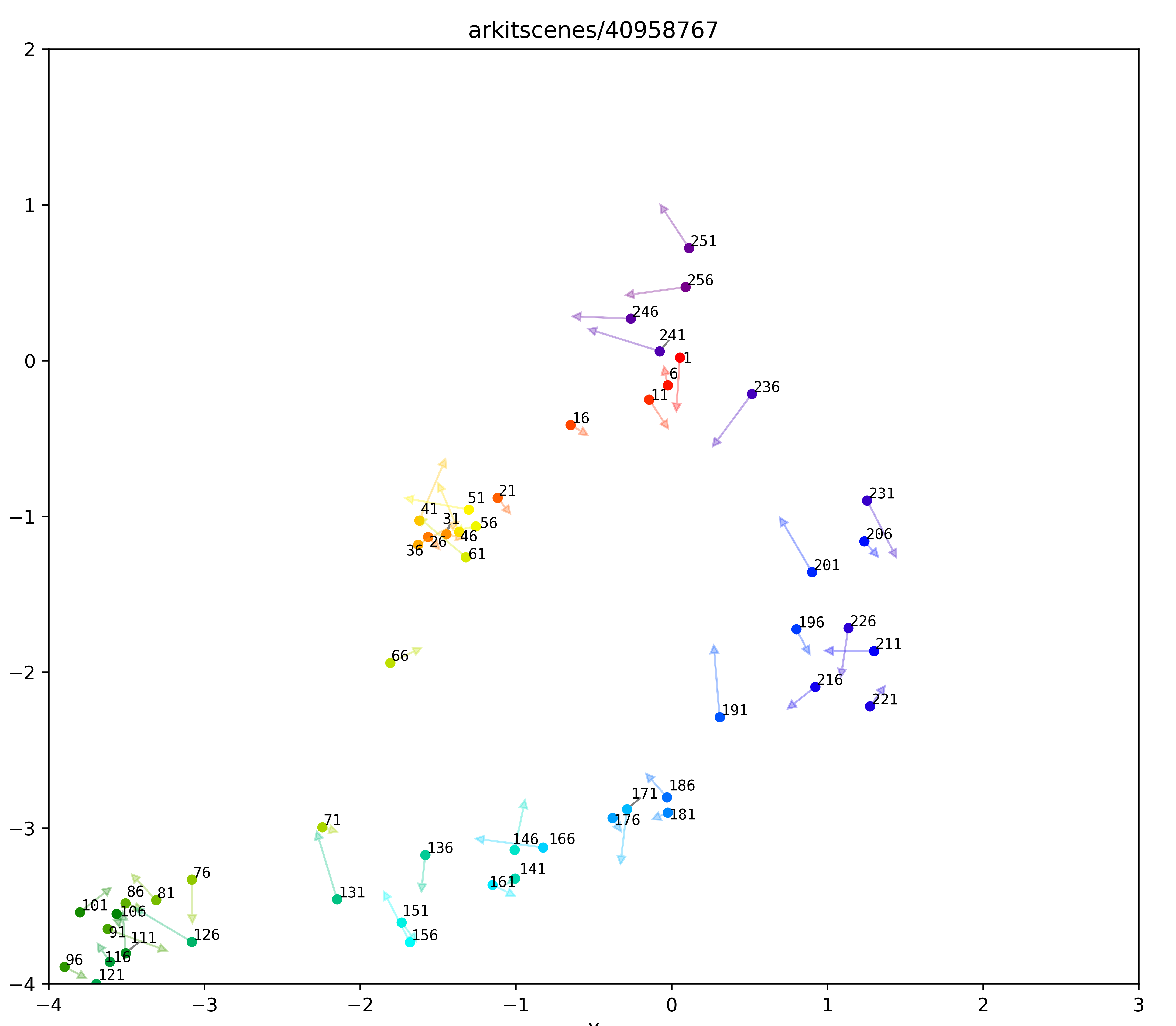}
        \caption{camera pose 1}
        \label{fig:sub1}
    \end{subfigure}
    \hfill %
    \begin{subfigure}[ht]{0.55\textwidth}
        \centering
        \includegraphics[width=\textwidth]{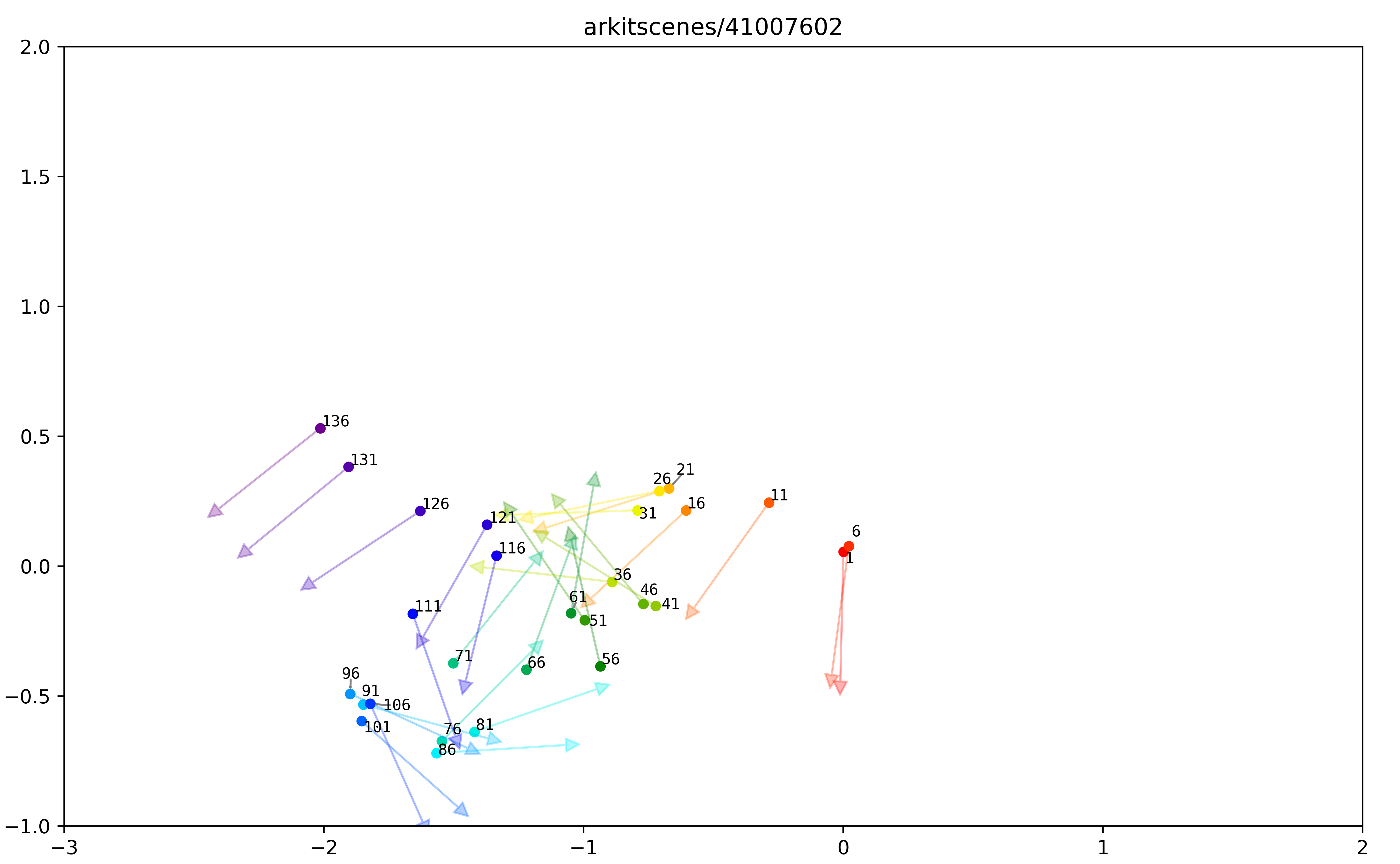}
        \caption{camera pose 2}
        \label{fig:sub2}
    \end{subfigure}
    \caption{Examples of camera poses in ArkitScenes}
    \label{fig:combined}
\end{figure}
The second stage focuses on selecting two critical frames from the previously identified translation segments. The selection criteria mandate that: (1) the camera movement direction must be parallel to the object arrangement direction, (2) this movement should be aligned with the horizontal axis, (3) the first frame should only capture objects A and B, while the second frame should only capture objects B and C, and (4) both frames must be free from motion blur and exhibit clear object visibility.

The third stage processes the rotation segments to extract three or four key frames. These frames must satisfy several conditions: (1) the camera positions should appear to originate from a stationary rotating camera, even if slight circular movement exists, (2) the camera orientations should align with standard cardinal directions (approximately 90 degrees apart), and (3) each frame should contain no more than three semantically distinct primary objects that occupy over 50\% of the frame area relative to the background.

For among view dynamics, image groups are manually selected from DL3DV-10K\cite{ling2023dl3dv10klargescalescenedataset} and WildRGB-D\cite{xia2024rgbd} datasets. We employ a single-stage selection process to identify four key frames representing cardinal viewpoints (front, left, right, and back) from 360-degree scene captures. The selection criteria are: (1) camera orientations must align with standard directions, ensuring that the central object, its background objects, and the camera's line of sight are collinear and parallel or perpendicular to standard scene elements such as tables or walls, (2) we reject sets where three or more frames share identical semantic background information, and (3) we discard sets where three or more frames have severely occluded background objects that cannot be reconstructed from information in the other frames.

For around view dynamics, image groups are manually curated from the DL3DV-10K\cite{ling2023dl3dv10klargescalescenedataset} dataset and assigned sequential identifiers. The front view (designated as view 1) must provide clear visibility of all relevant information. This view is established as the reference point for subsequent views in the sequence.

This structured approach to image selection and processing yields a rich dataset that supports subsequent model training and testing procedures. The methodology ensures comprehensive coverage of spatial relationships, occlusion states, and view-dependent object characteristics across multiple viewing scenarios. 

\noindent\textbf{Data Annotation.} After collecting and filtering the images, we follow a two-phase paradigm for annotation:
We establish a systematic image annotation protocol to ensure data consistency and accuracy. The annotation framework encompasses four key dimensions: spatial relationship identification, object grouping rules, semantic orientation determination, and occlusion level assessment. We provide a pdf of the annotation interface in the supplementary material.

Regarding spatial relationship identification, annotators are required to identify primary object entities within images and determine their spatial relationships. These relationships are primarily categorized into two types: front-back relationships typically involving two primary objects, with priority given to objects directly behind as key entities; and left-right relationships encompassing two to four primary objects, where adjacent objects with front-back relationships can be considered as a unified entity.

To enhance annotation efficiency and semantic completeness, this study introduces object grouping rules. Multiple objects can be annotated as a unified entity when they collectively form clear spatial relationships with other primary objects. Each object may include attribute descriptors (e.g., color, material) to enhance semantic expression. Combined object entities must maintain distinct spatial relationships with other primary objects.

For objects with definitive semantic fronts, the following information must be recorded: the object's inherent semantic front, the object's orientation relative to the current viewpoint (aligned, reversed, leftward, rightward, etc.), and the object's actual projected direction within the scene.

Occlusion levels are evaluated using a four-tier classification system: complete occlusion where the object is entirely invisible from the current viewpoint; major occlusion where primary object features are difficult to identify; minor occlusion where primary object features remain identifiable; and no occlusion where the object is fully visible. For cases of complete occlusion, the annotation system provides multi-view scene images, ensuring object visibility in at least one viewpoint to support subsequent cross-view question-answering system training. 

This annotation protocol provides a structured semantic foundation for subsequent automated question-answer pair generation while ensuring data quality and consistency. Through this standardized annotation process, we effectively capture key information including spatial relationships, compositional features, semantic orientations, and occlusion states of objects within scenes.  \\
\begin{figure*}[ht]
  \centering
  \includegraphics[width=1\linewidth]{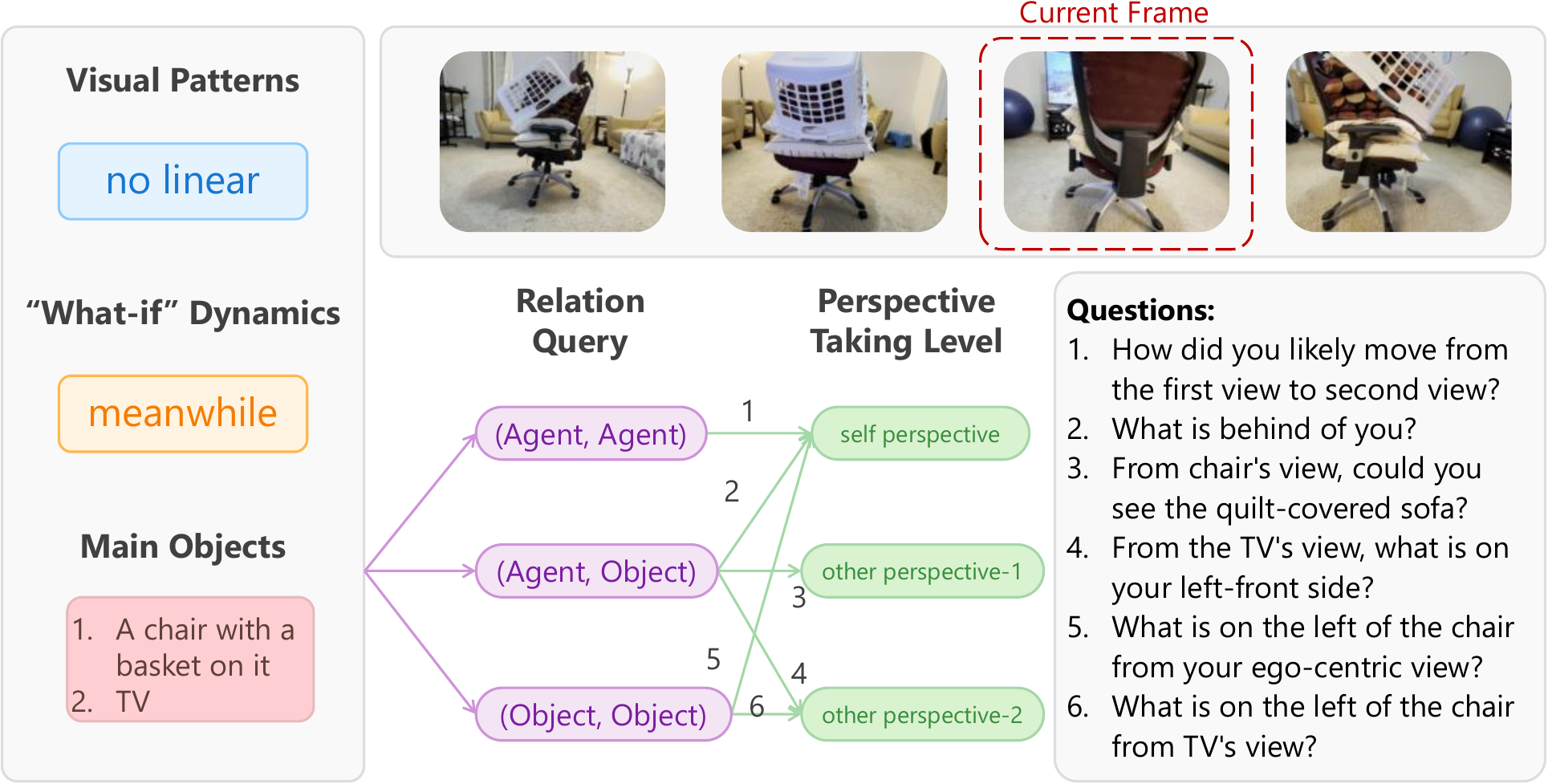}
  \caption{Example of different question-related label combinations to generate QA pairs.}
  \label{fig:data_gen}
\end{figure*}
\textbf{Examples for automatic QA generation pipeline.} 
Our automatic QA generation pipeline generates different types of questions using combinations of labels. Each question's label combination is encoded in its ID (e.g., "among\_group001\_q1\_1\_1"), while the original object and label information is preserved in the meta\_info field to track the context of question generation.

\subsection{Details of our \name Benchmark} 
\subsubsection{Three kinds of invisibility settings}

\textbf{Rotation.} In this setting, our camera remains stationary while rotating in place, capturing 2 to 4 orthogonal views. In each view, a central object remains visible in the foreground, while all views maintain equal importance in the spatial representation.

We evaluate models' understanding of spatial invisibility by asking questions such as 'When positioned at a particular viewpoint, what should be to your left or right (given that each view only reveals what's directly ahead)?' or 'After rotating a quarter or half turn, what objects would be in front of you, to your left, behind you, or to your right?' We expect models to construct a comprehensive spatial understanding by leveraging the \textbf{sequential nature of the views and consistent spatial cues} across images (such as lighting direction), thereby demonstrating their ability to reason about the complete environment despite only having access to partial visual information from each viewpoint.\\

\noindent\textbf{Around.} In this setting, we leverage \textbf{occlusion} phenomena to force MLLMs beyond simple 2D spatial recognition. When viewing objects from different angles, some objects become partially or fully hidden, requiring models to:
\begin{itemize}
    \item Maintain object permanence despite partial visibility
    \item Transform lateral relationships (left-right) from frontal views into depth relationships (front-back) for side views
    \item Integrate spatial information across multiple viewpoints to form a coherent 3D understanding
\end{itemize}

This approach prevents models from relying solely on direct visual cues and instead necessitates true 3D spatial reasoning by combining information from multiple perspectives.\\

\noindent\textbf{Among.} In this setting, the camera rotates around a central object, positioned between this central object and several surrounding objects. Four orthogonal views are captured, with each view showing the central object positioned in front of one of the surrounding objects.

This setup creates interesting visibility constraints across different perspectives. For instance, a surrounding object visible in one view may be invisible in another view because of the constraints imposed by the camera's field of view. Through establishing consistency relationships between these views, we can infer the relative positions of objects not directly visible from certain perspectives. When an object is not visible from a particular viewpoint, consistency and spatial reasoning can determine its position relative to the central object.

All views hold equal status in this framework, allowing for bidirectional establishment of invisibility relationships. This creates a coherent spatial reasoning system where information from each perspective contributes to a complete understanding of the three-dimensional arrangement, even when direct visual confirmation is unavailable from certain angles.

\subsubsection{Label taxonomy} \label{app_sec:taxonomy_def}
We use image related labels for better analysis and question related labels for automatic QA generation with different label combinations.

\noindent\textbf{Visual Patterns.} In our taxonomy of spatial configurations, we classify visual patternss into distinct categories based on their geometric relationships. Linear arrangements refer to configurations where objects are positioned along a single axis, forming a collinear pattern. Non-linear arrangements, conversely, are characterized by objects positioned such that the connecting lines between adjacent pairs form 90-degree angles, creating rectilinear patterns. This binary classification serves as a fundamental attribute in our spatial relationship labeling scheme, enabling precise description and analysis of scene compositions across various domains.

\noindent\textbf{ “What if” Dynamics.} “What if” Dynamics refers to the model's capability to comprehend and reason about dynamic perspective changes occurring within images or posed questions. We conceptualize viewpoint transitions as combinations of translation and rotation operations, resulting in four distinct categories:
\begin{itemize}

    \item Pure Translation: Cases where the viewpoint undergoes only translational movement without rotational change.
    \item Pure Rotation: Scenarios involving rotational transformation of the viewpoint while maintaining its positional coordinates.
    \item Simultaneous Translation-Rotation(Meanwhile): Instances where both translational and rotational operations occur concurrently.
    \item Sequential Translation-Rotation(Sequence): Cases where translation and rotation occur in sequence rather than simultaneously. Notably, in our dataset, this category is uniquely represented through textual descriptions in the questions rather than through explicit visual transformations.
\end{itemize}

The first three categories of “What if” dynamics are visually demonstrated through changes in view representation, while the sequential category requires models to interpret text-based descriptions of perspective changes. This taxonomy provides a systematic framework for evaluating spatial reasoning capabilities across diverse viewpoint transformation scenarios.

\noindent\textbf{Relation  Query.} We define three distinct categories of relation  queries that capture the fundamental nature of spatial reasoning tasks:
\begin{itemize}

    \item Agent-Agent: This pattern involves self-referential spatial positioning, where the observer must evaluate and potentially adjust their own position in space. It requires egocentric spatial reasoning and self-awareness of one's location relative to environmental constraints.

    \item Agent-Object: This pattern focuses on determining the orientation of an observed object relative to the observer's position. Unlike the P-P pattern, the emphasis here is on object perception rather than self-positioning, requiring the observer to make judgments about external entities while maintaining awareness of their own reference frame.

    \item Object-Object: This pattern involves reasoning about the spatial relationship between two discrete objects in the environment, independent of the observer's position. This allocentric spatial reasoning requires understanding relative positioning, distance, and orientation between entities without necessarily using oneself as a reference point.
\end{itemize}

These categorizations provide a structured approach to analyzing the cognitive demands of different spatial reasoning tasks and can inform both the design of spatial question answering systems and the evaluation of human spatial cognition abilities.

\noindent\textbf{Perspective Taking.} 
We propose a label called "Perspective Taking" that categorizes the complexity of viewpoint projection. This label distinguishes between three increasingly sophisticated levels of perspective reasoning:
\begin{itemize}

    \item Self Perspective: Reasoning based on the current camera view or the observer's own viewpoint. This represents the baseline where no perspective shift is required.

    \item Other's Perspective Taking-1: The ability to determine visibility relationships from another agent's viewpoint. This involves understanding what objects are visible or occluded from a different viewpoint (e.g., determining whether a specific object is within the field of view of another camera). The another agent's viewpoint is usually determined by an object with a clear orientation in the image.

    \item Other's Perspective Taking-2: The ability to understand how spatial relationships transform when viewed from another agent's perspective. This more advanced capability requires mental rotation and spatial transformation to reason about relative positions (e.g., determining whether, from another viewpoint, object X appears to be positioned behind object Y).
\end{itemize}

This classification aligns with developmental psychology research on perspective-taking abilities, where Level-1 perspective taking typically develops earlier than the more cognitively demanding Level-2 perspective taking.

We provide performance across different categories and labels in Table\ref{tab:part1} and \ref{tab:part2}. Upon detailed analysis of model performance across various capabilities, certain trends emerge. The O-O (Object-Object) task within Relation Pattern also demonstrates generally lower scores across the board, suggesting it is a less tractable problem for current models. Notably, InternVL2-8B struggles with the sequence task, exhibiting the lowest score among all evaluated models in that category.

Regarding model stability, Mantis(SigLip) demonstrates robust performance in both Object Arrangement and Relation Pattern sections, indicating a consistent capability in these spatial reasoning tasks. Similarly, Qwen2.5-VL-7B-Instruct maintains relatively stable performance within Viewpoint Dynamics. In contrast, InternVL2-8B shows a broader instability, with consistently lower overall scores and considerable performance fluctuations across different sub-categories, highlighting areas for further improvement in its generalizability and robustness.
\begin{table}[h!]
\centering
\caption{Performance of VLMs on \name across categories.(Part 1)}
\label{tab:part1}
\small
\begin{tabular}{@{}l S[table-format=2.2] *{6}{S[table-format=2.2]}@{}}
\toprule
\textbf{Model} & \textbf{Overall} & 
\multicolumn{2}{c}{\textbf{Object Arrangement}} & 
\multicolumn{3}{c}{\textbf{Perspective Taking}} \\
\cmidrule(lr){3-4} \cmidrule(l){5-7}
& & \textbf{Linear} & \textbf{Perp.} & \textbf{Self} & \textbf{Level1} & \textbf{Level2} \\
\midrule
LLaVA-Video-7B-Qwen2 & 41.96 & 30.12 & 43.11 & 42.19 & 60.76 & 33.80 \\
Mantis(SigLip) & 41.04 & \textbf{50.99} & 40.08 & 41.20 & 54.43 & 35.41 \\
GPT-4o & 38.81 & 29.16 & 39.75 & 39.07 & 46.20 & 31.86 \\
Qwen2.5-VL-3B-Instruct & 33.21 & 30.34 & 33.49 & 32.96 & 46.84 & 36.28 \\
LongVA-7B & 29.46 & 24.88 & 29.91 & 28.81 & 51.90 & \textbf{39.83} \\
Qwen2.5-VL-7B-Instruct & 29.26 & 21.35 & 30.02 & 28.77 & 46.84 & 36.81 \\
deepseek-vl2-small & \textbf{47.62} & 26.91 & \textbf{49.63} & \textbf{48.32} & 56.33 & 31.11 \\
Robobrain & 37.38 & 29.53 & 38.14 & 37.56 & 55.06 & 30.57 \\
Claude-sonnet-4 & 44.75 & 47.62 & 44.48 & {45.32} & 49.38 & 31.74 \\
Space-Mantis & 22.82 & 29.32 & 22.19 & 22.15 & 45.57 & 33.48 \\
InternVL2-8B & 18.68 & 13.11 & 19.22 & 17.89 & \textbf{64.56} & 27.99 \\
Space-Qwen & 33.28 & 26.32 & 33.95 & 33.06 & 46.84 & 35.63 \\
LLaVA-Onevision-7B& {47.43} & 44.09 & 47.75 & 48.04 & 51.27 & 33.48 \\
Spatial-MLLM & 32.06 & 20.92 & 33.13 & 31.79 & 46.84 & 35.20 \\
mPLUG-Owl3-7B & {44.85} & 26.91 & 46.59 & 45.15 & 60.13 & 35.74 \\
\bottomrule
\end{tabular}
\end{table}

\begin{table}[h!]
\centering
\caption{Performance of VLMs on \name across categories.(Part 2)}
\label{tab:part2}
\small
\begin{tabular}{@{}l *{6}{S[table-format=2.2]}@{}}
\toprule
\textbf{Model} & 
\multicolumn{3}{c}{\textbf{Relation Pattern}} & 
\multicolumn{3}{c}{\textbf{Viewpoint Dynamics}} \\
\cmidrule(lr){2-4} \cmidrule(l){5-7}
& \textbf{A-A} & \textbf{A-O} & \textbf{O-O} & 
\textbf{Rotation} & \textbf{Meanwhile} & \textbf{Sequence} \\
\midrule
LLaVA-Video-7B-Qwen2 & 36.22 & 57.61 & 26.67 & 35.71 & 30.12 & 73.45 \\
Mantis(SigLip) & 23.78 & 64.16 & 25.24 & 37.65 & 24.99 & 82.74 \\
GPT-4o & \textbf{49.30} & 48.38 & 16.70 & 32.65 & 31.09 & 59.73 \\
Qwen2.5-VL-3B-Instruct & 37.85 & 37.51 & 20.65 & 37.37 & 27.88 & 46.05 \\
LongVA-7B & 19.72 & 35.49 & 25.58 & 35.89 & 24.67 & 40.50 \\
Qwen2.5-VL-7B-Instruct & 31.41 & 34.67 & 15.63 & 38.76 & 22.87 & 43.76 \\
deepseek-vl2-small & 43.98 & \textbf{68.27} & 25.33 & 37.00 & 32.97 & \textbf{87.13} \\
Robobrain & 30.94 & 49.18 & 27.37 & 35.80 & 28.79 & 59.66 \\
Claude-sonnet-4 & 41.78 & 67.25 & 15.85 & \textbf{48.42} & \textbf{34.76} & 69.53 \\
Space-Mantis & 28.18 & 17.03 & 20.89 & 37.65 & 24.98 & 14.46 \\
InternVL2-8B & 15.67 & 12.47 & 24.58 & 36.45 & 21.78 & 7.36 \\
Space-Qwen & 31.59 & 38.14 & 26.13 & 38.02 & 28.51 & 44.58 \\
LLaVA-Onevision-7B & 42.28 & 65.87 & \textbf{29.79} & 36.45 & 33.80 & 84.38 \\
Spatial-MLLM & 27.72 & 37.75 & 25.80 & 38.39 & 26.84 & 44.19 \\
mPLUG-Owl3-7B & 47.80 & 62.29 & 18.83 & 37.84 & 31.02 & 81.55 \\
\bottomrule
\end{tabular}
\end{table}

\subsection{Examples}
\begin{figure*}[ht]
  \centering
  \includegraphics[width=1\linewidth]{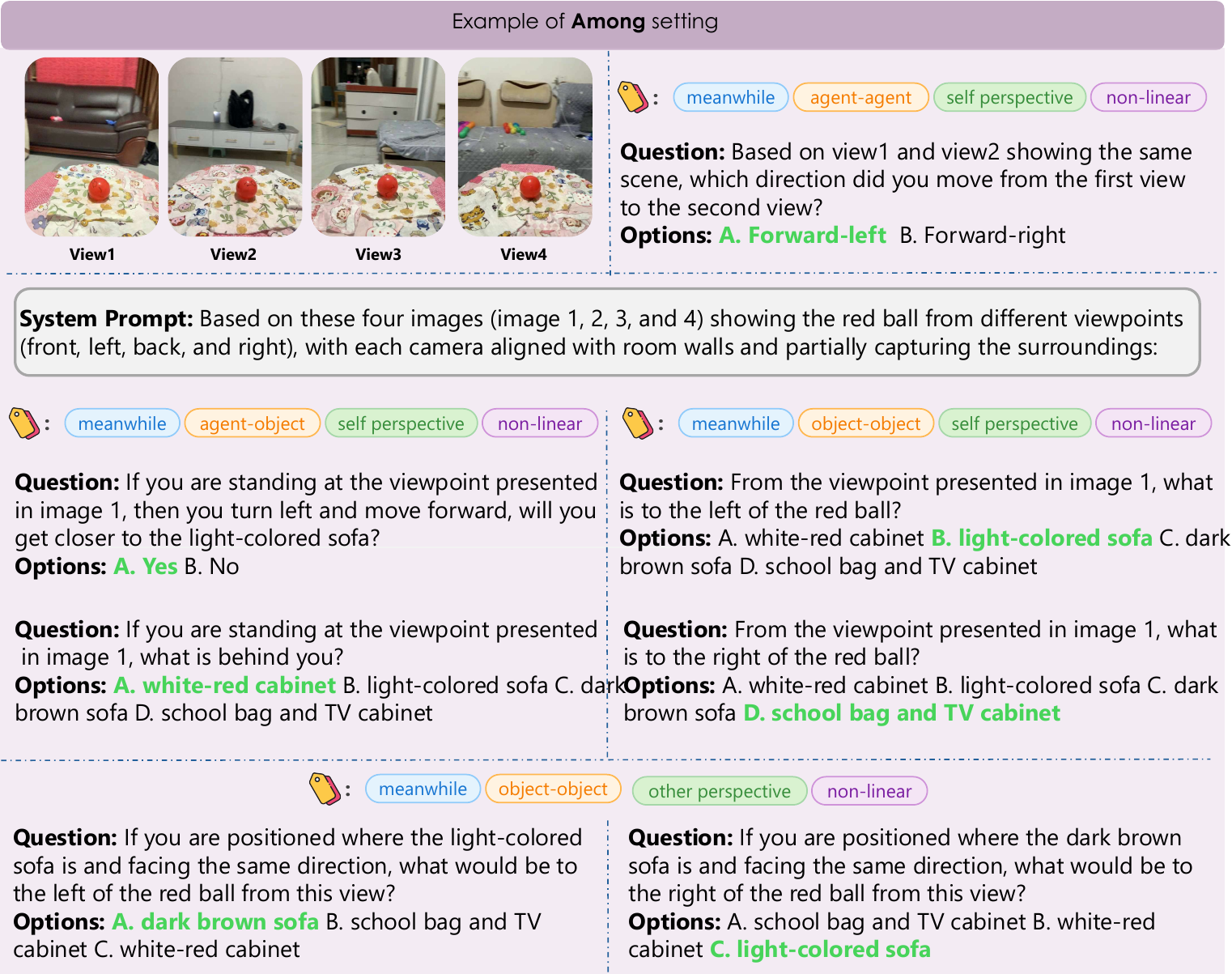}
  \caption{Example of among setting.}
  \label{fig:among_example}
\end{figure*}

\begin{figure*}[ht]
  \centering
  \includegraphics[width=1\linewidth]{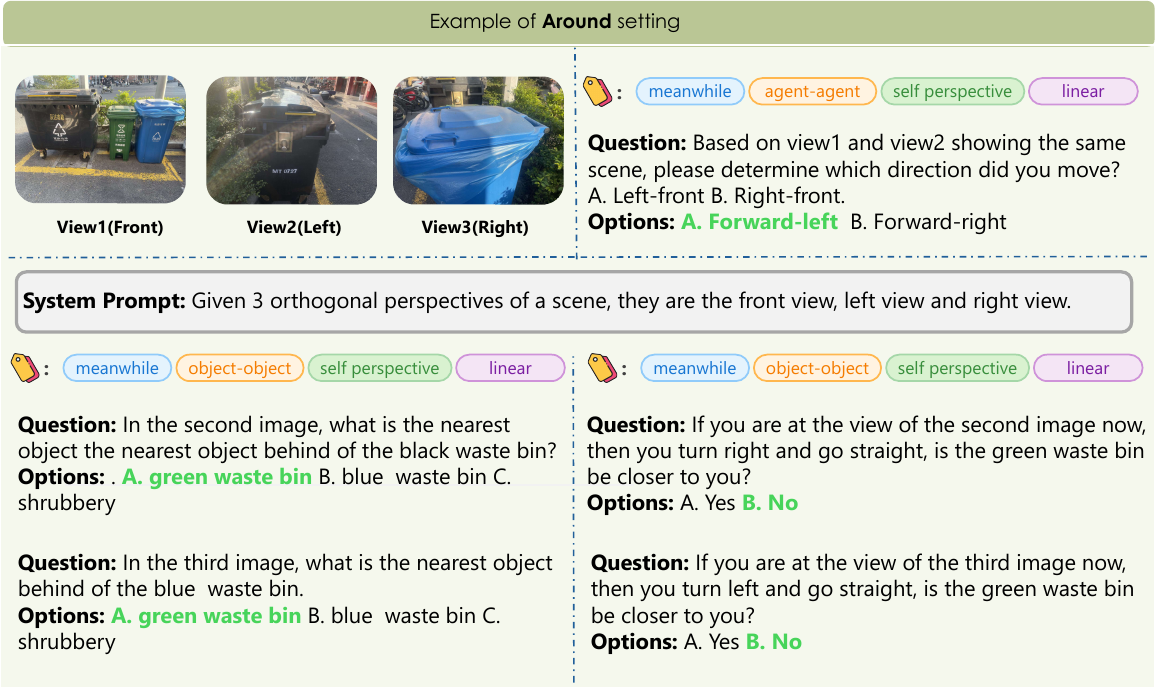}
  \caption{Example-1 of around setting.}
  \label{fig:around_example1}
\end{figure*}

\begin{figure*}[ht]
  \centering
  \includegraphics[width=1\linewidth]{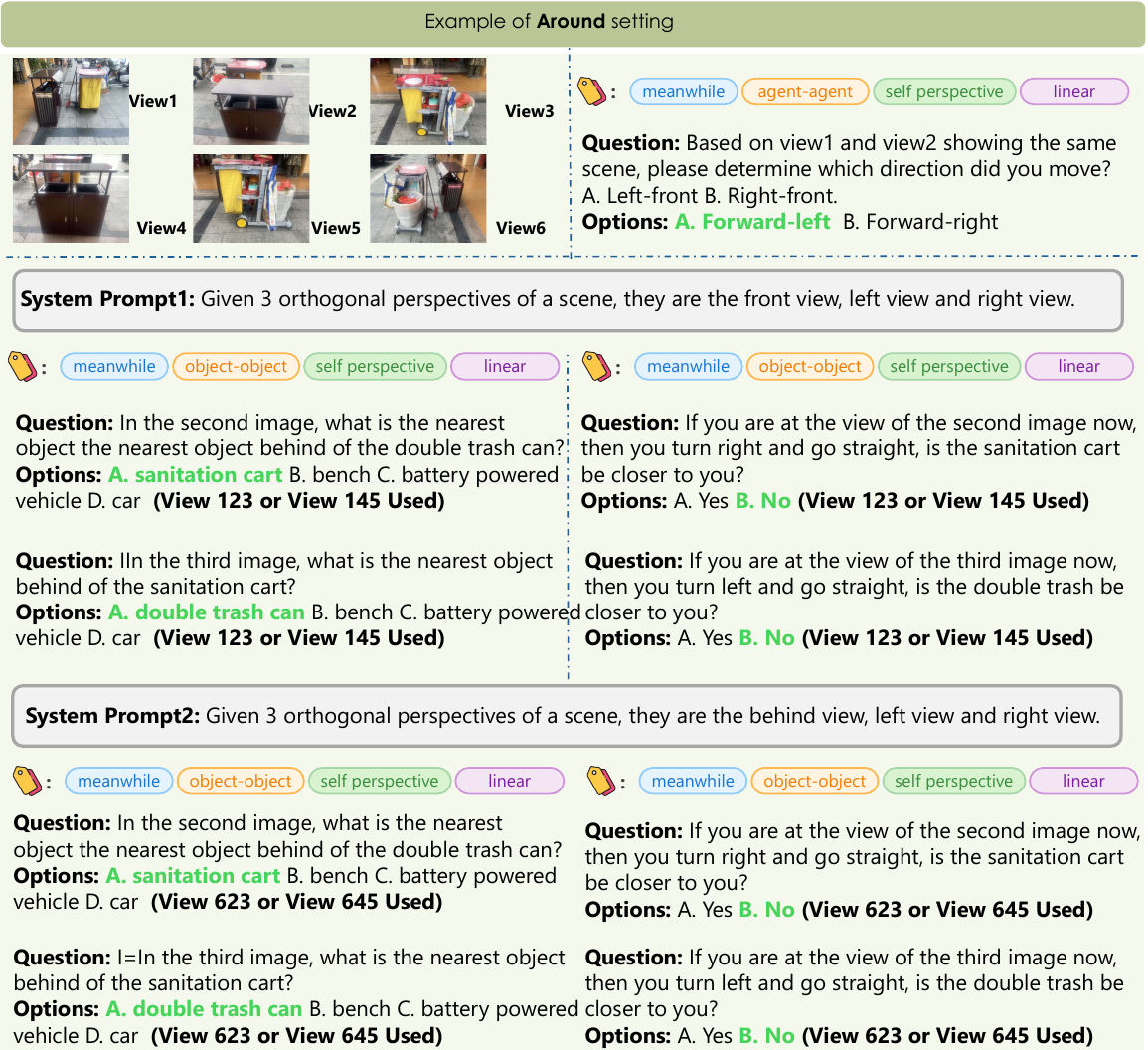}
  \caption{Example-2 of around setting.}
  \label{fig:around_example2}
\end{figure*}

We show some examples in Figure\ref{fig:around_example1}, \ref{fig:around_example2} and \ref{fig:among_example}.
\label{app_sec:mindcube_example}

\section{Evaluation on \name}\label{app_sec:mindcube_eval}
\subsection{Prompt Templates for Evaluation}
\begin{tcolorbox}[colback=black!5!white,colframe=black!75!white]

\textbf{Evaluation Prompt Prefix} \\
Based on these images, answer the question based on this rule: You only need to provide *ONE* correct answer selecting from the options listed below. For example, if you think the correct answer is `A. above' from ` A. above B. under C. front D. behind.', your response should only be `A. above'.

The Question is: 
\end{tcolorbox}
\subsection{Details in text only evaluation}

In the text-only evaluation, we replace the original image input with corresponding textual descriptions and assess the performance of models based on these descriptions. The purpose of this evaluation is to highlight how much information may be lost or distorted when the visual input is substituted with text-based representations, and to demonstrate the crucial role of visual data in the models’ performance.

We used two types of captions: \textbf{brief} and \textbf{dense}. The brief captions provide a concise overview of the image, while the dense captions offer a more detailed description with a focus on the spatial relationships between objects. Additionally, the models are evaluated using textual descriptions (text-only evaluation) based on these captions, with no access to the actual images.

\begin{tcolorbox}[colback=black!5!white,colframe=black!75!white]
\textbf{Prompt for Brief Captioning} \\
Describe this image briefly.
\end{tcolorbox}

\begin{tcolorbox}[colback=black!5!white,colframe=black!75!white]

\textbf{Prompt for Dense Captioning} \\
Describe this image in detail, specifially focusing on the spatial relationship between objects.
\end{tcolorbox}

\begin{tcolorbox}[colback=black!5!white,colframe=black!75!white]

\textbf{Text-only evaluation Prompt Prefix} \\
You need to gather information about each image based on the descriptions I provide below, and answer the given questions using those textual descriptions, without directly viewing the images.\\

Image 1: \textless Caption 1\textgreater  

...

Image N: \textless Caption N\textgreater  
\end{tcolorbox}

As shown in the Table~\ref{appendix: text-only eval}, all three models exhibit a noticeable performance decline when replacing the original image input with its corresponding text-based description. Specifically, the brief captions cause the most significant performance drop. For instance, RoboBrain-8B experiences a 7.83\% decrease with the brief captions, and LLaVA-OneVision-7B drops by 12.91\% in the same condition. Even when using dense captions, which offer more detail, there is still a performance reduction, although the decrease is slightly less pronounced compared to brief captions. In conclusion, while textual descriptions can convey some information, they fail to capture the richness and intricacies of visual data, leading to a marked reduction in performance across all models.

\begin{table}[h]
\centering
\caption{Text-only (T) evaluation vs. original evaluation with image inputs (I). The results highlight a significant performance drop when the original image input is replaced with the corresponding text-based caption, particularly with the brief captions. In all cases, model performance decreases notably, underscoring that our benchmark is \textit{vision-centric}.}
\label{appendix: text-only eval}
\begin{tabular}{lccc}
\toprule
Model & Brief (T) & Dense (T) & Original (I)\\ \midrule
RoboBrain-8B     
& 33.92\% \textcolor{green!60}{\scriptsize$\downarrow$7.83\%}
& 35.58\% \textcolor{green!60}{\scriptsize$\downarrow$6.17\%} 
& 41.75\% \\
LLaVA-OneVision-7B 
& 34.17\% \textcolor{green!60}{\scriptsize$\downarrow$12.91\%} 
& 35.92\% \textcolor{green!60}{\scriptsize$\downarrow$11.16\%} 
& 47.08\% \\
Qwen2.5-VL-7B-Instruct
& 27.00\% \textcolor{green!60}{\scriptsize$\downarrow$5.33\%} 
& 28.75\% \textcolor{green!60}{\scriptsize$\downarrow$3.58\%} 
& 32.33\% \\
\bottomrule
\end{tabular}
\end{table}
\subsection{Human Evaluation}\label{app_sec:human_eval}
We use our Tiny Benchmark— encompassing all task categories for evaluation by 5 human annotators, each
 of whom independently answers every question. Here is the results\ref{tab:human_performance}.
 
\begin{table}[h]
\centering
\caption{Comparison of Human and GPT-4 Performance (\%)}
\begin{tabular}{lcccc}
    \hline
    \textbf{Model/Annotator} & \textbf{GPT4-o} & \textbf{Human-max} & \textbf{Human-min} & \textbf{Human-avg} \\ 
    \hline
    Accuracy & 36.54 & 94.77 & 94.20 & 94.55 \\
    \hline
\end{tabular}
\label{tab:human_performance}
\end{table}

This observation demonstrates the disparity in spatial reasoning capabilities between humans and state-of-the-art multimodal large language models, where humans exhibit superior performance in solving spatial problems that remain challenging for advanced AI systems.
\subsection{Evaluation Setup}
To comprehensively evaluate model performance, we conducted experiments on a diverse suite of models. This suite includes models with native multi-image reasoning capabilities (e.g., LLaVA-Onevision~\citep{onevision}, LLaVA-Video~\citep{llava-video}, mPLUG-Owl3~\citep{mplug-owl3}, InternVL2.5~\citep{chen2024internvl}, QwenVL2.5~\citep{Qwen2.5-VL}, LongVA~\citep{longva}, DeepSeek-VL2~\citep{lu2024deepseekvlrealworldvisionlanguageunderstanding}), Gemma3~\cite{team2025gemma}, models fine-tuned on interleaved image-text data (e.g., Mantis~\citep{mantis}), leading proprietary APIs (e.g., GPT-5, Claude-4-Sonnet), and models specifically fine-tuned for spatial reasoning tasks (e.g., RoboBrain~\citep{ji2025robobrain}, Space-Mantis~\citep{space-mantis}, Space-Qwen~\citep{space-mantis}, and Spatial-MLLM~\cite{wu2025spatial}).

\subsection{Analysis in settings} \label{app_sec:setting_analysis}

\subsubsection{Around} 
\begin{figure*}
  \centering
  \includegraphics[width=0.9\linewidth]{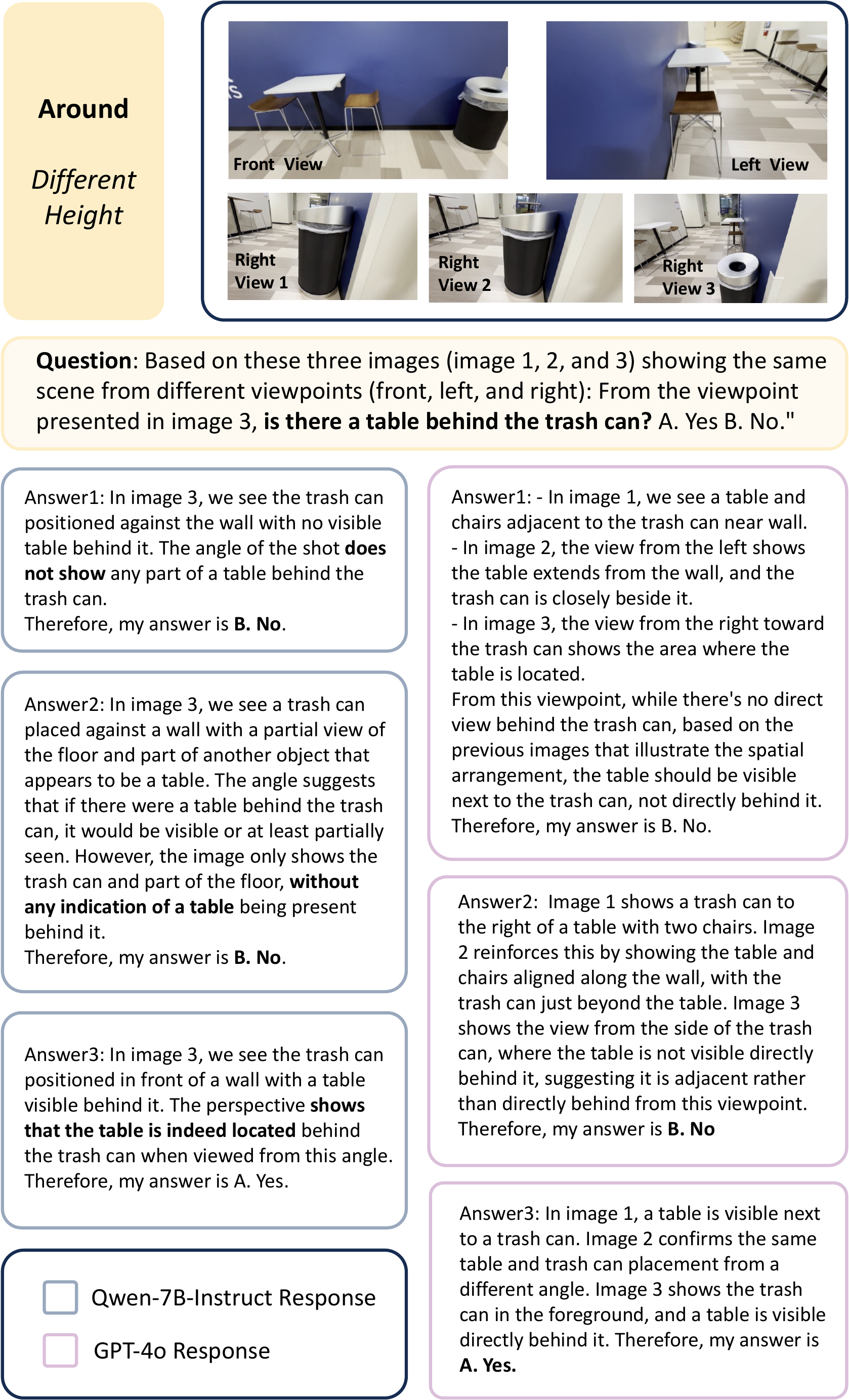}
  \caption{Example of different camera heights' impact in around setting.}
  \label{fig:around_case_ch}
\end{figure*}

First, we examine the relationship between occlusion degree and response accuracy across four visibility levels (fully visible, mostly visible, mostly occluded, fully occluded) to determine whether performance degrades proportionally with increasing occlusion. Second, we investigate the impact of camera height variation within the same lateral viewpoint, as different vertical perspectives yield distinct occlusion patterns that may challenge the model's ability to maintain spatial coherence. These paradigms evaluate whether models perform consistently when transferring spatial relationships across viewpoints, particularly in scenarios with significant object size discrepancies where smaller objects may be completely occluded from one angle but visible from another. This multifaceted analysis approach enables a more nuanced understanding of MLLMs' genuine 3D spatial reasoning capabilities beyond simple pattern recognition of 2D visual cues. We mainly evaluated GPT-4o and Qwen2.5-VL.\\\\
 \begin{figure*}[h]
  \centering
  \includegraphics[width=1\linewidth]{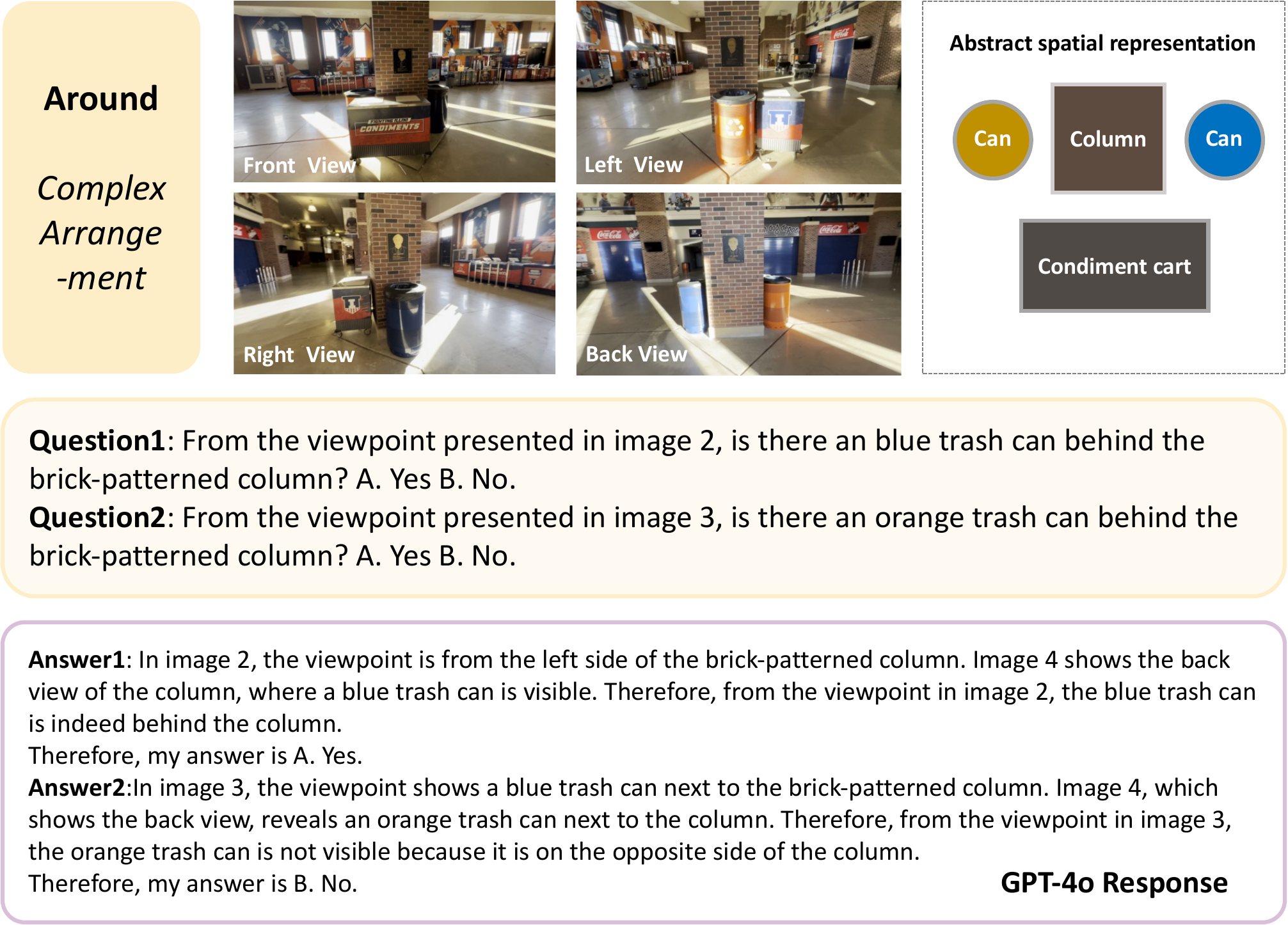}
  \caption{Example of complex spatial arrangement in around setting.}
  \label{fig:around_case_complex}
\end{figure*}
\noindent\textbf{Occlusion Degree Analysis.} Our analysis reveals a notable correlation between occlusion degree and model performance. Accuracy rates declined progressively with increasing occlusion, with an average decrease of 50.7\% between fully visible and fully occluded conditions (p \textless  0.01). Interestingly, the performance degradation was non-linear, with a precipitous drop occurring between the mostly visible and mostly occluded categories (28.7\% decrease), suggesting a potential threshold effect in the models' spatial reasoning capabilities. Error analysis in Figure\ref{fig:around_case_ch} further revealed that models frequently defaulted to proximity-based guessing when confronted with heavily occluded objects, rather than leveraging cross-view information to reason about hidden spatial relationships.\\\\
\textbf{Camera Height Impact Analysis.} Varying camera heights significantly affected model performance through different occlusion patterns. High-angle perspectives yielded 24.8\% higher accuracy than eye-level views by revealing tops of partially occluded objects and providing better scene context. This advantage was most pronounced in dense arrangements where top-down angles exposed spatial gaps between objects otherwise invisible from eye-level. Models clearly benefited from the holistic understanding afforded by elevated viewpoints, where global spatial relationships became more apparent. In contrast, eye-level perspectives with more severe occlusions led to poorer spatial reasoning, suggesting limited ability to mentally reconstruct hidden scene elements from partial visual information. \\\\
The integration of findings across all three dimensions indicates that current MLLMs possess partial but inconsistent 3D spatial reasoning capabilities. The models' performance appears heavily influenced by the visibility of key reference points across multiple viewpoints, suggesting \textbf{a reliance on visual correspondence matching rather than true 3D mental modeling}. The observed asymmetries in spatial transformation paradigms further support this hypothesis, as a robust 3D reasoning system would demonstrate consistent performance regardless of the specific transformation required.

\subsubsection{Among}
\begin{figure*}
  \centering
  \includegraphics[width=0.9\linewidth]{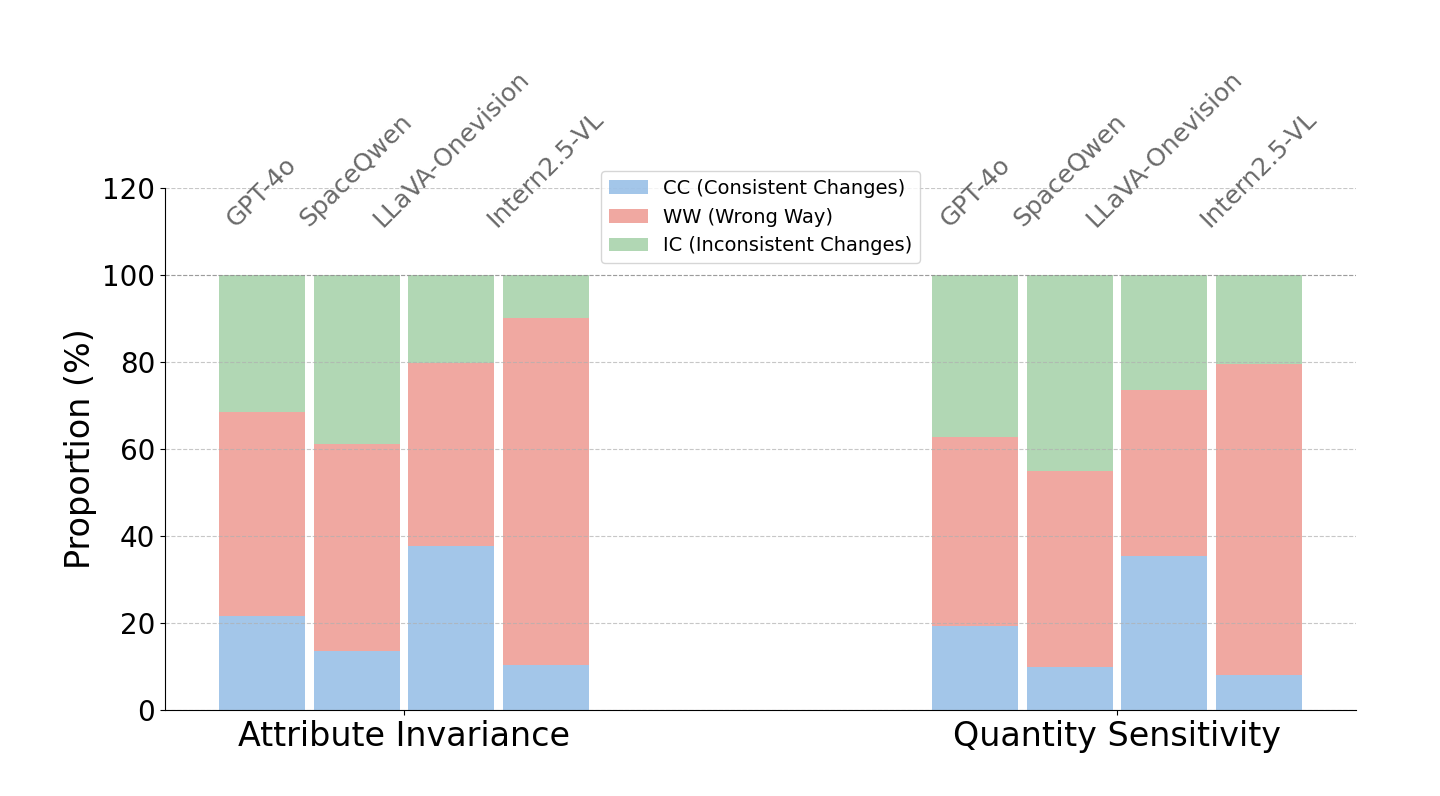}
  \caption{Paired question-answers inconsistency in two tests. We report the proportions of IC, CC and WW. Notably, SpaceQwen has a highest inconsistency(around 40\%). GPT-4o and LLaVA-Onevision exhibit more balanced performance.}
  \label{fig:ccww}
\end{figure*}

While the correctness of one single question indicates how often a model answers an isolated question correctly, it does not capture whether the model remains consistent when presented with objects with different visual attributes and different number of central objects while retaining the original peripheral scenes. 

\begin{figure*}[h]
  \centering
  \includegraphics[width=1\linewidth]{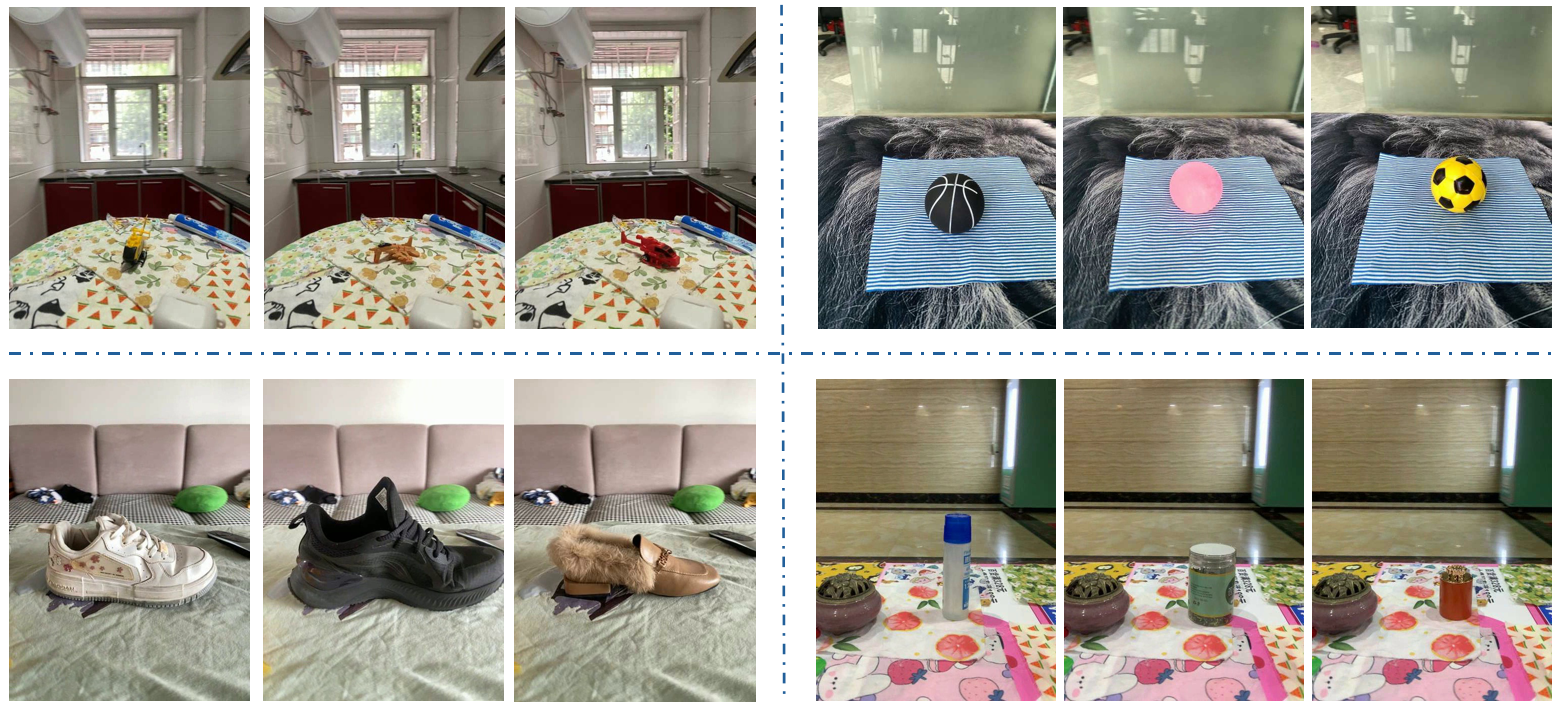}
  \caption{Examples in Attribute Invariance Test.}
  \label{fig:AIT}
\end{figure*}

To investigate this, we also propose two different tests:\\\\
\noindent\textbf{Attribute Invariance Test.} We modify only the visual attributes (e.g., color, category) of the central object while keeping the spatial configuration of all objects unchanged, as shown in Figure\ref{fig:AIT}. A robust spatial reasoning system should maintain consistent answers, as spatial relationships remain invariant despite superficial attribute changes.  
\label{sec:attribute}

\begin{figure*}[h]
  \centering
  \includegraphics[width=1\linewidth]{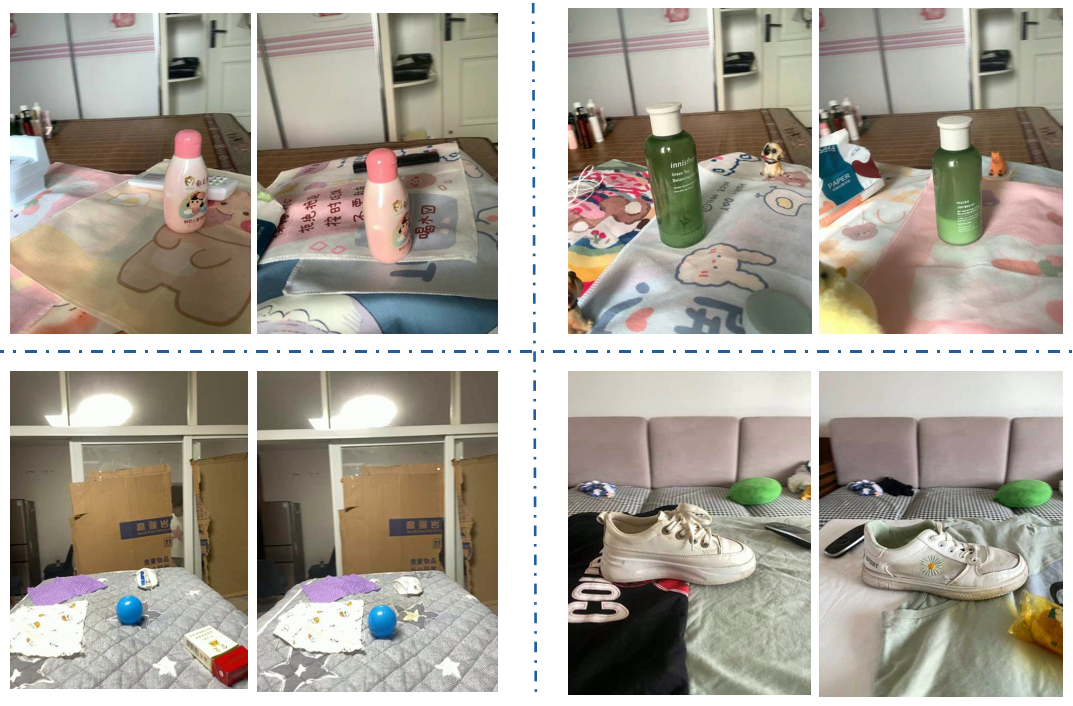}
  \caption{Examples in Quantity Sensitivity Test.}
  \label{fig:QST}
\end{figure*}

\noindent\textbf{Quantity Sensitivity Test.} We increase the number of central objects (e.g., from one to three) while retaining the original peripheral objects, as shown in Figure\ref{fig:QST}. This modification is hypothesized to enhance reasoning performance, as additional central objects provide more reference points for establishing cross-view correspondences and consistency.
\label{sec:quantity}

We also propose to look into the proportions of paired questions in tests where the answers are inconsistent with one another.
 First, we classify each paired instance into three scenarios:
 1) CC(Both Correct) when the model answers both the primary and paired question correctly, 2) WW (Both Wrong)
 when it fails both versions, and 3) IC (Inconsistent) when
 the model answers one version correctly but fails the other.

 As shown in Figure~\ref{fig:ccww}, we report the proportions of IC (in
consistent) outcomes across 4 MLLMs in two tests — two
 open-source (Intern2.5-VL, LLaVA-Onevision), a spatial model (SpaceQwen) and a closed-source GPT-4o. We have several observations: 1) SpaceQwen exhibits notably high inconsistency score IC (around
 40\%) on both tasks,
 2)  LLaVA-Onevision remain fairly
 balanced inconsistency and high performance across tests, while
InternVL vary significantly across tests.

Our systematic evaluation demonstrates MLLMs can achieve attribute-invariant spatial reasoning but struggle to utilize additional reference objects effectively. This highlights the need for: (1) enhanced geometric reasoning architectures, and (2) comprehensive benchmarks evaluating both attribute invariance and quantity sensitivity in 3D spatial understanding.

\subsection{Failure case analysis}
The observed pattern of errors indicates that models primarily rely on local relationship matching rather than inferring global spatial configurations, which represents a critical gap compared to human-like spatial reasoning abilities. Future architectural improvements should therefore focus on enhancing transitive spatial inference mechanisms and view-invariant scene representation to support more robust reasoning across multiple perspectives.
\begin{figure*}[h]
  \centering
  \includegraphics[width=1\linewidth]{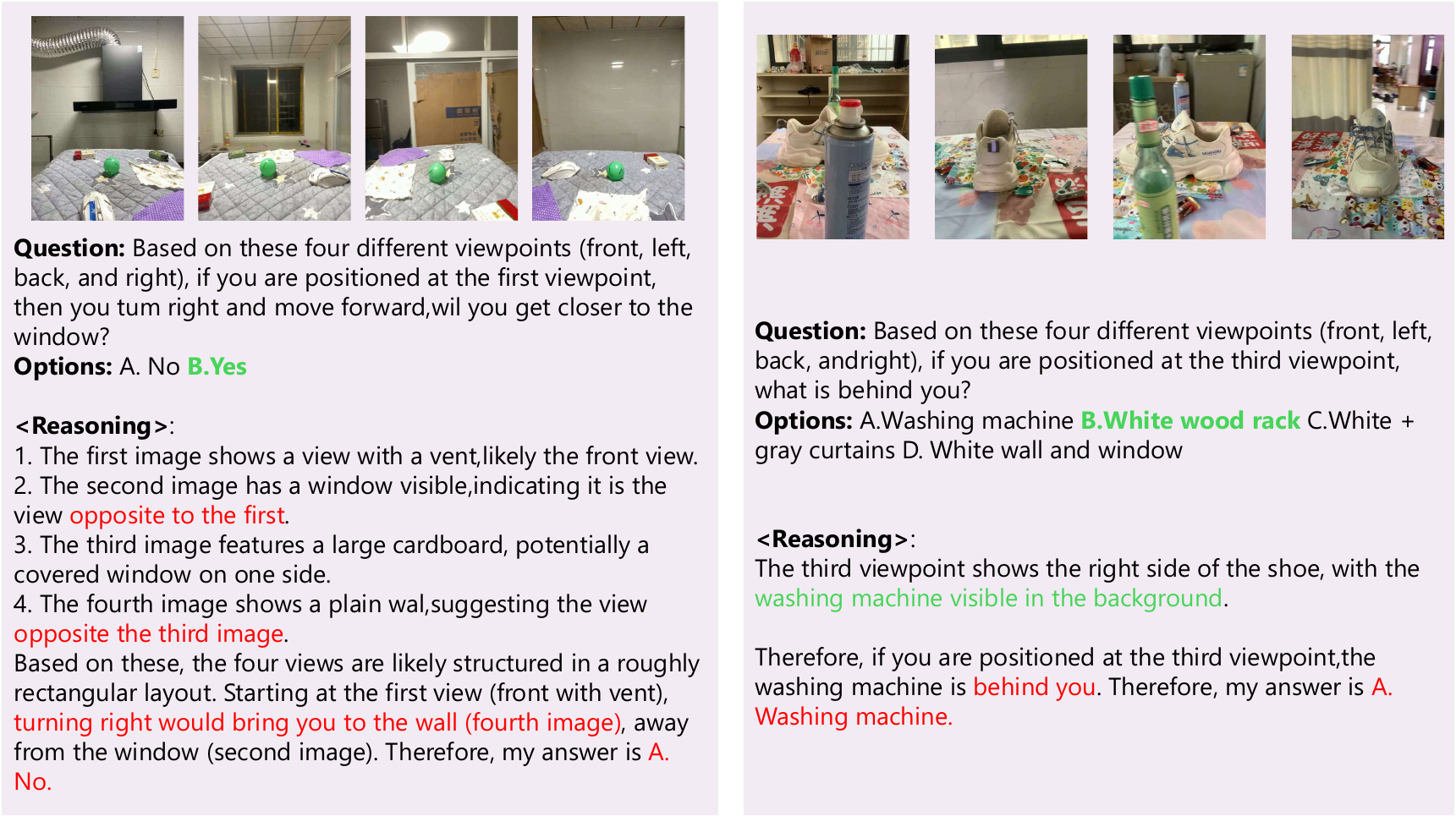}
  \caption{Failure case analysis. We show GPT4-o's reasoning process. In case 1, the model is unable to establish the spatial location corresponding to each view; In case 2, the model confuses the subject of the “behind” relationship.}
  \label{fig:collection}
\end{figure*}

\section{Data Structures as Cognitive Scaffolds, Evaluation Metrics, and Input-Output Configurations}
In this section, we provide detailed descriptions of the three data structures employed as cognitive scaffolds to approximate spatial mental models in VLMs, followed by formal definitions of the evaluation metrics employed across all experiments. Furthermore, we show the prompts for all the input-output configurations that were used across the following experiments.

\subsection{Data Structures as Cognitive Scaffolds}\label{app_sec:data_structure}
The human ability to navigate and reason about space, especially with incomplete information, is largely attributed to the formation of internal spatial mental models. These models, as extensively studied in cognitive science, are not necessarily veridical, metric-perfect replicas of the environment. Instead, they are often schematic and even distorted, yet functionally effective representations. These models can be especially useful for understanding the environment spatial layouts for agentic settings~\cite{yao2023react, wang2024rethinking}, such as embodied scenarios~\cite{liang2023code, driess2023palm, huang2023voxposer, huang2024rekep, li2024embodied, yang2025embodiedbench, tang2025uad}. Pioneering work by Barbara Tversky suggests that these internal constructs are more akin to "cognitive collages" – flexible assemblies of spatial information gleaned from various perspectives and experiences, rather than rigid, map-like blueprints~\cite{tversky1993cognitive}. These "cognitive collages" allow for the integration of fragmented observations and support reasoning across unseen perspectives. Johnson-Laird~\cite{johnson1983mental} posits that mental models, including those for space, serve as \textit{"structural analogs of the world,"} enabling individuals to simulate and infer spatial relationships, such as determining the relative positions of objects from sequential descriptions (e.g., "A is to the left of B; B is in front of C"). Research by Tversky~\cite{tversky1994spatial} has also demonstrated that individuals can construct rich, multi-dimensional mental representations even from linear, descriptive texts, and subsequently query these models from various psychological viewpoints.

Inspired by these cognitive theories, we explore three distinct data structures designed to act as cognitive scaffolds for VLMs. When VLMs are presented with limited visual input, these structures aim to approximate different facets of human spatial mental modeling: dynamic updating, integrated spatial layout representation, and inferential reasoning.

\subsubsection{View Interpolation for Dynamic Updating}
Human spatial mental models are not static; they are continuously updated with new sensory information and through mental simulation, such as imagining a change in viewpoint. To approximate this dynamic updating and mental animation capability~\cite{hegarty1992mental}, we employ view interpolation. This technique aims to bridge perceptual gaps between discrete, sparsely sampled views by generating intermediate visual frames.

\noindent\textbf{Our Setting:} In our experiments, view interpolation is implemented by inserting synthetic frames \textit{between} consecutive views provided to the model. For instance, if "1 interpolated frame" is specified, one new frame is generated and inserted between an initial view $V_n$ and the subsequent view $V_{n+1}$ (e.g., between View 1 and View 2). Similarly, "2 interpolated frames" would mean two synthetic frames are inserted in sequence between $V_n$ and $V_{n+1}$. For the interpolated frames, we either define a heuristic function to choose from the original datasets~\cite{baruch1arkitscenes, xia2024rgbd} where we sampled our data, or we use Stable Virtual Camera~\cite{zhou2025stable} to generate intermediate frames for those image groups without. This approach is intended to provide a smoother perceptual experience, potentially aiding the VLM in tracking object relations and maintaining spatial consistency across viewpoint shifts. (Refer to Figure~\ref{fig:data_examples} in the main paper for a conceptual illustration)

\subsubsection{Cognitive Maps for Integrated Spatial Layouts}
A core aspect of spatial cognition is the ability to form an allocentric (world-centered) or survey-like understanding of an environment, capturing the relative locations of objects. Tversky~\cite{tversky1993cognitive, tversky2003structures} highlights that such representations often involve different frames of reference and hierarchical structures. Cognitive maps in our context are 2D schematic representations that attempt to embody this integrated spatial layout. 

\textbf{Our Setting:} We investigate two variants of cognitive maps, both represented as structured data (e.g., JSON-like objects), to capture the spatial layout:

\begin{itemize}[leftmargin=12pt, topsep=-1pt, itemsep=0pt]
  \item We provide a 2D grid map of the scene that is related to the question to be answered.
  \item The map uses a 10$\times$10 grid, where \texttt{[0, 0]} is the top-left corner and \texttt{[9, 9]} is the bottom-right corner (i.e., bird’s-eye view).
  \item Directions are defined as follows:
    \begin{itemize}
      \item \texttt{up} = towards the top of the grid (decreasing y-value)
      \item \texttt{right} = towards the right of the grid (increasing x-value)
      \item \texttt{down} = towards the bottom of the grid (increasing y-value)
      \item \texttt{left} = towards the left of the grid (decreasing x-value)
      \item \texttt{inner} = into the 2D map (perpendicular to the grid, pointing away from you)
      \item \texttt{outer} = out of the 2D map (perpendicular to the grid, pointing toward you)
    \end{itemize}
  \item The map contains:
    \begin{itemize}
      \item \texttt{objects} — a list of all important items in the scene with their \texttt{position}
      \item \texttt{facing} — indicating the direction an object is oriented (when applicable)
      \item \texttt{views} — representing different camera viewpoints in the scene
    \end{itemize}
\end{itemize}

\begin{itemize}[leftmargin=12pt, topsep=-1pt, itemsep=0pt]
  \item \textbf{Augmented Cognitive Map:} This version explicitly integrates the observer's perspective by encoding the \texttt{positions} and \texttt{orientations} (facing directions) of the camera viewpoints within the map, alongside the objects and their locations. For instance, as depicted in our data examples (refer to Figure~\ref{fig:data_examples}, Cognitive Map - Augmented panel), an augmented map might define a list of \texttt{objects} with their \texttt{name} and \texttt{position} (e.g., \texttt{"Tissue box": \{ "position": [5, 5] \}}), and a separate list of \texttt{views} detailing each camera's \texttt{name} (e.g., \texttt{"View 1"}), \texttt{position} (e.g., \texttt{[3, 5]}), and \texttt{facing} direction (e.g., \texttt{"up"}).

  \item \textbf{Plain Cognitive Map (Object Only):} This is a more simplified, object-centric representation. It primarily focuses on the \texttt{spatial locations of objects} and, for some objects, their intrinsic \texttt{orientation} (facing direction) from a top-down survey perspective, without explicitly embedding camera view information within its structure. For example (refer to Figure~\ref{fig:data_examples}, Cognitive Map - Plain panel), a plain map might list objects like \texttt{"Potted plant"} with its \texttt{position} (e.g., \texttt{[5, 6]}) and \texttt{facing} direction (e.g., \texttt{"down"}), and another object like \texttt{"Sofa"} with only its \texttt{position} (e.g., \texttt{[4, 5]}). This type of map still allows for reasoning about object-to-object relationships and, where specified, object orientations, but abstracts away the explicit camera viewpoints that generated the scene understanding.
\end{itemize}

In both map types, coordinates represent positions on a 2D grid, and \texttt{facing} directions can be categorical (e.g., \texttt{"up"}, \texttt{"down"}, \texttt{"left"}, \texttt{"right"}, \texttt{"outer"}, \texttt{"inner"}). These structures aim to provide the VLM with an explicit, albeit potentially imperfect, schematic of the environment that it can then learn to generate and utilize for spatial reasoning tasks.

 As for the format, our JSON format has been widely adopted as a computational model providing a flexible structure for VLMs, designed to offer a bird's-eye view representation encoding the relative positions and orientations of objects \cite{yang2024thinkingspacemultimodallarge}. This representation aligns, at a high level, with the functional principles of cognitive maps in cognitive science. Our goal is to equip VLMs with a scaffold that approximates the functional role of a cognitive map to enable explicit reasoning, rather than replicating its exact neurological basis.

The use of JSON is a principled choice for interfacing with text-native VLMs, following standard practices for eliciting structured outputs. VLMs fundamentally operate on sequences of language tokens, making JSON a naturally fitting text-based format. JSON provides a structured and computationally effective means to evaluate complex spatial outputs, constituting one of the standard methods for eliciting structured knowledge from LLMs and VLMs. Although differentiable vectorized representations represent a promising research direction, current integration attempts have been widely recognized as ineffective, particularly owing to limitations in VLM comprehension.

\subsubsection{Free Form Reasoning}

Spatial mental models are not just static representations; they are actively used for inference and problem-solving~\cite{tversky1994spatial}. To approximate this procedural and inferential aspect, we utilize free-form reasoning, implemented as a natural language Chain-of-Thought (CoT)~\cite{wei2022chain} process. This encourages the VLM to externalize its step-by-step reasoning process when deducing an answer to a spatial query.

\noindent\textbf{Our Setting:} The VLM is prompted to generate a textual reasoning chain before outputting the final answer. This process is guided by a three-step principle, exemplified by the reasoning chain shown in Figure~\ref{fig:data_examples}, the reasoning chain panel. For the steps shown in that example, they are: (1) \textit{Initial Observation and Grounding:} The model first processes each available view, identifying key objects and their immediate spatial relationships within that specific viewpoint. For instance, the example chain begins with: \texttt{"In View 1, I see a potted plant, tissue box, and hand sanitizer from left to right, with a sofa behind."}  This step grounds the reasoning in direct visual evidence from individual perspectives. (2) \textit{Cross-View Integration and Environment Consolidation:}Next, the model attempts to identify consistent objects or environmental cues across the different views to recognize that they depict the same underlying 3D scene. The example reasoning continues: \texttt{"In View 2, I see the same potted plant, so both views are from the same environment."}  This step is crucial for building a unified understanding of the space from discrete observations. (3) \textit{Question-Guided Spatial Inference:} Finally, based on the specific question posed and the integrated understanding from the previous steps, the model performs step-by-step logical and spatial inferences to arrive at the answer. In the example, this involves relating the object positions across views relative to the observer's position in View 2: \texttt{"Since the hand sanitizer is rightmost in View 1, it's spatially furthest behind the potted plant when looking in View 2.\\~In View 2, the potted plant is closest to me, so the hand sanitizer is\\~the furthest from me."}

\subsection{Evaluation Metrics}\label{app_sec:evaluation_metrics}
To quantitatively assess how these data structures affect the performance of VLMs in the spatial mental modeling presented in \name, and to evaluate the quality of the generated cognitive maps, we employed the following metrics: (1) \textit{QA Accuracy}, and (2) \textit{Graph Metrics for Generated Cognitive Maps}.

\subsubsection{QA Accuracy}

QA Accuracy serves as the core metric for evaluating task performance. It quantifies the proportion of questions that the vision-language model (VLM) answers correctly out of the total number of questions. A higher QA Accuracy indicates better alignment between the model’s responses and the ground truth.

The metric is formally defined as:

$$
\text{QA Accuracy} = \frac{N_{\text{correct}}}{N_{\text{total}}} \times 100\%
$$

where $N_{\text{correct}}$ denotes the number of correctly answered questions, and $N_{\text{total}}$ is the total number of questions evaluated.

\subsubsection{Graph Metrics for Cognitive Maps}
\label{sec:graph_metrics}

To quantitatively evaluate the quality of a generated cognitive map, we use a set of structured graph-based metrics. The overall process consists of several key steps:

\begin{enumerate}[leftmargin=12pt, topsep=-1pt, itemsep=0pt]
    \item \textbf{Validity Check.} First, we ensure that the generated map is syntactically and semantically valid—i.e., it has a correct JSON format, contains interpretable object positions, and includes at least one valid object.
    
    \item \textbf{Rotation Normalization.} Since we do not enforce a fixed orientation for generated maps (to allow for flexible generation from vision-language models), we evaluate the similarity between the generated map and the ground truth across a set of 3D rotations. We always choose the best-aligned rotation to compute our similarity scores.

    \item \textbf{Structural Matching.} We define a relation graph between object pairs in each map, capturing directional and proximity-based relationships. A core part of the evaluation is determining whether these relationships in the ground truth are preserved in the generated map.

    \item \textbf{Similarity Metrics.} We compute coverage (how many ground-truth objects are present), directional similarity (relative spatial relations), and facing similarity (object orientation). These are aggregated into an overall similarity score.
    
    \item \textbf{Rotation-Invariant Isomorphism.} We also evaluate whether a generated map is graph-isomorphic to the ground truth under any allowed 3D rotation, providing a strict measure of structural correctness.
\end{enumerate}

Below, we provide precise mathematical definitions for each of these components.

\paragraph{Notation.}
A \emph{cognitive map} is a finite set of objects
\(
\mathcal O=\{o_1,\dots,o_n\}
\)
where each object \(o_i\) is associated with  
(i) a 2-D position vector \(p_i=(x_i,y_i)\in\mathbb R^{2}\) and  
(ii) an optional facing label \(f_i\in\{\text{up},\text{right},\text{down},\text{left},\text{inner},\text{outer}\}\cup\{\varnothing\}\).
For two maps, we distinguish (1) the \emph{ground-truth} map \((\mathcal O^{\star},p^{\star},f^{\star})\) and (2) a \emph{generated} map \((\mathcal O^{g},p^{g},f^{g})\).

The set of objects that appear in both maps is  
\(
\mathcal O^{c}=\mathcal O^{\star}\cap\mathcal O^{g}.
\)

\paragraph{Extended directional relation.}
We define a directional or proximity-based relationship between any ordered object pair \((o_i,o_j)\) based on their spatial arrangement and optional facing annotations. This relation is captured via the function:

\[
\mathrm{dir}(o_i,o_j)=
\begin{cases}
\text{right} & |x_j-x_i|\textgreater |y_j-y_i|\text{ and }x_j\textgreater x_i,\\
\text{left}  & |x_j-x_i|\textgreater |y_j-y_i|\text{ and }x_j<x_i,\\
\text{down}  & |y_j-y_i|\ge |x_j-x_i|\text{ and }y_j\textgreater y_i,\\
\text{up}    & |y_j-y_i|\ge |x_j-x_i|\text{ and }y_j< y_i,\\
\text{inner} & \|p_j-p_i\|_2< \delta\text{ and }(f_i=\text{inner}\ \lor\ f_j=\text{outer}),\\
\text{outer} & \|p_j-p_i\|_2< \delta\text{ and }(f_i=\text{outer}\ \lor\ f_j=\text{inner}),\\
\varnothing  & \text{otherwise},
\end{cases}
\]
with threshold \(\delta=0.5\) as in the implementation.  
These relations form a \emph{relation matrix}:
\[
R(o_i,o_j)=\mathrm{dir}(o_i,o_j).
\]

\paragraph{Coverage.}
Coverage measures how many ground-truth objects are successfully retrieved in the generated map:
\[
\mathrm{Cov}=\frac{|\mathcal O^{c}|}{|\mathcal O^{\star}|}\in[0,1].
\]

\paragraph{Directional similarity.}
We now evaluate how well the generated map preserves the directional relationships among object pairs from the ground truth. Define:
\[
\mathcal P^{\star}=\{(o_i,o_j)\in\mathcal O^{c}\times\mathcal O^{c}\mid i\neq j,\;R^{\star}(o_i,o_j)\neq\varnothing\}.
\]
Then the directional similarity score is given by:
\[
S_{\text{dir}}=
\frac{\bigl|\{(o_i,o_j)\in\mathcal P^{\star}\mid R^{g}(o_i,o_j)=R^{\star}(o_i,o_j)\}\bigr|}
     {|\mathcal P^{\star}|}\in[0,1],
\]
which corresponds to the proportion of directional relations in the ground truth that are correctly matched in the generated map.

\paragraph{Facing similarity.}
For objects with defined facing directions, we compare their orientation across the two maps:
\[
\mathcal F^{\star}=\{o_i\in\mathcal O^{c}\mid f^{\star}_i\neq\varnothing\}.
\]
Then:
\[
S_{\text{face}}=
\frac{\bigl|\{o_i\in\mathcal F^{\star}\mid f^{g}_i=f^{\star}_i\}\bigr|}
     {|\mathcal F^{\star}|}\in[0,1].
\]

\paragraph{Overall similarity.}
To aggregate the directional and facing similarities, we use a weighted combination:
\[
S_{\text{overall}}=\alpha\cdot S_{\text{dir}}+(1-\alpha)\cdot S_{\text{face}} \in[0,1],
\]
where $\alpha=0.7$ places greater emphasis on spatial layout than orientation.

\paragraph{Rotation-invariant isomorphism.}
To ensure fair comparison regardless of orientation, we define a set of 3D rotations:
\(\mathcal R=\{R_1,\dots,R_m\}\), including all 90° turns about the $z$-axis, and one 90° turn about each of the $x$- and $y$-axes.

We say the maps are \emph{rotation-invariant isomorphic} if there exists a rotation such that their relation matrices match completely:
\[
\exists\,k\in\{1,\dots,m\}\;\forall\,o_i,o_j\in\mathcal O^{\star}:
\quad R^{\star}(o_i,o_j)=R^{g}_{(k)}(o_i,o_j),
\]
where \(R^{g}_{(k)}\) is the relation matrix computed after applying \(R_k\) to the generated map.

\paragraph{Graph validity.}
Finally, a generated map is deemed \emph{valid} if: (1) It is well-formed JSON, (2) All fields conform to expected formats and constraints, and (3) At least one object has a valid position.

Together, the tuple  
\(
\bigl(\mathrm{Cov},\,S_{\text{dir}},\,S_{\text{face}},\,S_{\text{overall}},\,\text{Iso}_{\text{rot}}\bigr)
\)
provides a comprehensive, rotation-aware evaluation of how closely a generated cognitive map matches ground truth structure and orientation.

\subsection{Prompts for All Input-Output Configurations} \label{app_sec:example_configuration}
Below, we provide all prompts for the input-output configurations we investigate in our work.

\subsubsection{Example for \texttt{Raw-QA}}\label{app_sec:config_rawqa}

\begin{tcolorbox}[colback=boxbackground, colframe=accentblue, 
title=Example Prompt for \texttt{Raw-QA}, boxrule=0.5mm, width=\textwidth, arc=2mm, auto outer arc=true, breakable]
\begin{center}
    \includegraphics[width=0.8\textwidth]{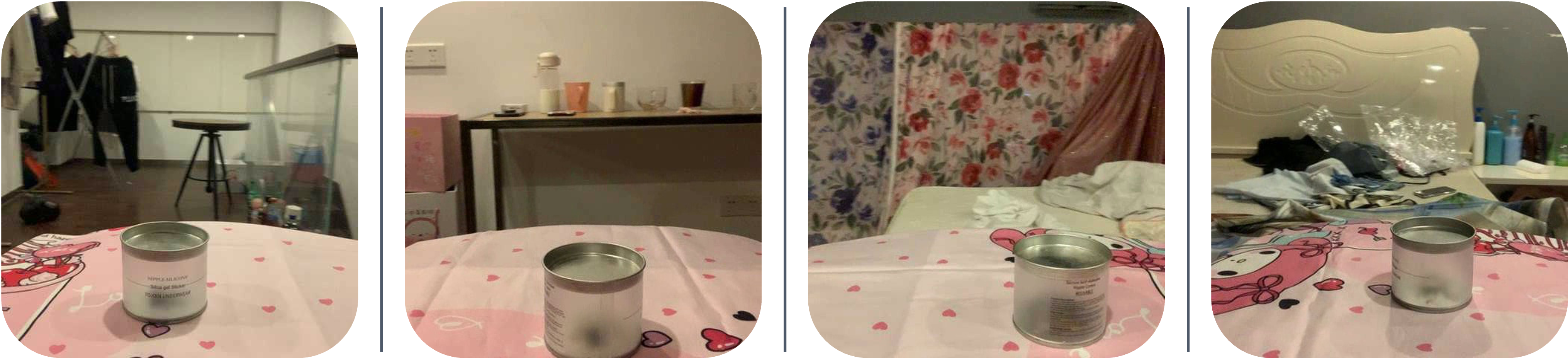}
\end{center}
[Task]
\newline
Your task is to analyze the spatial arrangement of objects in the scene by examining the provided images, which show the scene from different viewpoints.
\newline
[Answer Instruction]
\newline
You only need to provide *ONE* correct answer selecting from the options listed below. For example, if you think the correct answer is 'A. Above' from 'A. Above B. Under C. Front D. Behind', your response should **only** be '\textless answer\textgreater A. Above\textless /answer\textgreater '.
\newline
[Question]
\newline
Based on these four images (image 1, 2, 3, and 4) showing the white jar from different viewpoints (front, left, back, and right), with each camera aligned with room walls and partially capturing the surroundings: From the viewpoint presented in image 4, what is to the left of the white jar?
\newline
A. Table with cups on it B. Clothes rack C. Bed sheet with a floral pattern D. White headboard
\end{tcolorbox}

\subsubsection{Example for \texttt{\iclr{FFR}}} \label{app_sec:config_ff_rsn}
\begin{tcolorbox}[colback=boxbackground, colframe=accentblue, 
title=Example Prompt for \texttt{\iclr{FFR}}: Free-Form Reasoning, boxrule=0.5mm, width=\textwidth, arc=2mm, auto outer arc=true, breakable]
\begin{center}
    \includegraphics[width=0.8\textwidth]{figs/app_example_figure.png}
\end{center}
[Task]
\newline
Your task is to analyze the spatial arrangement of objects in the scene by examining the provided images, which show the scene from different viewpoints.
\newline
[Answer Instruction]
\newline
Please do step by step reasoning first, then give your final answer. For example, if you think the correct answer is 'A. Above' from 'A. Above B. Under C. Front D. Behind', your response should be this format: '\textless think\textgreater (replace with your reasoning here)\textless /think\textgreater \textless answer\textgreater A. Above\textless /answer\textgreater '.
\newline
[Question]
\newline
Based on these four images (image 1, 2, 3, and 4) showing the white jar from different viewpoints (front, left, back, and right), with each camera aligned with room walls and partially capturing the surroundings: From the viewpoint presented in image 4, what is to the left of the white jar?
\newline
A. Table with cups on it B. Clothes rack C. Bed sheet with a floral pattern D. White headboard
\end{tcolorbox}

\subsubsection{Example for \texttt{VI-1} and \texttt{VI-2}} \label{app_sec:config_vi}
\begin{tcolorbox}[colback=boxbackground, colframe=accentblue, 
title=Prompt for \texttt{VI-1}: View Interpolation with 1 Frame, boxrule=0.5mm, width=\textwidth, arc=2mm, auto outer arc=true, breakable]
\begin{center}
    \includegraphics[width=0.8\textwidth]{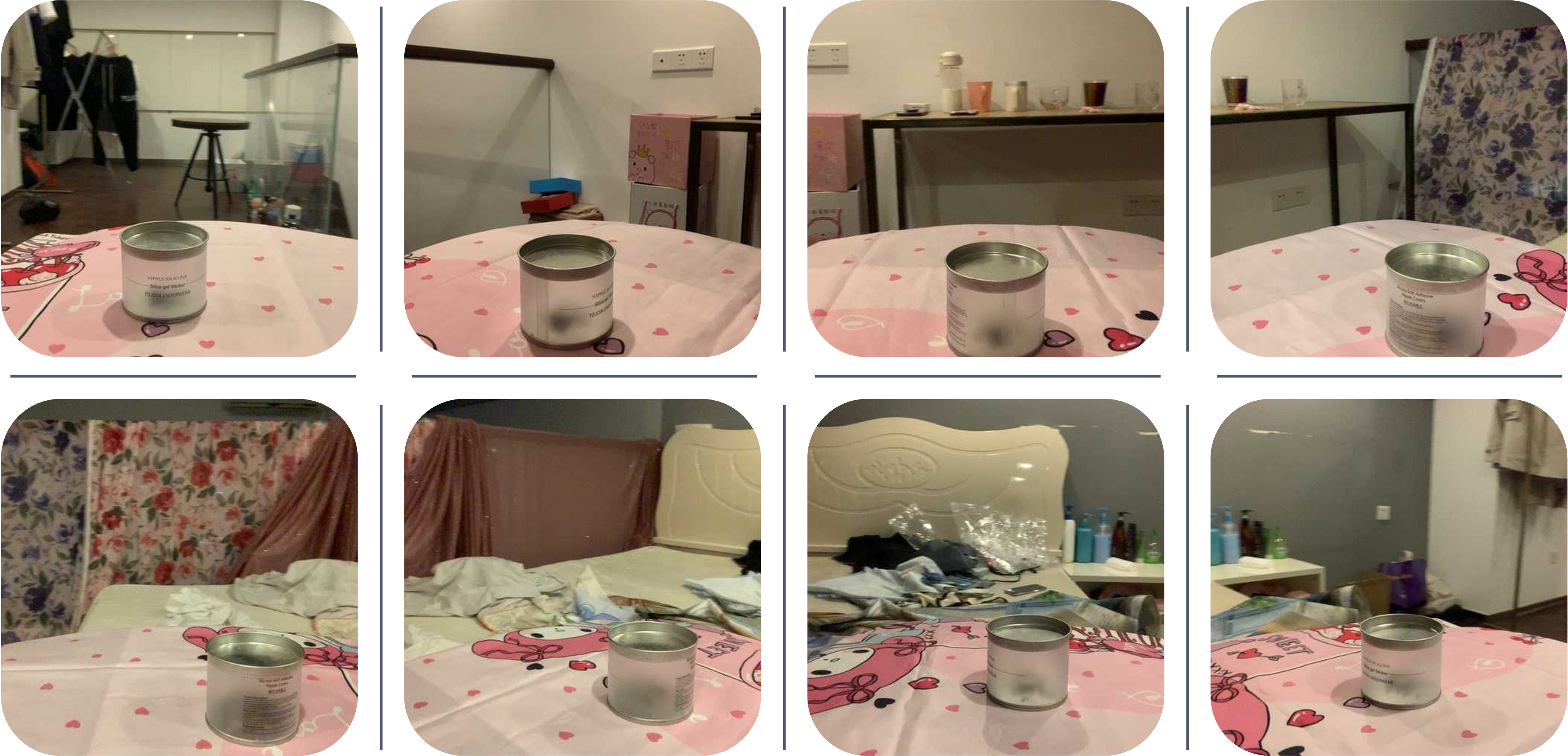}
\end{center}
[Task]
\newline
Your task is to analyze the spatial arrangement of objects in the scene by examining the provided images, which show the scene from different viewpoints.
\newline
[Answer Instruction]
\newline
You only need to provide *ONE* correct answer selecting from the options listed below. For example, if you think the correct answer is 'A. Above' from 'A. Above B. Under C. Front D. Behind', your response should **only** be '\textless answer\textgreater A. Above\textless /answer\textgreater '.
\newline
[Question]
\newline
Based on these 8 images showing the white jar from different viewpoints (from front (image 1) to left (image 3), from left (image 3) to back (image 5), from back (image 5) to right (image 7), from right (image 7) back to front (image 1)), with each camera aligned with room walls and partially capturing the surroundings: From the viewpoint presented in image 7, what is to the left of the white jar?
\newline
A. Table with cups on it B. Clothes rack C. Bed sheet with a floral pattern D. White headboard
\end{tcolorbox}

\begin{tcolorbox}[colback=boxbackground, colframe=accentblue, 
title=Prompt for \texttt{VI-2}: View Interpolation with 2 Frames, boxrule=0.5mm, width=\textwidth, arc=2mm, auto outer arc=true, breakable]
\begin{center}
    \includegraphics[width=\textwidth]{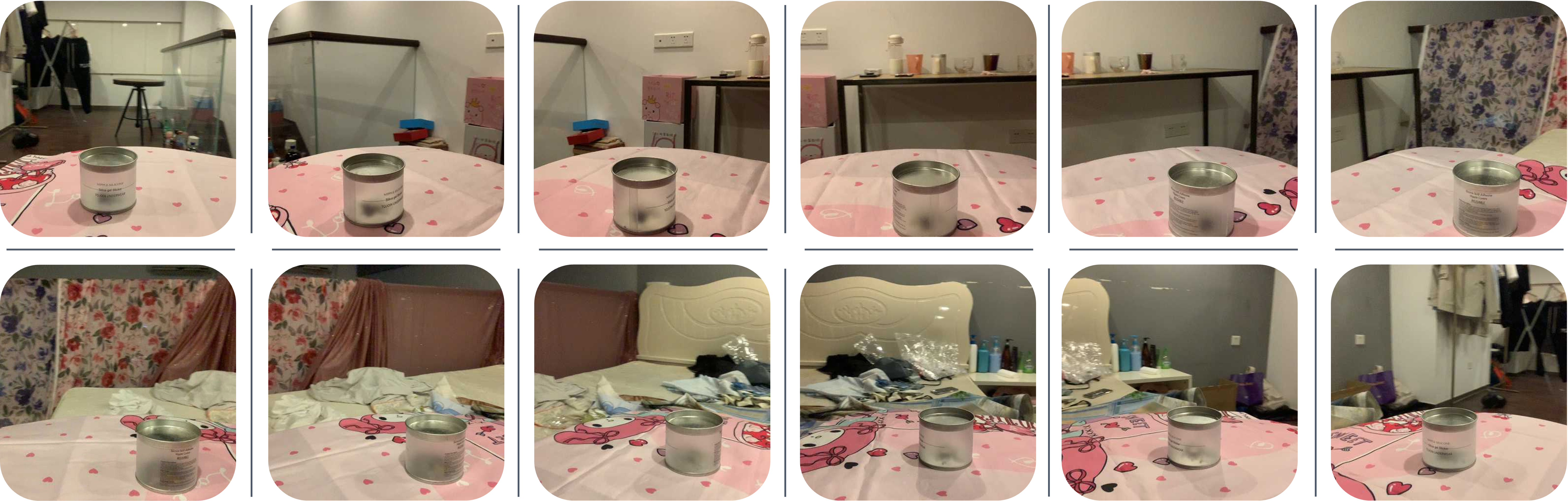}
\end{center}
[Task]
\newline
Your task is to analyze the spatial arrangement of objects in the scene by examining the provided images, which show the scene from different viewpoints.
\newline
[Answer Instruction]
\newline
You only need to provide *ONE* correct answer selecting from the options listed below. For example, if you think the correct answer is 'A. Above' from 'A. Above B. Under C. Front D. Behind', your response should **only** be '\textless answer\textgreater A. Above\textless /answer\textgreater '.
\newline
[Question]
\newline
Based on these 12 images showing the white jar from different viewpoints (from front (image 1) to left (image 4), from left (image 4) to back (image 7), from back (image 7) to right (image 10), from right (image 10) back to front (image 1)), with each camera aligned with room walls and partially capturing the surroundings: From the viewpoint presented in image 10, what is to the left of the white jar?
\newline
A. Table with cups on it B. Clothes rack C. Bed sheet with a floral pattern D. White headboard
\end{tcolorbox}

\subsubsection{Example for \texttt{Aug-CGMap-In}} \label{app_sec:config_ag_cgmap_in}
\begin{tcolorbox}[colback=boxbackground, colframe=accentblue, 
title=Prompt for \texttt{Aug-CGMap-In}: Grounded Augmented Cognitive Map as Input, boxrule=0.5mm, width=\textwidth, arc=2mm, auto outer arc=true, breakable]
\begin{center}
    \includegraphics[width=0.8\textwidth]{figs/app_example_figure.png}
\end{center}
[Task]
\newline
Your task is to analyze the spatial arrangement of objects in the scene by examining the provided images, which show the scene from different viewpoints. Also, we provide you a cognitive map that shows the general layout for the scene. Please use the cognitive map to reason and answer the question.
\newline
[Answer Instruction]
\newline
You only need to provide *ONE* correct answer selecting from the options listed below. For example, if you think the correct answer is 'A. Above' from 'A. Above B. Under C. Front D. Behind', your response should **only** be '\textless answer\textgreater A. Above\textless /answer\textgreater '.
\newline
[Cognitive Map Format]
\newline
We provide you a 2D grid map of the scene that is related to the question you should answer. Below is the description of the map:
\newline
- The map uses a 10x10 grid where [0,0] is at the top-left corner and [9,9] is at the bottom-right corner 
\newline
- The map is shown in the bird's view
\newline
- Directions are defined as:
\newline
  * up = towards the top of the grid (decreasing y-value)
\newline
  * right = towards the right of the grid (increasing x-value)
\newline
  * down = towards the bottom of the grid (increasing y-value)
\newline
  * left = towards the left of the grid (decreasing x-value)
\newline
  * inner = straight into the 2D map (perpendicular to the grid, pointing away from you)
\newline
  * outer = straight out of the 2D map (perpendicular to the grid, pointing towards you)
\newline
- "objects" lists all important items in the scene with their positions
\newline
- "facing" indicates which direction an object is oriented towards (when applicable)
\newline
- "views" represents the different camera viewpoints in the scene
\newline
Below is the cognitive map of the scene related to the question. Please use it to reason and answer the question.
\begin{verbatim}
```json
{
  "objects": [
    {"name": "white jar", "position": [5, 5]},
    {"name": "bed sheet with a floral pattern", 
    "position": [5, 8]},
    {"name": "white headboard", "position": [2, 5]},
    {"name": "clothes rack", "position": [5, 2]},
    {"name": "table with cups on it", "position": [8, 5]}
  ],
  "views": [
    {"name": "Image 1", "position": [5, 6], "facing": "up"},
    {"name": "Image 2", "position": [4, 5], "facing": "right"},
    {"name": "Image 3", "position": [5, 4], "facing": "down"},
    {"name": "Image 4", "position": [6, 5], "facing": "left"}
  ]
}
```
\end{verbatim}
[Question]
\newline
Based on these four images (image 1, 2, 3, and 4) showing the white jar from different viewpoints (front, left, back, and right), with each camera aligned with room walls and partially capturing the surroundings: From the viewpoint presented in image 4, what is to the left of the white jar?
\newline
A. Table with cups on it B. Clothes rack C. Bed sheet with a floral pattern D. White headboard
\end{tcolorbox}

\subsubsection{Example for \texttt{Aug-CGMap-Out}} \label{app_sec:config_ag_cgmap_out}
\begin{tcolorbox}[colback=boxbackground, colframe=accentblue, 
title=Prompt for \texttt{Aug-CGMap-Out}: Ask VLM to Output Augmented Cognitive Map and Direct Answer, boxrule=0.5mm, width=\textwidth, arc=2mm, auto outer arc=true, breakable]
\begin{center}
    \includegraphics[width=0.8\textwidth]{figs/app_example_figure.png}
\end{center}
[Task]
\newline
Your task is to analyze the spatial arrangement of objects in the scene by examining the provided images, which show the scene from different viewpoints. You will then create a detailed cognitive map representing the scene using a 10x10 grid coordinate system.
\newline
[Rules]
\newline
1. Focus ONLY on these categories of objects in the scene: \{white jar, bed sheet with a floral pattern, white headboard, clothes rack, table with cups on it\}
\newline
2. Create a cognitive map with the following structure in the bird's view:
\newline
   - A 10x10 grid where [0,0] is at the top-left corner and [9,9] is at the bottom-right corner
\newline
   - up = towards the top of the grid (decreasing y)
\newline
   - right = towards the right of the grid (increasing x)
\newline
   - down = towards the bottom of the grid (increasing y)
\newline
   - left = towards the left of the grid (decreasing x)
\newline
   - inner = straight into the 2D map (perpendicular to the grid, pointing away from you)
\newline
   - outer = straight out of the 2D map (perpendicular to the grid, pointing towards you)
\newline
   - Include positions of all objects from the specified categories
\newline
   - Estimate the center location (coordinates [x, y]) of each instance within provided categories
\newline
   - If a category contains multiple instances, include all of them
\newline
   - Each object's estimated location should accurately reflect its real position in the scene, preserving the relative spatial relationships among all objects
\newline
   - Combine and merge information from the images since they are pointing to the same scene, calibrating the object locations accordingly
\newline
   - Include camera positions and directions for each view
\newline
3. Carefully integrate information from all views to create a single coherent spatial representation.
\newline
[Answer Instruction]
\newline
1. Given the provided views and main objects mentioned in the above rules, you **MUST** present your cognitive map in the following JSON format **before your answer**:
\begin{verbatim}
```json
{
  "objects": [
    {"name": "object_name", "position": [x, y], 
      "facing": "direction"},
    {"name": "object_without_orientation", "position": [x, y]}
  ],
  "views": [
    {"name": "View/Image 1", "position": [x, y], 
      "facing": "direction"},
    {"name": "View/Image 2", "position": [x, y], 
      "facing": "direction"}
  ]
}
```
\end{verbatim}
2. Next, provide *ONE* correct answer selecting from the options. Your answer field must be in the format like "A. Above".
\newline
3. In general, your response's format should be like "Based on my observation, the answer is:
\textless cogmap\textgreater (Replace with your cogmap here)\textless /cogmap\textgreater \textless answer\textgreater (Replace with your answer here)\textless /answer\textgreater ". Your option must be from the available options.
\newline
[Question]
\newline
Based on these four images (image 1, 2, 3, and 4) showing the white jar from different viewpoints (front, left, back, and right), with each camera aligned with room walls and partially capturing the surroundings: From the viewpoint presented in image 4, what is to the left of the white jar?
\newline
A. Table with cups on it B. Clothes rack C. Bed sheet with a floral pattern D. White headboard
\end{tcolorbox}

\subsubsection{Example for \texttt{Plain-CGMap-Out}} \label{app_sec:config_plain_cgmap_out}
\begin{tcolorbox}[colback=boxbackground, colframe=accentblue, 
title=Prompt for \texttt{Plain-CGMap-Out}: Ask VLM to Output Plain Cognitive Map and Direct Answer, boxrule=0.5mm, width=\textwidth, arc=2mm, auto outer arc=true, breakable]
\begin{center}
    \includegraphics[width=0.8\textwidth]{figs/app_example_figure.png}
\end{center}
[Task]
\newline
Your task is to analyze the spatial arrangement of objects in the scene by examining the provided images, which show the scene from different viewpoints. You will then create a detailed cognitive map representing the scene using a 10x10 grid coordinate system.
\newline
[Rules]
\newline
1. Focus ONLY on these categories of objects in the scene: \{white jar, bed sheet with a floral pattern, white headboard, clothes rack, table with cups on it\}
\newline
2. Create a cognitive map with the following structure in the bird's view:
\newline
   - A 10x10 grid where [0, 0] is at the top-left corner and [9, 9] is at the bottom-right corner
\newline
   - up = towards the top of the grid (decreasing y)
\newline
   - right = towards the right of the grid (increasing x)
\newline
   - down = towards the bottom of the grid (increasing y)
\newline
   - left = towards the left of the grid (decreasing x)
\newline
   - Include positions of all objects from the specified categories
\newline
   - Estimate the center location (coordinates [x, y]) of each instance within provided categories
\newline
   - If a category contains multiple instances, include all of them
\newline
   - Object positions must maintain accurate relative spatial relationships
\newline
   - Combine and merge information from the images since they are pointing to the same scene, calibrating the object locations with grid coordinates accordingly
\newline
3. Carefully integrate information from all views to create a single coherent spatial representation.
\newline
[Answer Instruction]
\newline
1. Given the provided views and main objects mentioned in the above rules, you **MUST** present your cognitive map in the following JSON format **before your reasoning**:
\begin{verbatim}
```json
{
    "object_category_1": {"position": [x, y]},
    "object_category_2": {"position": [x, y], 
        "facing": "direction"}, 
    # if the object is asked for orientation
    ...
}
```
\end{verbatim}
2. Next, provide *ONE* correct answer selecting from the options. Your answer field must be in the format like "A. Above"
\newline
3. In general, your response's format should be like "Based on my observation, the answer is:
\textless cogmap\textgreater (Replace with your cogmap here)\textless /cogmap\textgreater \textless answer\textgreater (Replace with your answer here)\textless /answer\textgreater ". Your option must be from the available options.
\newline
[Question]
\newline
Based on these four images (image 1, 2, 3, and 4) showing the white jar from different viewpoints (front, left, back, and right), with each camera aligned with room walls and partially capturing the surroundings: From the viewpoint presented in image 4, what is to the left of the white jar?
\newline
A. Table with cups on it B. Clothes rack C. Bed sheet with a floral pattern D. White headboard
\end{tcolorbox}

\subsubsection{Example for \texttt{Plain-CGMap-FFR-Out}} \label{app_sec:config_plain_cgmap_ffr_out}
\begin{tcolorbox}[colback=boxbackground, colframe=accentblue, 
title=Prompt for \texttt{Plain-CGMap-FFR-Out}: Ask VLM to Output Plain Cognitive Map and Free-Form Reasoning, boxrule=0.5mm, width=\textwidth, arc=2mm, auto outer arc=true, breakable]
\begin{center}
    \includegraphics[width=0.8\textwidth]{figs/app_example_figure.png}
\end{center}
[Task]
\newline
Your task is to analyze the spatial arrangement of objects in the scene by examining the provided images, which show the scene from different viewpoints. You will then create a detailed cognitive map representing the scene using a 10x10 grid coordinate system.
\newline
[Rules]
\newline
1. Focus ONLY on these categories of objects in the scene: \{white jar, bed sheet with a floral pattern, white headboard, clothes rack, table with cups on it\}
\newline
2. Create a cognitive map with the following structure in the bird's view:
\newline
   - A 10x10 grid where [0, 0] is at the top-left corner and [9, 9] is at the bottom-right corner
\newline
   - up = towards the top of the grid (decreasing y)
\newline
   - right = towards the right of the grid (increasing x)
\newline
   - down = towards the bottom of the grid (increasing y)
\newline
   - left = towards the left of the grid (decreasing x)
\newline
   - Include positions of all objects from the specified categories
\newline
   - Estimate the center location (coordinates [x, y]) of each instance within provided categories
\newline
   - If a category contains multiple instances, include all of them
\newline
   - Object positions must maintain accurate relative spatial relationships
\newline
   - Combine and merge information from the images since they are pointing to the same scene, calibrating the object locations with grid coordinates accordingly
\newline
3. Carefully integrate information from all views to create a single coherent spatial representation.
\newline
[Answer Instruction]
\newline
1. Given the provided views and main objects mentioned in the above rules, you **MUST** present your cognitive map in the following JSON format **before your reasoning**:
\begin{verbatim}
```json
{
    "object_category_1": {"position": [x, y]},
    "object_category_2": {"position": [x, y], 
    "facing": "direction"}, 
    # if the object is asked for orientation
    ...
}
```
\end{verbatim}
2. Next, please also provide your reasons step by step in details, then provide *ONE* correct answer selecting from the options. Your answer field must be in the format like "A. Above"
\newline
3. In general, your response's format should be like "Based on my observation, the answer is:
\textless cogmap\textgreater (Replace with your cogmap here)\textless /cogmap\textgreater \textless think\textgreater (Replace with your reasoning here)\textless /think\textgreater \textless answer\textgreater (Replace with your answer here)\textless /answer\textgreater ". Your option must be from the available options.
\newline
[Question]
\newline
Based on these four images (image 1, 2, 3, and 4) showing the white jar from different viewpoints (front, left, back, and right), with each camera aligned with room walls and partially capturing the surroundings: From the viewpoint presented in image 4, what is to the left of the white jar?
\newline
A. Table with cups on it B. Clothes rack C. Bed sheet with a floral pattern D. White headboard
\end{tcolorbox}

\subsubsection{Example for \texttt{Aug-CGMap-FFR-Out}} \label{app_sec:config_ag_cgmap_ffr_out}
\begin{tcolorbox}[colback=boxbackground, colframe=accentblue, 
title=Prompt for \texttt{Aut-CGMap-FFR-Out}: Ask VLM to Output Augmented Cognitive Map and Free-Form Reasoning, boxrule=0.5mm, width=\textwidth, arc=2mm, auto outer arc=true, breakable]
\begin{center}
    \includegraphics[width=0.8\textwidth]{figs/app_example_figure.png}
\end{center}
[Task]
\newline
Your task is to analyze the spatial arrangement of objects in the scene by examining the provided images, which show the scene from different viewpoints. You will then create a detailed cognitive map representing the scene using a 10x10 grid coordinate system.
\newline
[Rules]
\newline
1. Focus ONLY on these categories of objects in the scene: \{white jar, bed sheet with a floral pattern, white headboard, clothes rack, table with cups on it\}
\newline
2. Create a cognitive map with the following structure in the bird's view:
\newline
   - A 10x10 grid where [0,0] is at the top-left corner and [9,9] is at the bottom-right corner
\newline
   - up = towards the top of the grid (decreasing y)
\newline
   - right = towards the right of the grid (increasing x)
\newline
   - down = towards the bottom of the grid (increasing y)
\newline
   - left = towards the left of the grid (decreasing x)
\newline
   - inner = straight into the 2D map (perpendicular to the grid, pointing away from you)
\newline
   - outer = straight out of the 2D map (perpendicular to the grid, pointing towards you)
\newline
   - Include positions of all objects from the specified categories
\newline
   - Estimate the center location (coordinates [x, y]) of each instance within provided categories
\newline
   - If a category contains multiple instances, include all of them
\newline
   - Each object's estimated location should accurately reflect its real position in the scene, preserving the relative spatial relationships among all objects
\newline
   - Combine and merge information from the images since they are pointing to the same scene, calibrating the object locations accordingly
\newline
   - Include camera positions and directions for each view
\newline
3. Carefully integrate information from all views to create a single coherent spatial representation.
\newline
[Answer Instruction]
\newline
1. Given the provided views and main objects mentioned in the above rules, you **MUST** present your cognitive map in the following JSON format **before your reasoning**:
\begin{verbatim}
```json
{
  "objects": [
    {"name": "object_name", "position": [x, y], 
    "facing": "direction"},
    {"name": "object_without_orientation", "position": [x, y]}
  ],
  "views": [
    {"name": "View/Image 1", "position": [x, y], 
      "facing": "direction"},
    {"name": "View/Image 2", "position": [x, y], 
      "facing": "direction"}
  ]
}
```
\end{verbatim}
2. Next, please also provide your reasons step by step in details, then provide *ONE* correct answer selecting from the options. Your answer field must be in the format like "A. Above"
\newline
3. In general, your response's format should be like "Based on my observation, the answer is:
\textless cogmap\textgreater (Replace with your cogmap here)\textless /cogmap\textgreater \textless think\textgreater (Replace with your reasoning here)\textless /think\textgreater \textless answer\textgreater (Replace with your answer here)\textless /answer\textgreater ". Your option must be from the available options.
\newline
[Question]
\newline
Based on these four images (image 1, 2, 3, and 4) showing the white jar from different viewpoints (front, left, back, and right), with each camera aligned with room walls and partially capturing the surroundings: From the viewpoint presented in image 4, what is to the left of the white jar?
\newline
A. Table with cups on it B. Clothes rack C. Bed sheet with a floral pattern D. White headboard
\end{tcolorbox}

\subsubsection{Example for \texttt{CGMap-In-FFR-Out}} \label{app_sec:config_cgmap_in_ffr_out}
\begin{tcolorbox}[colback=boxbackground, colframe=accentblue, 
title=Prompt for \texttt{CGMap-In-FFR-Out}: Input VLM with Grounded Cognitive Map and Output with Free-Form Reasoning, boxrule=0.5mm, width=\textwidth, arc=2mm, auto outer arc=true, breakable]
\begin{center}
    \includegraphics[width=0.8\textwidth]{figs/app_example_figure.png}
\end{center}
[Task]
\newline
Your task is to analyze the spatial arrangement of objects in the scene by examining the provided images, which show the scene from different viewpoints. Also, we provide you a cognitive map that shows the general layout for the scene. Please use the cognitive map to reason and answer the question.
\newline
[Answer Instruction]
\newline
Please do step by step reasoning first, then give your final answer. For example, if you think the correct answer is 'A. Above' from 'A. Above B. Under C. Front D. Behind', your response should be this format: '\textless think\textgreater (replace with your reasoning here)\textless /think\textgreater \textless answer\textgreater A. Above\textless /answer\textgreater '.
[Cognitive Map Format]
\newline
We provide you a 2D grid map of the scene that is related to the question you should answer. Below is the description of the map:
\newline
- The map uses a 10x10 grid where [0,0] is at the top-left corner and [9,9] is at the bottom-right corner
\newline
- The map is shown in the bird's view
\newline
- Directions are defined as:
\newline
  * up = towards the top of the grid (decreasing y-value)
\newline
  * right = towards the right of the grid (increasing x-value)
\newline
  * down = towards the bottom of the grid (increasing y-value)
\newline
  * left = towards the left of the grid (decreasing x-value)
\newline
  * inner = straight into the 2D map (perpendicular to the grid, pointing away from you)
\newline
  * outer = straight out of the 2D map (perpendicular to the grid, pointing towards you)
\newline
- "objects" lists all important items in the scene with their positions
\newline
- "facing" indicates which direction an object is oriented towards (when applicable)
\newline
- "views" represents the different camera viewpoints in the scene
\newline
Below is the cognitive map of the scene related to the question. Please use it to reason and answer the question.
\begin{verbatim}
```json
{
  "objects": [
    {"name": "white jar", "position": [5, 5]},
    {"name": "bed sheet with a floral pattern", 
        "position": [5, 8]},
    {"name": "white headboard", "position": [2, 5]},
    {"name": "clothes rack", "position": [5, 2]},
    {"name": "table with cups on it", "position": [8, 5]}
  ],
  "views": [
    {"name": "Image 1", "position": [5, 6], "facing": "up"},
    {"name": "Image 2", "position": [4, 5], "facing": "right"},
    {"name": "Image 3", "position": [5, 4], "facing": "down"},
    {"name": "Image 4", "position": [6, 5], "facing": "left"}
  ]
}
```
\end{verbatim}
[Question]
\newline
Based on these four images (image 1, 2, 3, and 4) showing the white jar from different viewpoints (front, left, back, and right), with each camera aligned with room walls and partially capturing the surroundings: From the viewpoint presented in image 4, what is to the left of the white jar?
\newline
A. Table with cups on it B. Clothes rack C. Bed sheet with a floral pattern D. White headboard
\end{tcolorbox}

\section{Which Scaffolds Best Guide Spatial Thinking in Unchanged VLMs?} \label{app_sec:prompt}
The primary experimental setup for evaluating how different structural scaffolds can guide spatial reasoning in unchanged VLMs, has been comprehensively described in Section 3 of the main paper. This current section is dedicated to presenting qualitative examples of model outputs, serving as case studies for the different scaffolding configurations tested.

For precise details on the input prompt structures and specific instructions provided to the \modelname for each experimental condition (e.g., \texttt{Raw-QA}, \texttt{\iclr{FFR}}, \texttt{Plain-CGMap-FFR-Out}), please refer to Appendix~\ref{app_sec:example_configuration}.

\subsection{VLM Response Examples for Configurations in Section~\ref{app_sec:example_configuration}}
\begin{tcolorbox}[colback=boxbackground, colframe=accentblue, 
title=Output for \texttt{Raw-QA}: Input Prompt Refer to \ref{app_sec:config_rawqa}, boxrule=0.5mm, width=\textwidth, arc=2mm, auto outer arc=true, breakable]
\textcolor{mygreen}{\bf C. Bed sheet with a floral pattern}
\end{tcolorbox}

\bigskip

\begin{tcolorbox}[colback=boxbackground, colframe=accentblue, 
title=Output for \texttt{VI}: Input Prompt Refer to \ref{app_sec:config_vi}, boxrule=0.5mm, width=\textwidth, arc=2mm, auto outer arc=true, breakable]
\textcolor{red}{\bf B. Clothes rack}
\end{tcolorbox}

\bigskip

\begin{tcolorbox}[colback=boxbackground, colframe=accentblue, 
title=Output for \texttt{Aug-CGMap-In}: Input Prompt Refer to \ref{app_sec:config_ag_cgmap_in}, boxrule=0.5mm, width=\textwidth, arc=2mm, auto outer arc=true, breakable]
\textcolor{mygreen}{\bf C. Bed sheet with a floral pattern}
\end{tcolorbox}

\bigskip

\begin{tcolorbox}[colback=boxbackground, colframe=accentblue, 
title=Output for \texttt{\iclr{FFR}}: Input Prompt Refer to \ref{app_sec:config_ff_rsn}, boxrule=0.5mm, width=\textwidth, arc=2mm, auto outer arc=true, breakable]
\textless think\textgreater To determine what is to the left of the white jar from the viewpoint presented in image 4, let's analyze the scene:
\newline
- In image 4, the white jar is placed on a table covered with a pink tablecloth featuring heart patterns.
\newline
- The table has a glass top and is positioned against a wall.
\newline
- To the left of the jar, there is a piece of furniture that appears to be a bed or a bed frame.
\newline
Given this information, the object to the left of the white jar is most likely a bed or a bed frame. \textless /think\textgreater 
\newline
\textless answer\textgreater \textcolor{mygreen}{\bf C. Bed sheet with a floral pattern}\textless /answer\textgreater 
\end{tcolorbox}

\bigskip

\begin{tcolorbox}[colback=boxbackground, colframe=accentblue, 
title=Output for \texttt{Aug-CGMap-FFR-Out}: Input Prompt Refer to \ref{app_sec:config_ag_cgmap_ffr_out}, boxrule=0.5mm, width=\textwidth, arc=2mm, auto outer arc=true, breakable]
\begin{verbatim}
```json
{
  "objects": [
    {"name": "white jar", "position": [150, 446], 
        "facing": "left"},
    {"name": "bed sheet with a floral pattern", 
        "position": [0, 317], 
    "facing": "left"},
    {"name": "clothes rack", "position": [0, 108], 
    "facing": "left"},
    {"name": "table with cups on it", "position": [188, 200], 
    "facing": "left"},
    {"name": "white headboard", "position": [188, 317], 
    "facing": "left"}
  ],
  "views": [
    {"name": "Image 1", "position": [0, 0], "facing": "left"},
    {"name": "Image 2", "position": [0, 0], "facing": "left"},
    {"name": "Image 3", "position": [0, 0], "facing": "left"},
    {"name": "Image 4", "position": [0, 0], "facing": "left"}
  ]
}
```
\end{verbatim}
\textless CogMap\textgreater 
\newline
\textless Your cognitive map\textgreater 
\newline
\textless Reasoning\textgreater 
\newline
From the viewpoint presented in image 4, we can see that the white jar is positioned in the foreground, facing left. The bed sheet with a floral pattern is located directly behind the white jar, covering the lower part of the image. The clothes rack is visible on the left side of the image, slightly above the bed sheet. The table with cups on it is situated further back, to the right of the white jar. The white headboard is also visible, but it is not as prominent as the other objects due to its position behind the bed sheet.
\newline
Therefore, my answer is A. Table with cups on it.
\newline
\textless /Answer\textgreater 
\newline
Therefore, my answer is \textcolor{red}{\bf A. Table with cups on it}.
\end{tcolorbox}

\bigskip

\begin{tcolorbox}[colback=boxbackground, colframe=accentblue, 
title=Output for \texttt{Plain-CGMap-FFR-Out}: Input Prompt Refer to \ref{app_sec:config_plain_cgmap_ffr_out}, boxrule=0.5mm, width=\textwidth, arc=2mm, auto outer arc=true, breakable]
\begin{verbatim}
```json
{
    "white_jar": {"position": [150, 440], "facing": "left"},
    "table_with_cups_on_it": {"position": [187, 260], 
    "facing": "right"}
}
```
\end{verbatim}
\textless CogMap\textgreater 
\newline
\textless Your cognitive map\textgreater 
\newline
\textless Reasoning\textgreater 
\newline
From the viewpoint presented in image 4, we can see that the white jar is positioned on the bed sheet with a floral pattern. The table with cups on it is located to the right of the white jar, as indicated by its position in the image. The clothes rack is not visible in this particular view, and the white headboard is not directly related to the white jar or the table with cups on it. Therefore, the correct answer is A. Table with cups on it.
\newline
\textless Answer\textgreater 
\newline
Therefore, my answer is \textcolor{red}{\bf A. Table with cups on it}.
\end{tcolorbox}

\bigskip

\begin{tcolorbox}[colback=boxbackground, colframe=accentblue, 
title=Output for \texttt{CGMap-In-FFR-Out}: Input Prompt Refer to \ref{app_sec:config_cgmap_in_ffr_out}, boxrule=0.5mm, width=\textwidth, arc=2mm, auto outer arc=true, breakable]
\textcolor{mygreen}{\bf C. Bed sheet with a floral pattern}
\end{tcolorbox}

\subsection{Additional Graph Metrics for Generated Graphs}
To quantitatively assess the quality of the cognitive maps produced by our models, we employ a suite of graph-based metrics detailed in Section C.2.2. Table~\ref{tab:graph_metrics_comparison} presents a comparison of the cognitive maps generated by two of our methods: \texttt{Aug-CGMap-FFR-Out} and \texttt{Plain-CGMap-FFR-Out}.

The results highlight several key findings:
\begin{itemize}[leftmargin=12pt, topsep=-1pt, itemsep=0pt]
\item \textbf{High Validity}: Both methods demonstrate a strong ability to generate syntactically and semantically valid maps, achieving high \textbf{Valid Rates} (\textbf{94.95\%} for \texttt{Plain-CGMap-FFR-Out} and \textbf{92.57\%} for \texttt{Aug-CGMap-FFR-Out}). This indicates that the models successfully learned the required JSON structure for the cognitive maps.
\item \textbf{Structural Complexity}: Achieving perfect structural replication of the ground truth remains challenging, as shown by the modest \textbf{Isomorphism Rates}. The \texttt{Plain-CGMap-FFR-Out} method performs significantly better, with \textbf{7.43\%} of its maps being structurally identical (isomorphic) to the ground truth, compared to a mere \textbf{0.10\%} for the augmented map method.
\item \textbf{Superior Similarity Performance}: A clear performance difference in semantic similarity is evident. The \texttt{Aug-CGMap-FFR-Out} method, which explicitly includes camera views, achieves a substantially higher \textbf{Overall Similarity} (\textbf{51.12\%}) and is superior in representing both the relative directional relationships (\textbf{Avg. Dir. Sim.} of \textbf{43.57\%}) and the correct orientation of individual objects (\textbf{Avg. Facing Sim.} of \textbf{68.75\%}). In contrast, while \texttt{Plain-CGMap-FFR-Out} maintains higher validity and isomorphism, it lags behind in all three similarity metrics.
\end{itemize}

\begin{table}[h!]
\centering
\caption{
    Comparison of graph metrics for cognitive maps generated by different methods. The metrics evaluate the quality of the generated maps against the ground truth.
    \textbf{Valid Rate}: percentage of syntactically and semantically valid maps.
    \textbf{Isomorphism Rate}: percentage of maps that are structurally identical (isomorphic) to the ground truth, accounting for rotation.
    \textbf{Overall Sim. (Similarity)}: a weighted score combining directional and facing similarity ($S_{\text{overall}} = \alpha \cdot S_{\text{dir}} + (1-\alpha) \cdot S_{\text{face}}$).
    \textbf{Avg. Dir. Sim. (Average Directional Similarity)}: correctness of relative spatial relations between objects.
    \textbf{Avg. Facing Sim. (Average Facing Similarity)}: correctness of object orientations.
    All values are percentages (\%).
}
\vspace{1.5em}
\label{tab:graph_metrics_comparison}
\resizebox{\textwidth}{!}{%
\begin{tabular}{lccccc}
\toprule
\textbf{Method} & \textbf{Valid Rate} & \textbf{Isomorphism Rate} & \textbf{Overall Sim.} & \textbf{Avg. Dir. Sim.} & \textbf{Avg. Facing Sim.} \\
\midrule
\texttt{Aug-CGMap-FFR-Out} & 92.57 & 0.10 & 51.12 & 43.57 & 68.75 \\
\texttt{Plain-CGMap-FFR-Out} & 94.95 & 7.43 & 37.44 & 28.29 & 58.78 \\
\bottomrule
\end{tabular}%
}
\end{table}

\subsection{\iclr{Further Analysis on View Interpolation}}~\label{app:vi_frozen}
To rigorously assess the impact of view interpolation and ensure fair comparison, we conducted extensive additional experiments covering  comparisons with optimal interpolation settings and scaling laws across model sizes. Please note that due to limitations in the raw dataset, the rotation setting does not support view interpolation beyond five steps; consequently, the experiments with up to 7-step view interpolation presented here are conducted exclusively on the around and among settings.

\noindent\textbf{Comparison with Optimal View Interpolation.}
To ensure fairness, we compared our \texttt{Plain-CGMap} against the best possible performance of the View Interpolation (VI) baseline on Qwen2.5-VL-3B. As detailed in Table~\ref{tab:optimal_vi}, our method (\texttt{Plain-CGMap-FFR-Out}, 47.41\%) outperforms the baseline even at its peak performance (VI-1, 46.47\%). This confirms that the Cognitive Map provides a structural advantage over simply increasing visual frame density. Furthermore, adding standard Free-Form Reasoning (\texttt{FFR}) to interpolated views harms performance as density increases (dropping from 45.53\% to 40.77\%), indicating that the bottleneck lies in the perception ability to organize visual floods rather than reasoning capacity alone.

\begin{table}[h!]
    \centering
    \resizebox{\linewidth}{!}{
    \begin{tabular}{l|c|ccccccc}
    \toprule
    Method (Qwen2.5VL-3B) & VI-0 (Raw) & VI-1 & VI-2 & VI-3 & VI-4 & VI-5 & VI-6 & VI-7 \\
    \midrule
    \texttt{Plain-CGMap-FFR-Out} & \textbf{47.41} & 44.94 & 44.59 & 43.18 & 44.35 & 43.41 & 42.82 & 45.28 \\
    \texttt{RawQA} & 43.76 & \textbf{46.47} & 45.53 & 44.94 & 44.35 & 44.24 & 45.18 & 44.59 \\
    \texttt{FFR} & 45.53 & 43.71 & 43.83 & 41.65 & 41.00 & 40.36 & 40.89 & 40.77 \\
    \bottomrule
    \end{tabular}
    }
    \caption{Comparison with Optimal View Interpolation (Qwen2.5-VL-3B).}
    \label{tab:optimal_vi}
\end{table}

\noindent\textbf{Scaling Analysis: Does View Interpolation Scale?}
We extended our evaluation to larger models, including Qwen2.5-VL-7B, Qwen3-VL-8B, Qwen3-VL-235B, and GPT-5, to test if higher capacity naturally resolves interpolation issues.
Results for 7B/8B models (Table~\ref{tab:scaling_small}) show no consistent scaling law. While Qwen2.5-VL-7B benefits from density (peaking at VI-4), Qwen3-VL-8B exhibits unstable performance despite being architecturally advanced.

\begin{table}[h!]
    \centering
    \begin{tabular}{l|cccccccc}
    \toprule
    Config & VI-0 & VI-1 & VI-2 & VI-3 & VI-4 & VI-5 & VI-6 & VI-7 \\
    \midrule
    Qwen2.5-VL-7B & 37.80 & 34.90 & 35.60 & 45.30 & \textbf{47.40} & 46.50 & 46.80 & 46.80 \\
    Qwen3-VL-8B & 33.80 & 36.60 & \textbf{37.60} & 35.20 & 35.30 & 33.90 & 35.80 & 35.80 \\
    \bottomrule
    \end{tabular}
    \caption{Scaling Analysis on Qwen2.5-VL-7B and Qwen3-VL-8B.}
    \label{tab:scaling_small}
\end{table}

Moreover, for massive-scale models (Table~\ref{tab:scaling_large}), performance negatively correlates with view density. GPT-5 peaks at the sparse 1-frame setting (46.59\%) and declines to 42.35\% as density increases to 7 frames. Similarly, Qwen3-VL-235B drops to $\sim$36\% with interpolation. This suggests that without structured mapping, interpolation artifacts act as noise rather than useful signals, even for SOTA models.

The view interpolation data is publicly available in our dataset repository.

\begin{table}[h!]
    \centering
    \small
    \begin{tabular}{l|ccccccc}
    \toprule
    Model & VI-1 & VI-2 & VI-3 & VI-4 & VI-5 & VI-6 & VI-7 \\
    \midrule
    \textbf{GPT-5} & \textbf{46.59} & 45.18 & 44.24 & 44.59 & 42.59 & 43.53 & 42.35 \\
    \textbf{Qwen3-VL-235B} & \textbf{38.94} & 38.59 & 37.29 & 37.18 & 35.29 & 36.12 & 36.00 \\
    \bottomrule
    \end{tabular}
    \caption{Performance of Large-Scale Models (GPT-5, Qwen3-VL-235B) across view densities.}
    \label{tab:scaling_large}
\end{table}

\subsection{\iclr{Explicit Reasoning with Visual-of-Thought}}~\label{app:Explicit}
\paragraph{Explicit Visual Sketching via External Tools.}
While text-structured maps (e.g., JSON) introduce symbolic order, they fundamentally remain implicit token sequences. To bridge this gap, we investigated whether externalizing spatial map into \textit{explicit visual representations} could further enhance reasoning. Inspired by ViLaSR\citet{wu2025reinforcingspatialreasoningvisionlanguage}, we implemented a pipeline where the VLM uses an external plotting engine (Matplotlib) rather than generating text maps directly.

\begin{figure}[ht]
  \centering
  \begin{subfigure}{0.48\linewidth}
    \centering
    \includegraphics[width=\linewidth]{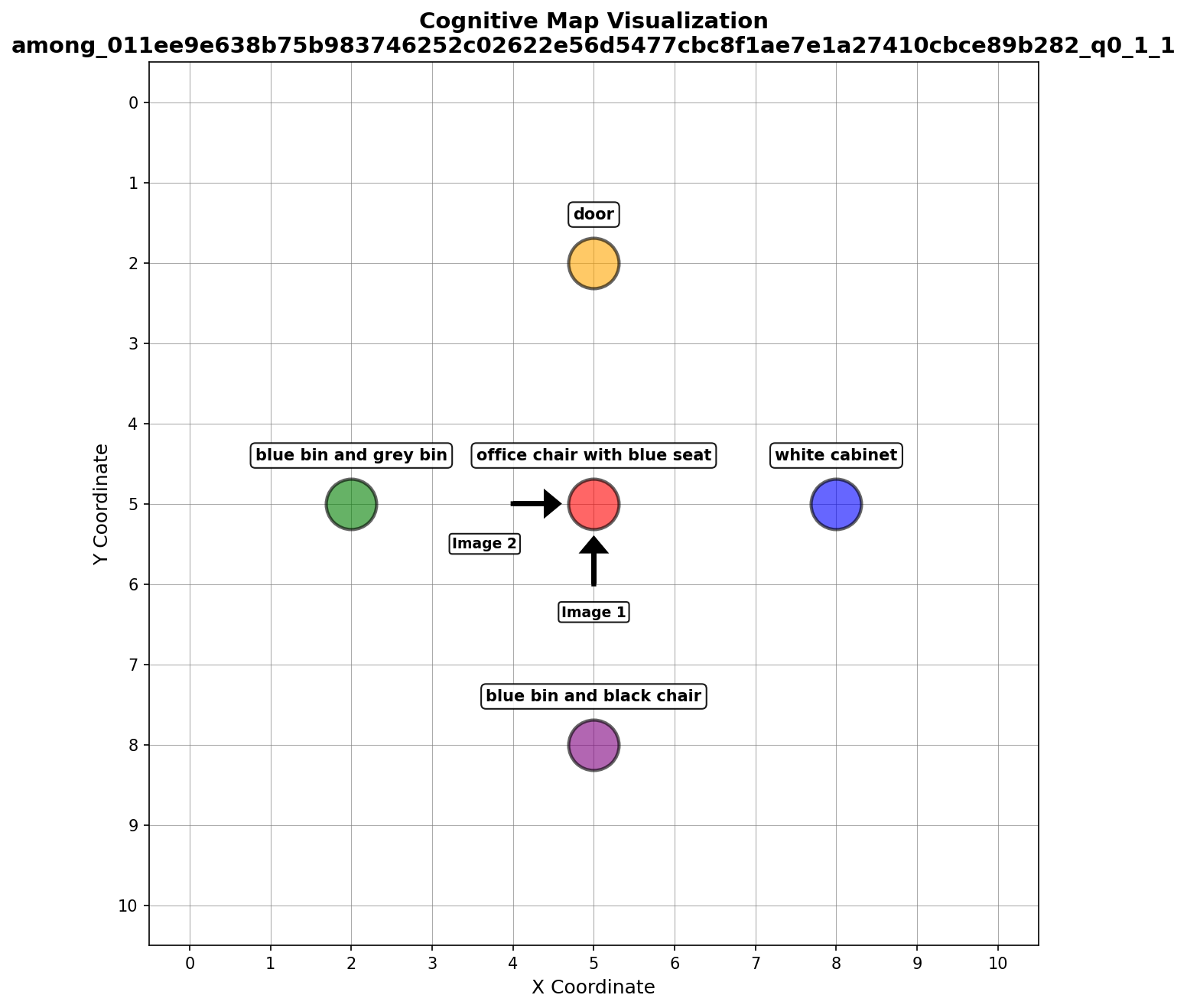}
    \caption{Explicit Visual Map in Among Setting.}
    \label{fig:map_text}
  \end{subfigure}
  \hfill
  \begin{subfigure}{0.48\linewidth}
    \centering
    \includegraphics[width=\linewidth]{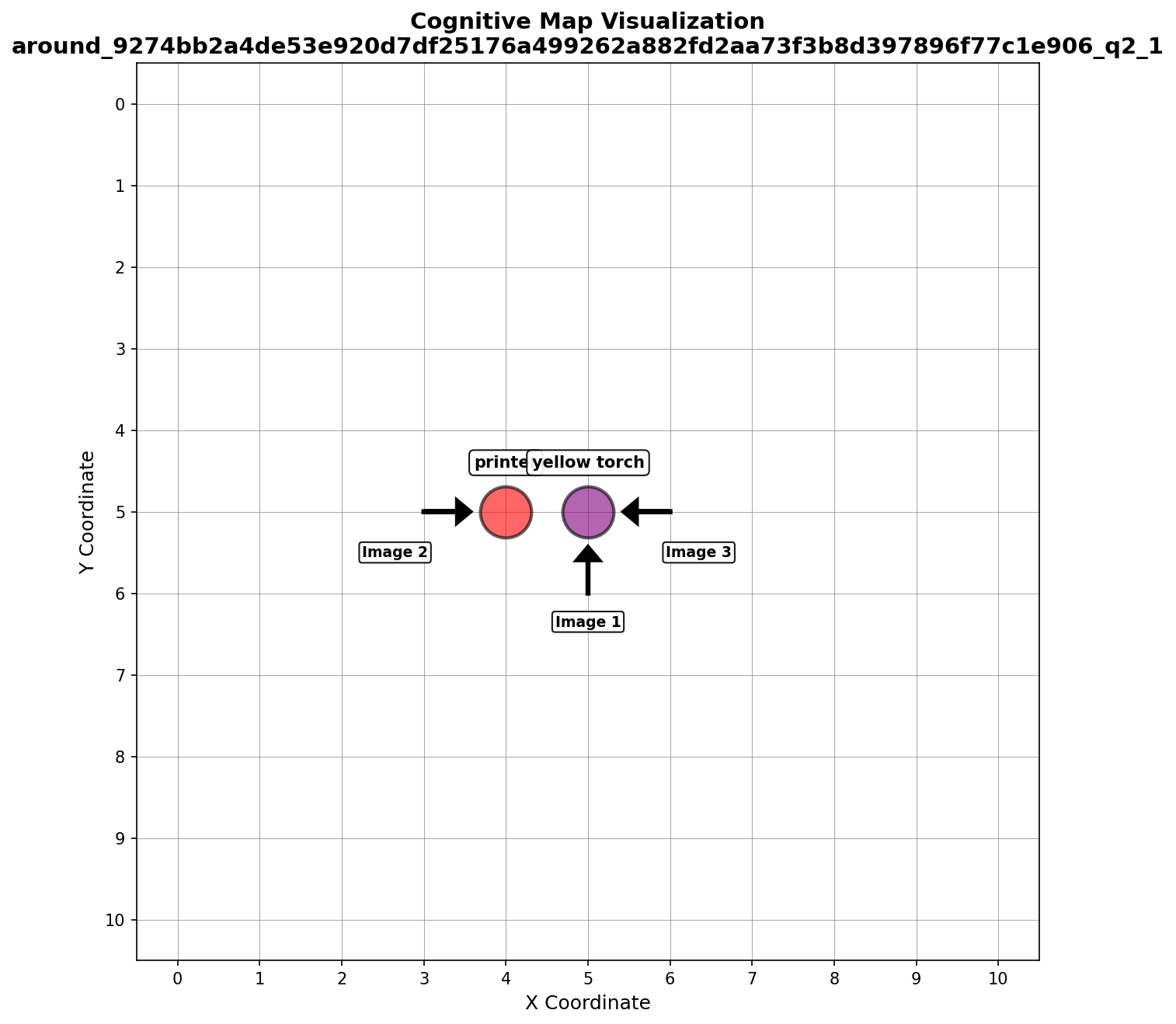}
    \caption{Explicit Visual Map in Around Setting.}
    \label{fig:map_visual}
  \end{subfigure}
  \caption{\textbf{Comparison of Map Representations.} The visual map, generated via external tools, renders objects and viewpoints onto a $10\times10$ grid, providing explicit geometric grounding.}
  \label{fig:map_comparison}
\end{figure}

We designed two visual-centric configurations:
\begin{enumerate}
    \item \textbf{Visual-Map-as-Input (\texttt{Img-CGMap-In}):} Instead of ingesting raw JSON tokens, the model receives a rendered $10\times10$ grid image where objects and viewpoints are plotted explicitly. This tests the model's ability to comprehend provided visual layouts.
    \item \textbf{Visual-Map-for-Reasoning (\texttt{Img-CGMap}):} We adopt a multi-turn tool-use framework. The model first acts as a coder to generate drawing coordinates (e.g., \texttt{(obj, [x,y])}), which are executed to render a visual sketch. This sketch is then fed back to the model as a new visual prompt to guide the final spatial reasoning. \textbf{The entire multi-turn interaction pipeline and prompt design were strictly aligned with the ViSaLR framework.}
\end{enumerate}

\begin{table}[h]
    \centering
    \resizebox{\linewidth}{!}{
    \begin{tabular}{l|c|c|ccc}
    \toprule
    \textbf{Configuration} & \textbf{Map Modality} & \textbf{Overall} & \textbf{Rotation} & \textbf{Among} & \textbf{Around} \\
    \midrule
    Raw-QA & - & 37.81 & 34.00 & 36.00 & 45.20 \\
    \midrule
    CGMap-In (Original) & Text (JSON) & 41.43 & \textbf{37.00} & 41.67 & 44.40 \\
    \textbf{Img-CGMap-In (New)} & Visual (Grid) & \textbf{42.10} & 31.50 & \textbf{44.17} & \textbf{45.60} \\
    \midrule
    Plain-CGMap (Original) & Text (JSON) & 41.33 & 25.00 & 39.67 & \textbf{58.40} \\
    \textbf{Img-CGMap (New)} & Visual (Grid) & \textbf{43.13} & \textbf{32.75} & \textbf{41.13} & 55.06 \\
    \bottomrule
    \end{tabular}
    }
    \caption{Performance comparison between implicit text-based maps (JSON) and explicit visual-based maps (Image/Grid). Visual configurations consistently outperform their textual counterparts.}
    \label{tab:visual_vs_text_map}
\end{table}

\noindent\textbf{Results Analysis.} As shown in Table~\ref{tab:visual_vs_text_map}, visual map configurations consistently outperform their textual counterparts. Specifically, \texttt{Img-CGMap} achieves an overall accuracy of \textbf{43.13\%}, surpassing the text-based \texttt{Plain-CGMap} (41.33\%). Notably, the ``Among'' spatial relation benefits significantly from the visual grid (41.13\% vs. 39.67\%), likely because relative positioning is more intuitive in pixel space than in coordinate space. This experiment validates that leveraging external tools to create explicit visual sketches grounds the model's reasoning more effectively than symbolic text, offering a promising direction for future development.

\section{Can We Teach VLMs to Build and Leverage Spatial Representations?}
In the main paper, we demonstrated that prompting frozen VLMs with external scaffolds offers limited improvements. This highlighted a core limitation: the models themselves aren't effectively forming internal spatial representations or reasoning through space. To address this, we investigated whether supervised fine-tuning (SFT) could teach VLMs to build and leverage these spatial models internally. This section of the appendix provides further details on our SFT methodology, starting with the crucial step of data curation.

\subsection{Supervised Fine-Tuning Data Curation}
Effective SFT heavily relies on the quality and nature of the training data. To teach our VLMs the desired spatial reasoning capabilities, we meticulously curated two primary types of data: cognitive maps and free-form reasoning chains. These were designed to provide strong supervisory signals for the model to learn how to represent and reason about space.

\subsubsection{Cognitive Map Generation} \label{app_sec:sft_cogmap_gen}
As discussed in Section~\ref{app_sec:data_structure}, cognitive maps serve as 2D schematic representations of object layouts. For the SFT phase, we needed to generate ground truth cognitive maps that the VLM could learn to produce. Our approach to generating these maps was grounded in the object arrangement annotations described in Section~\ref{app_sec:mindcube_collection}. We aimed for representations that were not only accurate but also in a format that the VLM could feasibly learn to generate.

The generation process was automated via a script that processes input JSONL files, where each line item contains scene details including images and, crucially, \texttt{meta\_info} describing the objects, their potential orientations, and the camera viewpoint setup. For every item, the script first identifies its specific spatial arrangement "setting" (e.g., "around," "among," "translation," or "rotation") by parsing the item's unique ID. Based on this setting, dedicated functions apply a set of predefined rules and heuristics to determine the 2D coordinates (on a 10x10 grid) and facing directions for both the objects and the camera views.

For instance, in the "around" setting, objects (typically 2-4) are placed in a predetermined linear arrangement near the grid's center (e.g., at coordinates like [4,5], [5,5]), and camera views are positioned at cardinal directions relative to these objects, based on the specific camera angles pertinent to the question. In the "rotation" setting, the camera is fixed at the center ([5,5]), and its facing direction changes across views, while object positions are defined relative to the camera's current orientation. Similar rule-based placements are implemented for "among" (objects in a cross or T-shape with views from specific angles) and "translation" (objects arranged linearly to depict relationships like "on" or "down to") settings. Object orientations, if applicable, are also assigned based on the input \texttt{meta\_info}.

Finally, the generated layout of objects and views is formatted into a structured JSON string, representing the cognitive map. This JSON cogmap, along with templated instructional prompts (\texttt{cogmap\_input} for VLM input format guidance and \texttt{cogmap\_output} for VLM output task description), is added to the original data item. The overall generation logic is summarized in Algorithm~\ref{alg:cogmap_generation}.

\begin{algorithm}[htbp]
\caption{Cognitive Map Generation} \label{alg:cogmap_generation}
\begin{algorithmic}[1]
\Require Dataset $D$ containing items with spatial arrangement annotations
\Ensure Updated dataset with cognitive maps in JSON format

\ForAll{$item \in D$}
    \State $setting \gets$ Extract setting type from $item.id$
    \State Initialize empty cognitive map $cogmap$
    
    \Comment{Position objects and views based on setting type}
    \If{$setting = \text{``around''}$}
        \State Position 2-4 objects in a line with coordinates like [4,5], [5,5], etc.
        \State Place views at cardinal positions based on camera angles
    \ElsIf{$setting = \text{``among''}$}
        \State Place center object at [5,5] and surrounding objects at [5,8], [2,5], [5,2], [8,5]
        \State Position views based on specified camera angles
    \ElsIf{$setting = \text{``translation''}$}
        \State Position objects according to their spatial relationships (e.g., ``on'', ``down'')
        \State Place views to highlight these spatial relationships
    \ElsIf{$setting = \text{``rotation''}$}
        \State Arrange objects based on rotation type (clockwise, counterclockwise, etc.)
        \State Fix camera at [5,5] with varying facing directions
    \EndIf
    
    \Comment{Add orientation information where applicable}
    \ForAll{$object \in cogmap.objects$}
        \If{$object$ has orientation}
            \State Add facing direction (``up'', ``down'', ``left'', ``right'')
        \EndIf
    \EndFor
    
    \State Format $cogmap$ as structured JSON
    \State Add formatted cognitive map to $item$
\EndFor
\State \Return Updated dataset $D$
\end{algorithmic}
\end{algorithm}

\subsubsection{Free-Form Reasoning Generation} \label{app_sec:sft_ffr_gen}
While cognitive maps provide a structured, global understanding of the scene, effective spatial reasoning also involves a procedural, step-by-step thought process. To instill this capability in our VLMs, we generated a dataset of grounded free-form reasoning chains. These chains were designed to verbalize the mental simulation process required to answer the spatial questions in \name.

The generation of these reasoning chains was closely tied to the question-answer (QA) templates developed in Section 2. For each specific setting (e.g., rotation, among, around), we manually constructed reasoning chains following a consistent set of principles to ensure logical coherence and clear grounding in the provided visual information and the question asked.

The core principles guiding the generation of these reasoning chains were:
\begin{enumerate}[leftmargin=12pt, topsep=-1pt, itemsep=0pt]
    \item \textbf{Initial Scene Understanding.} The reasoning begins by processing each input image individually. This involves identifying key objects visible in that view and noting their explicit spatial relationships with other objects within that same view. This step emulates the initial perceptual intake a human might perform.
    \item \textbf{Cross-View Consistency and Environment Integration.} After individual view analysis, the reasoning emphasizes that although different images are provided, they all depict the \textit{same underlying spatial environment}. This is often achieved by identifying and highlighting an anchor object or a consistent set of objects that appear across multiple views, thereby helping to establish a unified mental model of the scene.
    \item \textbf{Question-Driven Inference.} With a foundational understanding of the scene established from the views, the subsequent steps in the reasoning chain are directly guided by the specifics of the question. This involves: (1) \textbf{Mental Simulation:} If the question involves a hypothetical change in viewpoint or a "what-if" scenario (e.g., "what if you turn left?"), the reasoning chain explicitly verbalizes this mental transformation. (2) \textbf{Perspective Taking:} If the question requires adopting a different perspective (e.g., "from the sofa's perspective"), the reasoning chain articulates this shift. (3) \textbf{Spatial Relationship Deduction:} The chain logically deduces the queried spatial relationship by integrating information from the relevant views, applying spatial concepts (like left-of, behind, further from), and referencing the established mental model of the scene.
\end{enumerate}

\begin{figure}[htbp]
    \centering
    \includegraphics[width=\linewidth]{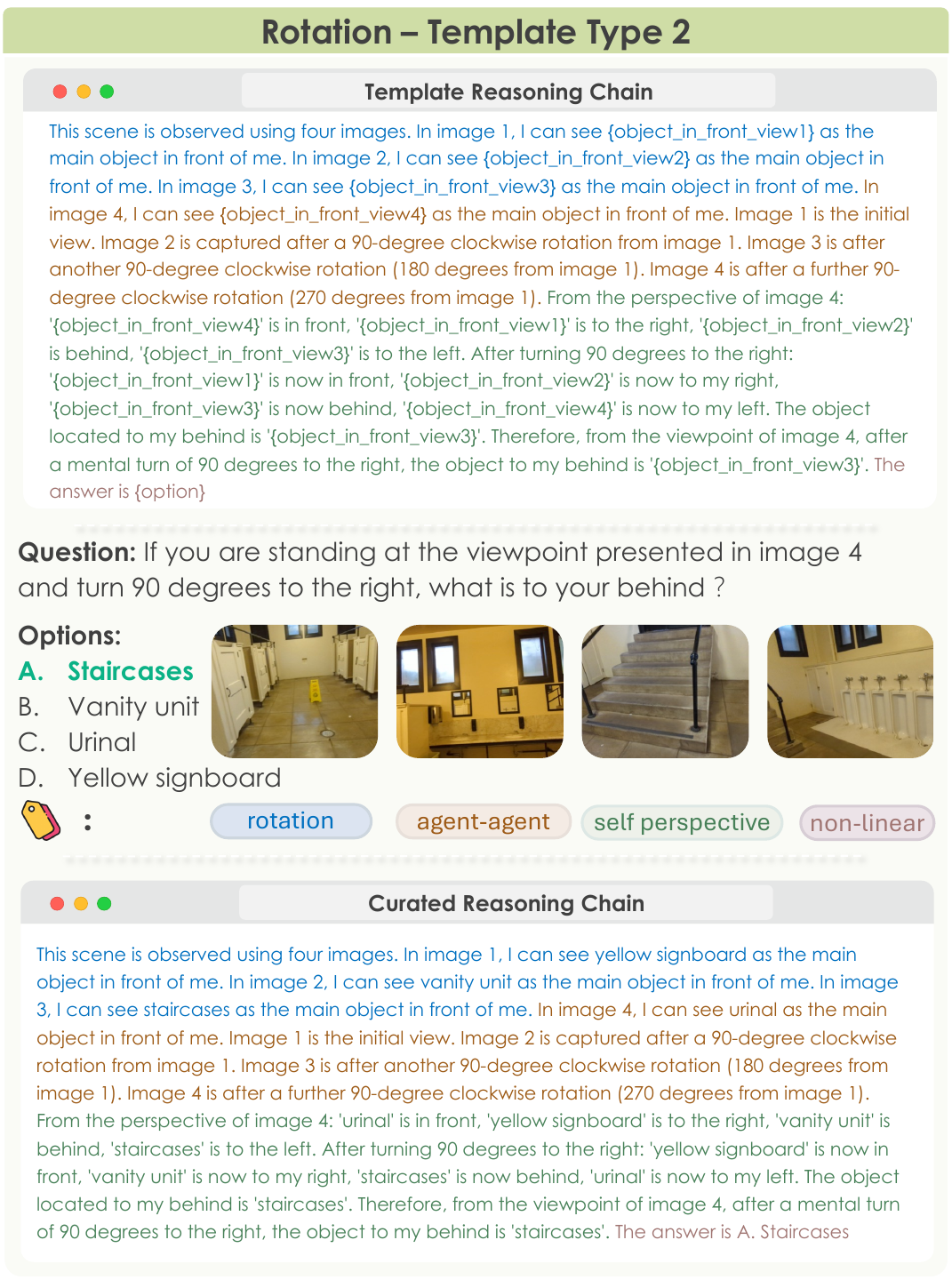}
    \caption{Example reasoning chain template for \textsc{Rotation}}
    \label{fig:reasoning_rotation}
\end{figure}

\begin{figure}[htbp]
    \centering
    \includegraphics[width=\linewidth]{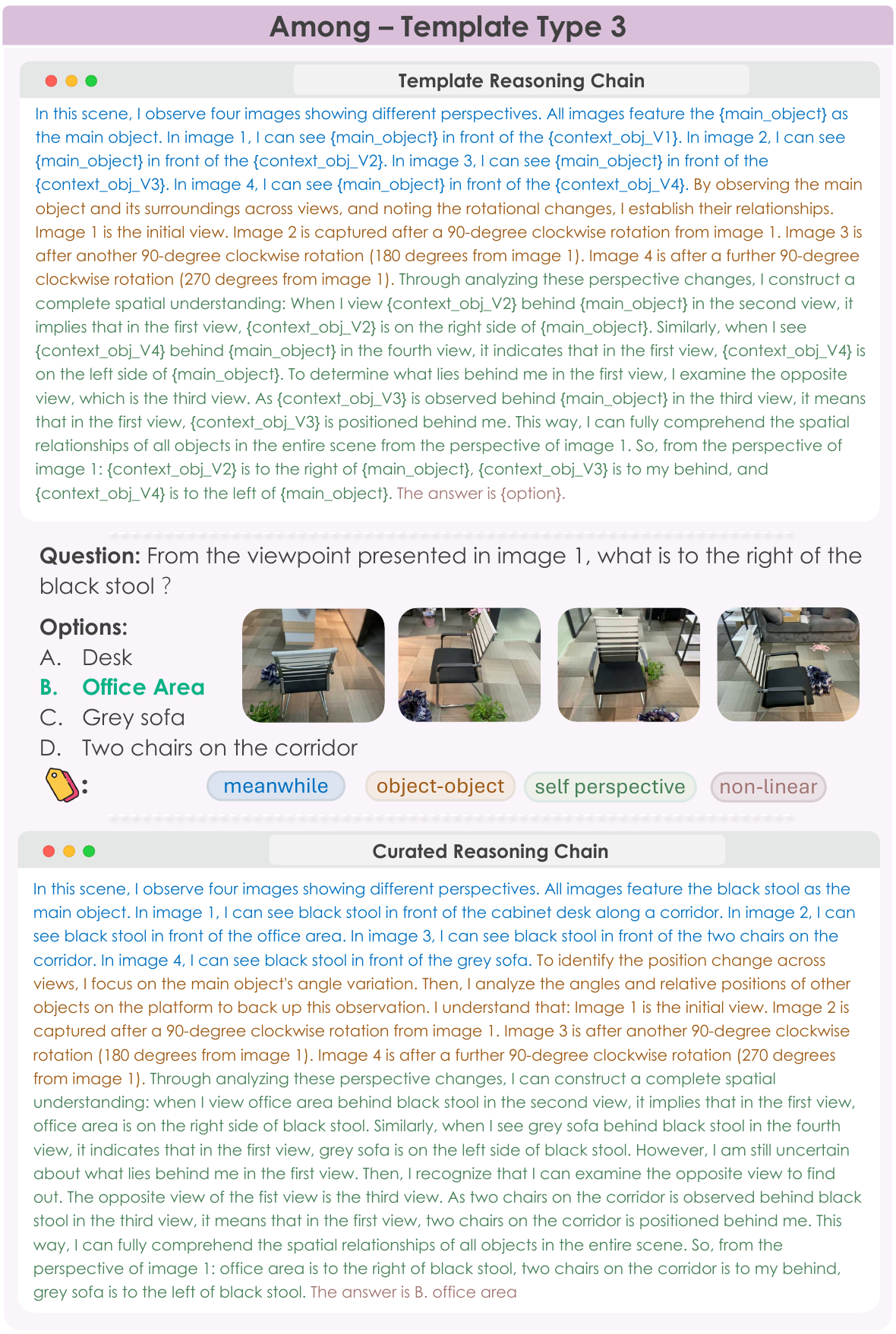}
    \caption{Example reasoning chain template for \textsc{Among}}
    \label{fig:reasoning_among}
\end{figure}

\begin{figure}[htbp]
    \centering
    \includegraphics[width=.92\linewidth]{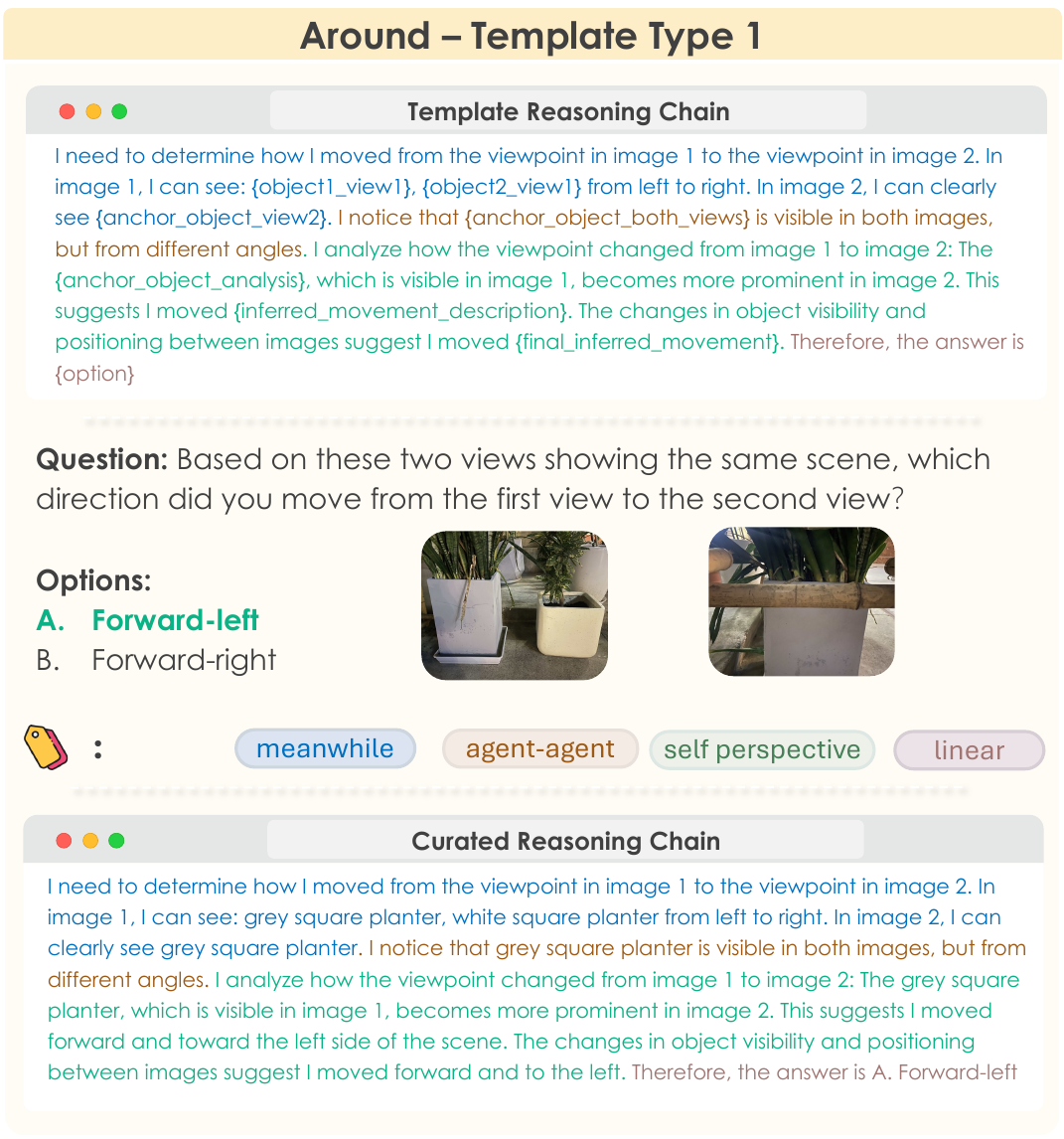}
    \caption{Example reasoning chain template for \textsc{Around}}
    \label{fig:reasoning_around}
    \vspace*{-1.5em}
\end{figure}

This structured approach to generating reasoning chains aimed to provide clear, step-by-step examples of spatial thought processes for the VLM to learn from. Figure~\ref{fig:reasoning_rotation},~\ref{fig:reasoning_among} and~\ref{fig:reasoning_around} show a template example combined with the filled case for \textsc{Rotation}, \textsc{Among}, \textsc{Around}, respectively.

\subsection{Detailed Experimental Setup} \label{app_sec:sft_exp_setup}

\begin{wraptable}{r}{0.5\textwidth}
\centering
\vspace{-1.5em} %
\caption{Training hyperparameters for SFT experiments with \modelname.}
\label{tab:sft_training_settings_appendix}
\begin{tabular}{ll}
\toprule
\textbf{Parameter}         & \textbf{Value}            \\
\midrule
Dataset size               & 10,000 QA pairs           \\
Epochs                     & 3                         \\
Learning rate              & 1e-5                      \\
Scheduler                  & Cosine                    \\
Fine-tuning type           & Full-parameter            \\
Batch Size       & 256 \\ 
GPUs used                  & 2 × NVIDIA H100  \\
Max image resolution       & 90,000 pixels             \\
Min image resolution       & 784 pixels                \\
Model Max Length           & 8192 tokens               \\
Weight Decay               & 0                         \\
Warmup Ratio               & 0.03                      \\
Max Grad Norm              & 1                         \\
Precision                  & BF16                      \\
Optimizer                  & AdamW      \\ %
\bottomrule
\end{tabular}
\vspace{-2ex} %
\end{wraptable}

In this section, we provide a more granular view of the experimental parameters employed during the Supervised Fine-Tuning (SFT) phase of our research. As stated in the main text, these experiments were designed to teach Vision-Language Models (VLMs) to build and leverage internal spatial representations. The base model for these SFT experiments was \modelname.

We utilized a consistent training script for all SFT experiments, ensuring comparability across different configurations. The primary variation across these runs was the specific dataset used (datasets variable in the script), corresponding to the different SFT task configurations discussed in Section 4.1, such as \texttt{Aug-CGMap-Out}. Other hyperparameters were kept constant to isolate the effects of the different training signals.

The core training hyperparameters are summarized in Table \ref{tab:sft_training_settings_appendix} and further detailed by the provided training script.


The training was conducted using a distributed setup managed by \texttt{torchrun} and leveraged DeepSpeed with a ZeRO Stage 3 optimization strategy for efficient full-parameter fine-tuning. Specifically, we set \texttt{NPROC\_PER\_NODE} to 2, utilizing two NVIDIA H100 GPUs, though the script template showed \texttt{CUDA\_VISIBLE\_DEVICES=0,1,2,3} and \texttt{NPROC\_PER\_NODE} defaulting to 4, our table and resource claims point to 2 GPUs being used for these runs. The \texttt{per\_device\_train\_batch\_size} was set to 4, and with \texttt{gradient\_accumulation\_steps} at 32, this resulted in an effective batch size of 256.

The learning rate was $1 \times 10^{-5}$ with a cosine learning rate scheduler and a warmup ratio of 0.03 over 3 training epochs. We enabled full fine-tuning of the vision encoder, MM-MLP, and the LLM components (\texttt{--tune\_mm\_vision True --tune\_mm\_mlp True --tune\_mm\_llm True}). The maximum sequence length for the model was set to 8192 tokens (\texttt{--model\_max\_length 8192}), and the maximum image input pixels were capped at 90,000 (\texttt{--max\_pixels 90000}), with a minimum of 784. Gradient checkpointing was enabled to manage memory usage (\texttt{--gradient\_checkpointing True}). Checkpoints were saved every 5 steps (\texttt{--save\_steps 5}), keeping a total of 12 checkpoints (\texttt{--save\_total\_limit 12}).

This consistent and robust experimental framework, detailed in Sections 3.1, 3.2, and herein, allowed for a fair comparison of the different SFT strategies aimed at improving the spatial reasoning capabilities of \modelname.

\subsection{VLM Response Examples After SFT for Configurations in  Section \ref{app_sec:example_configuration}}
\begin{tcolorbox}[colback=boxbackground, colframe=accentblue, 
title=Output for \texttt{Raw-QA}: Input Prompt Refer to \ref{app_sec:config_rawqa}, boxrule=0.5mm, width=\textwidth, arc=2mm, auto outer arc=true, breakable]
\textless answer\textgreater \textcolor{red}{\bf D. White headboard}\textless /answer\textgreater 
\end{tcolorbox}

\bigskip

\begin{tcolorbox}[colback=boxbackground, colframe=accentblue, 
title=Output for \texttt{Aug-CGMap-Out}: Input Prompt Refer to \ref{app_sec:config_ag_cgmap_out}, boxrule=0.5mm, width=\textwidth, arc=2mm, auto outer arc=true, breakable]
Based on my observation, the answer is:
\newline
\textless cogmap\textgreater 
\begin{verbatim}
```json
{
  "objects": [
    {"name": "white jar", "position": [5, 5]},
    {"name": "bed sheet with a floral pattern", 
        "position": [5, 8]},
    {"name": "white headboard", "position": [2, 5]},
    {"name": "clothes rack", "position": [5, 2]},
    {"name": "table with cups on it", "position": [8, 5]}
  ],
  "views": [
    {"name": "Image 1", "position": [5, 6], "facing": "up"},
    {"name": "Image 2", "position": [4, 5], "facing": "right"},
    {"name": "Image 3", "position": [5, 4], "facing": "down"},
    {"name": "Image 4", "position": [6, 5], "facing": "left"}
  ]
}
'''
\end{verbatim}
\textless /cogmap\textgreater 
\textless answer\textgreater \textcolor{mygreen}{\bf C. Bed sheet with a floral pattern}\textless /answer\textgreater 
\end{tcolorbox}

\bigskip

\begin{tcolorbox}[colback=boxbackground, colframe=accentblue, 
title=Output for \texttt{Plain-CGMap-Out}: Input Prompt Refer to \ref{app_sec:config_plain_cgmap_out}, boxrule=0.5mm, width=\textwidth, arc=2mm, auto outer arc=true, breakable]
Based on my observation, the answer is:
\newline
\textless cogmap\textgreater 
\begin{verbatim}
```json
{
    "white jar": {"position": [5, 5]}, 
    "bed sheet with a floral pattern": {"position": [5, 8]}, 
    "white headboard": {"position": [2, 5]}, 
    "clothes rack": {"position": [5, 2]}, 
    "table with cups on it": {"position": [8, 5]}
}
'''
\end{verbatim}
\textless /cogmap\textgreater \textless answer\textgreater 
\textcolor{red}{\bf B. Clothes rack}
\textless /answer\textgreater 
\end{tcolorbox}

\bigskip

\begin{tcolorbox}[colback=boxbackground, colframe=accentblue, 
title=Output for \texttt{\iclr{FFR}}: Input Prompt Refer to \ref{app_sec:config_ff_rsn}, boxrule=0.5mm, width=\textwidth, arc=2mm, auto outer arc=true, breakable]
\textless think\textgreater In this scene, I observe four images showing different perspectives. All images feature the white jar as the main object. In image 1, I can see white jar in front of the table with cups on it. In image 2, I can see white jar in front of the bed sheet with a floral pattern. In image 3, I can see white jar in front of the clothes rack. In image 4, I can see white jar in front of the white headboard. To identify the position change across views, I focus on the main object's angle variation. Then, I analyze the angles and relative positions of other objects on the platform to back up this observation. I understand that:  Image 1 is the initial view. Image 2 is captured after a 90-degree clockwise rotation from image 1. Image 3 is after another 90-degree clockwise rotation (180 degrees from image 1). Image 4 is after a further 90-degree clockwise rotation (270 degrees from image 1). Through analyzing these perspective changes, I can construct a complete spatial understanding: when I view bed sheet with a floral pattern behind white jar in the second view, it implies that in the first view, bed sheet with a floral pattern is on the right side of white jar. Similarly, when I see white headboard behind white jar in the fourth view, it indicates that in the first view, white headboard is on the left side of white jar. However, I am still uncertain about what lies behind me in the first view. Then, I recognize that I can examine the opposite view to find out. The opposite view of the fist view is the third view. As clothes rack is observed behind white jar in the third view, it means that in the first view, clothes rack is positioned behind me. This way, I can fully comprehend the spatial relationships of all objects in the entire scene. So, from the perspective of image 4: table with cups on it is to the right of white jar, bed sheet with a floral pattern is to my behind, clothes rack is to the left of white jar. So the answer is B. Clothes rack\textless /think\textgreater \textless answer\textgreater \textcolor{red}{\bf B. Clothes rack}\textless /answer\textgreater 
\end{tcolorbox}

\bigskip

\begin{tcolorbox}[colback=boxbackground, colframe=accentblue, 
title=Output for \texttt{Aug-CGMap-FFR-Out}: Input Prompt Refer to \ref{app_sec:config_ag_cgmap_ffr_out}, boxrule=0.5mm, width=\textwidth, arc=2mm, auto outer arc=true, breakable]
Based on my observation, the answer is:
\newline
\textless cogmap\textgreater 
\newline
\begin{verbatim}
```json
{
  "objects": [
    {"name": "white jar", "position": [5, 5]},
    {"name": "bed sheet with a floral pattern", 
        "position": [5, 8]},
    {"name": "white headboard", "position": [2, 5]},
    {"name": "clothes rack", "position": [5, 2]},
    {"name": "table with cups on it", "position": [8, 5]}
  ],
  "views": [
    {"name": "Image 1", "position": [5, 6], "facing": "up"},
    {"name": "Image 2", "position": [4, 5], "facing": "right"},
    {"name": "Image 3", "position": [5, 4], "facing": "down"},
    {"name": "Image 4", "position": [6, 5], "facing": "left"}
  ]
}
'''
\end{verbatim}
\textless /cogmap\textgreater 
\textless think\textgreater In this scene, I observe four images showing different perspectives. All images feature the white jar as the main object. In image 1, I can see white jar in front of the table with cups on it. In image 2, I can see white jar in front of the clothes rack. In image 3, I can see white jar in front of the bed sheet with a floral pattern. In image 4, I can see white jar in front of the white headboard. To identify the position change across views, I focus on the main object's angle variation. Then, I analyze the angles and relative positions of other objects on the platform to back up this observation. I understand that:  Image 1 is the initial view. Image 2 is captured after a 90-degree clockwise rotation from image 1. Image 3 is after another 90-degree clockwise rotation (180 degrees from image 1). Image 4 is after a further 90-degree clockwise rotation (270 degrees from image 1). Through analyzing these perspective changes, I can construct a complete spatial understanding: when I view clothes rack behind white jar in the second view, it implies that in the first view, clothes rack is on the right side of white jar. Similarly, when I see white headboard behind white jar in the fourth view, it indicates that in the first view, white headboard is on the left side of white jar. However, I am still uncertain about what lies behind me in the first view. Then, I recognize that I can examine the opposite view to find out. The opposite view of the fist view is the third view. As bed sheet with a floral pattern is observed behind white jar in the third view, it means that in the first view, bed sheet with a floral pattern is positioned behind me. This way, I can fully comprehend the spatial relationships of all objects in the entire scene. So, from the perspective of image 4: clothes rack is to the right of white jar, bed sheet with a floral pattern is to my behind, table with cups on it is to the left of white jar. So the answer is A. Table with cups on it.\textless /think\textgreater \textless answer\textgreater \textcolor{red}{\bf A. Table with cups on it}\textless /answer\textgreater 
\end{tcolorbox}

\bigskip

\begin{tcolorbox}[colback=boxbackground, colframe=accentblue, 
title=Output for \texttt{Plain-CGMap-FFR-Out}: Input Prompt Refer to \ref{app_sec:config_plain_cgmap_ffr_out}, boxrule=0.5mm, width=\textwidth, arc=2mm, auto outer arc=true, breakable]
Based on my observation, the answer is:
\newline
\textless cogmap\textgreater 
\newline
\begin{verbatim}
```json
{
    "white jar": {"position": [5, 5]}, 
    "bed sheet with a floral pattern": {"position": [5, 8]}, 
    "white headboard": {"position": [2, 5]}, 
    "clothes rack": {"position": [5, 2]}, 
    "table with cups on it": {"position": [8, 5]}
}
'''
\end{verbatim}
\textless /cogmap\textgreater 
\textless think\textgreater In this scene, I observe four images showing different perspectives. All images feature the white jar as the main object. In image 1, I can see white jar in front of the table with cups on it. In image 2, I can see white jar in front of the clothes rack. In image 3, I can see white jar in front of the bed sheet with a floral pattern. In image 4, I can see white jar in front of the white headboard. To identify the position change across views, I focus on the main object's angle variation. Then, I analyze the angles and relative positions of other objects on the platform to back up this observation. I understand that:  Image 1 is the initial view. Image 2 is captured after a 90-degree clockwise rotation from image 1. Image 3 is after another 90-degree clockwise rotation (180 degrees from image 1). Image 4 is after a further 90-degree clockwise rotation (270 degrees from image 1). Through analyzing these perspective changes, I can construct a complete spatial understanding: when I view clothes rack behind white jar in the second view, it implies that in the first view, clothes rack is on the right side of white jar. Similarly, when I see white headboard behind white jar in the fourth view, it indicates that in the first view, white headboard is on the left side of white jar. However, I am still uncertain about what lies behind me in the first view. Then, I recognize that I can examine the opposite view to find out. The opposite view of the fist view is the third view. As bed sheet with a floral pattern is observed behind white jar in the third view, it means that in the first view, bed sheet with a floral pattern is positioned behind me. This way, I can fully comprehend the spatial relationships of all objects in the entire scene. So, from the perspective of image 4: clothes rack is to the right of white jar, bed sheet with a floral pattern is to my behind, table with cups on it is to the left of white jar. So the answer is A. Table with cups on it.\textless /think\textgreater \textless answer\textgreater \textcolor{red}{\bf A. Table with cups on it}\textless /answer\textgreater 
\end{tcolorbox}

\subsection{Detailed Graph Metric Results for SFT Graph-Related Experiments}

This section provides a detailed look at the Supervised Fine-Tuning (SFT) training dynamics to support the main paper's conclusions. The figures below plot key metrics over training steps for four map-generation settings. A comparative analysis highlights that jointly training map generation and reasoning is the most effective strategy.

When training on map generation alone, as in the \texttt{Plain-CGMap-Out} and \texttt{Aug-CGMap-Out} settings, the graph quality metrics show rapid convergence. However, the final QA accuracy is limited, reaching 54.29\% for \texttt{Plain-CGMap-Out} and 52.48\% for \texttt{Aug-CGMap-Out}.

In contrast, the joint training approaches (\texttt{Plain-CGMap-FFR-Out} and \texttt{Aug-CGMap-FFR-Out}), despite a slower initial convergence on graph quality metrics, ultimately achieve far superior performance in task accuracy. The \texttt{Plain-CGMap-FFR-Out} setting proves to be the most effective, reaching a QA Accuracy of 57.81\%. The \texttt{Aug-CGMap-FFR-Out} setting also yields results with QA accuracy climbing to about 54\%. This demonstrates the superiority of joint training for achieving the best task accuracy, even though the generated maps are less structurally perfect than those from map-only training.

\begin{figure}[!ht]
    \centering
    \includegraphics[width=\linewidth]{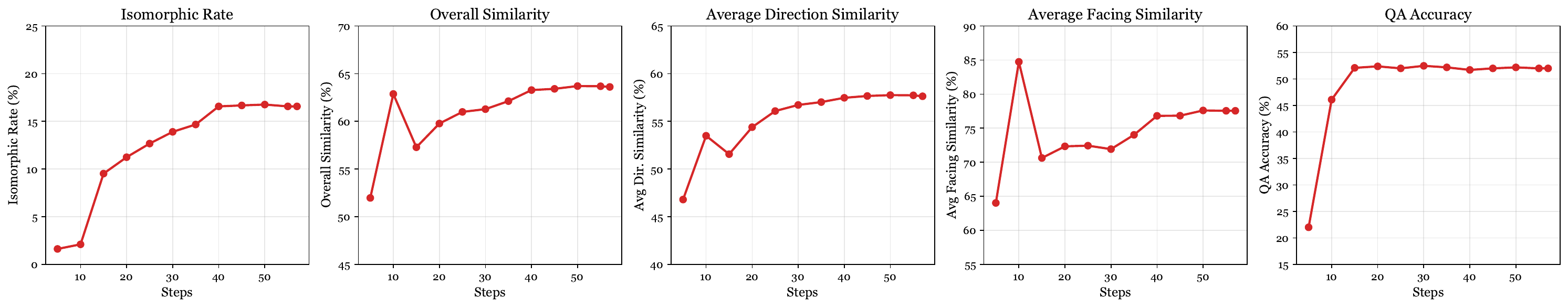}
    \caption{Training dynamics for the \texttt{Aug-CGMap-Out} setting.}
    \label{fig:sft-aug-cgmap-out}
\end{figure}

\begin{figure}[!ht]
    \centering
    \includegraphics[width=\linewidth]{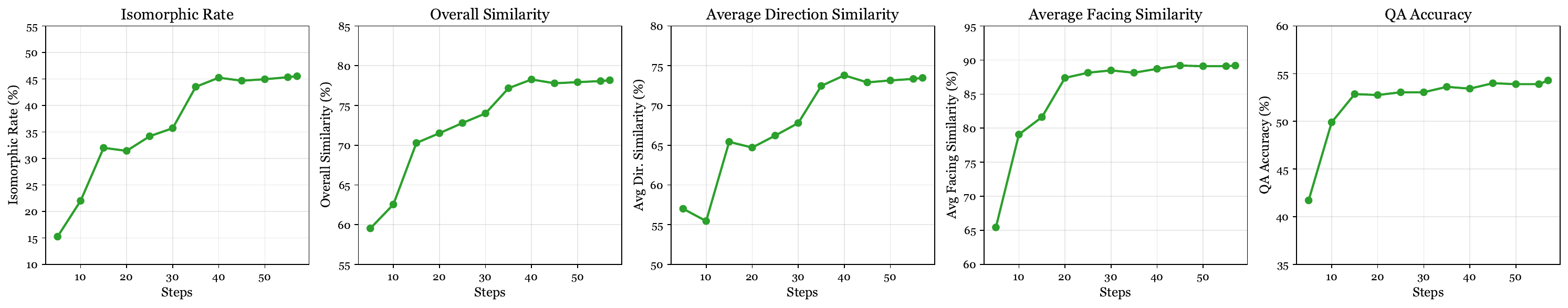}
    \caption{Training dynamics for the \texttt{Plain-CGMap-Out} setting.}
    \label{fig:sft-plain-cgmap-out}
\end{figure}

\begin{figure}[!ht]
    \centering
    \includegraphics[width=\linewidth]{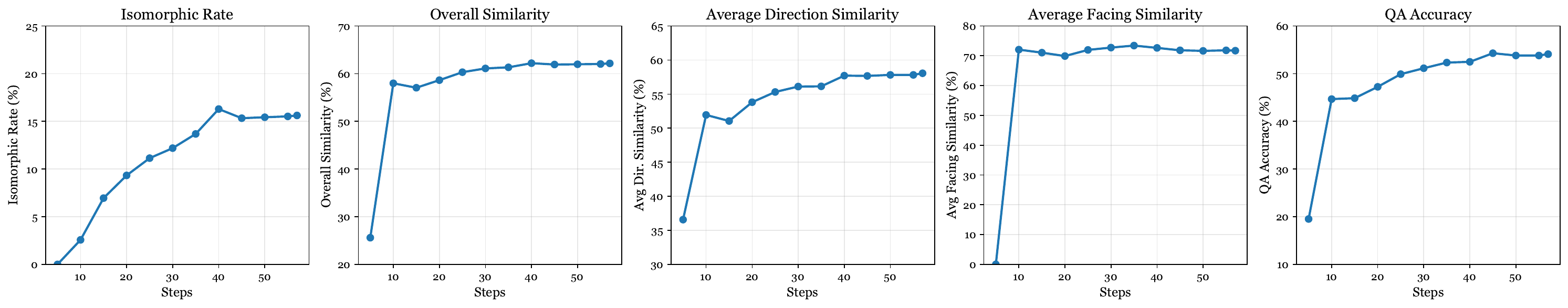}
    \caption{Training dynamics for the \texttt{Aug-CGMap-FFR-Out} setting.}
    \label{fig:sft-aug-cgmap-ffr-out}
\end{figure}

\begin{figure}[!ht]
    \centering
    \includegraphics[width=\linewidth]{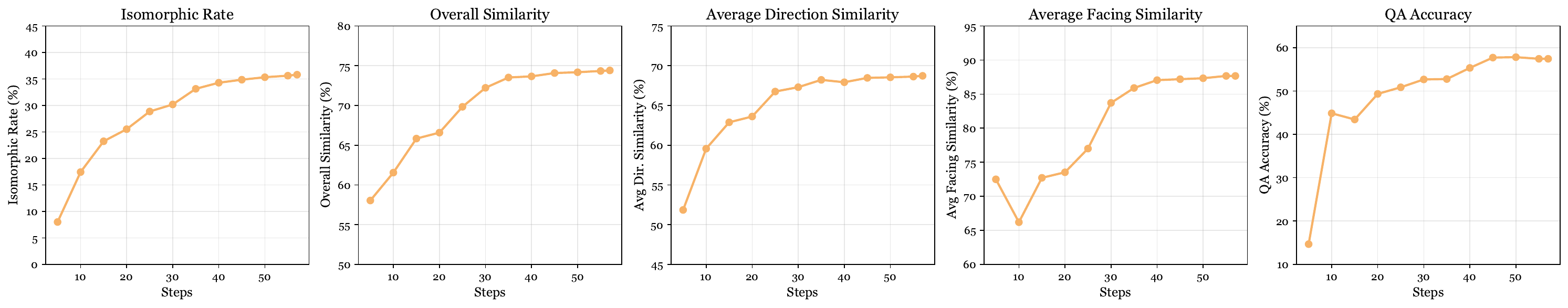}
    \caption{Training dynamics for the \texttt{Plain-CGMap-FFR-Out} setting, showing superior final performance.}
    \label{fig:sft-plain-cgmap-ffr-out}
\end{figure}

\subsection{Which Part of VLM is the Bottleneck for Spatial Understanding?}

To develop more efficient fine-tuning strategies, it is crucial to understand which component of a Vision-Language Model (VLM)---the vision encoder responsible for perception or the Large Language Model (LLM) responsible for reasoning---presents the primary bottleneck for spatial understanding. To investigate this, we conduct a bottleneck analysis by selectively fine-tuning different parts of the VLM and observing the impact on performance.

We evaluate four distinct training configurations on the \texttt{Raw-QA} task, with results captured at an early stage of training (step 57) to assess the initial learning dynamics. The configurations are: (1) the baseline performance of the pre-trained model without any fine-tuning; (2) fine-tuning only the vision encoder while keeping the LLM frozen; (3) fine-tuning only the LLM while keeping the vision encoder frozen; and (4) the standard approach of fine-tuning all parts of the model.

\begin{table}[htbp]
\centering
\caption{VLM Training Bottleneck Analysis (Step=57, in \%). Performance is measured on the \namesub benchmark under the Raw-QA setting.}
\label{tab:vlm_bottleneck}
\vspace{1.5em}
\begin{tabular}{lcccc}
\toprule
\textbf{Training Method} & \textbf{Overall} & \textbf{Rotation} & \textbf{Among} & \textbf{Around} \\
\midrule
Raw-QA (no fine-tuning) & 37.81 & 34.00 & 36.00 & 45.20 \\
Freeze LLM (Vision Encoder Only) & 37.81 & 30.50 & 37.00 & 45.60 \\
Freeze Vision Encoder (LLM Only) & 51.43 & 34.00 & 50.00 & 68.80 \\
Tune All Parts & 52.67 & 34.50 & 52.50 & 67.60 \\
\bottomrule
\end{tabular}
\end{table}

The results, presented in Table~\ref{tab:vlm_bottleneck}, offer several key insights. First, there is a dramatic performance leap from the no-fine-tuning baseline (37.81\% overall), but only when the language model is trained. Methods involving LLM fine-tuning achieve over 51\% accuracy, underscoring the necessity of adapting the model's reasoning capabilities.

Most strikingly, the performance bottleneck is almost exclusively concentrated in the LLM. Tuning only the LLM (\texttt{Freeze Vision Encoder}) yields an overall accuracy of 51.43\%, capturing nearly the full performance gain of end-to-end fine-tuning (52.67\%). In stark contrast, tuning only the vision encoder (\texttt{Freeze LLM}) provides no improvement whatsoever over the baseline (37.81\%). This indicates that the bottleneck is not shared between modules. For this spatial task, adapting the model's language-based reasoning is critical, while adapting its visual perception is surprisingly ineffective.

Intriguingly, the fact that fine-tuning only the vision encoder fails to improve performance is in itself a significant finding. A possible explanation is that the pre-trained visual features are already sufficient to extract the necessary objects and their properties. The core challenge of the task seems to lie not in \emph{what} is seen, but in \emph{how to reason} about the spatial relationships across a series of views—a task primarily handled by the LLM. In conclusion, our analysis suggests that the most significant gains come from adapting the reasoning module. For efficient tuning, freezing the vision encoder and focusing solely on the LLM proves to be a highly effective strategy, achieving nearly top-tier performance at a fraction of the computational cost.

\subsection{Branching from \texttt{Raw-QA} SFT Checkpoint}

In our main experiments, we fine-tuned the model for each specific task format starting from the base pre-trained VLM. A natural question arises: can a curriculum-based SFT approach further improve performance? Specifically, we investigate whether first fine-tuning the model on the simplest task format---`Raw-QA`, which only requires outputting the final answer---can establish a better foundation for learning to leverage more complex reasoning formats.

To test this hypothesis, we conducted a set of branching experiments. We took the checkpoint from the model fully fine-tuned on the `Raw-QA` task. Then, we used this specialized checkpoint as the initial weights for further fine-tuning on other scaffolding tasks, namely \texttt{Aug-CGMap-In}, \texttt{FF Rsn}, and \texttt{Aug-CGMap-FFR-Out}. It is important to note that during this second stage of fine-tuning, the model's output for all tasks was still constrained to be only the final answer option. This setup allows us to isolate the effect of the cognitive scaffolds on the model's internal reasoning process, rather than its ability to generate complex text.

The results, presented in Table~\ref{tab:sft_from_rawqa}, show a consistent and notable improvement across all branched tasks compared to their counterparts trained from scratch. 
For example, both \texttt{Aug-CGMap-In} and \texttt{Aug-CGMap-FFR-Out} reach an impressive overall accuracy of 49.00\%. Even the \texttt{FF Rsn} method benefits from this two-stage approach, with its overall accuracy rising to 46.82\%. These findings suggest that a two-stage SFT strategy is highly effective. By first grounding the model in the fundamental objective of the task (i.e., finding the correct answer) and then teaching it to process and leverage more complex cognitive scaffolds, we can achieve superior spatial reasoning performance. This indicates that the model, once primed for the core task, becomes more adept at utilizing the provided spatial context, even if it does not explicitly generate the reasoning chain or cognitive map.

\begin{table}[h!]
\centering
\caption{Performance of various methods after being fine-tuned from a \texttt{Raw-QA} SFT checkpoint. This two-stage training approach led to performance gains across all methods. All accuracies are reported as percentages (\%).}
\label{tab:sft_from_rawqa}
\vspace{1.5em}
\begin{tabular}{lcccc}
\toprule
\textbf{Method} & \textbf{Overall} & \textbf{Rotation} & \textbf{Among} & \textbf{Around} \\
\midrule
\texttt{Raw-QA}                  & 46.36 & 33.50 & 51.20 & 46.75 \\
\texttt{Aug-CGMap-In}            & 49.00 & 35.50 & 53.20 & 50.50 \\
\texttt{FF Rsn}                  & 46.82 & 37.00 & 50.60 & 47.00 \\
\texttt{Aug-CGMap-FFR-Out}       & 49.00 & 37.00 & 53.20 & 49.75 \\
\bottomrule
\end{tabular}
\end{table}

\subsection{Hyperparameter Tuning Results}

We conducted new hyperparameter tuning experiments to further validate our approach. Specifically, we performed a series of experiments to tune the hyperparameters for our Supervised Fine-Tuning (SFT) settings, as detailed in Table\ref{tab:sft_hyperparams}.

As shown in the table, our original hyperparameter configuration of a learning rate of $10^{-5}$, a batch size of 512, and a warmup ratio of 0.03 yielded the highest accuracy of 52.67\%. The results from these experiments confirm that the hyperparameters used in our initial submission are effective and well-optimized for the task, further substantiating our claims.
\begin{table}[h]
\centering
\caption{SFT Hyperparameter Tuning Results}
\label{tab:sft_hyperparams}
\begin{tabular}{lc}
\toprule
\textbf{SFT (learning rate, batch size, warmup ratio)} & \textbf{Acc (\%)} \\
\midrule
($1\!\times\!10^{-5}$, 512, 0.03) -- Ours & \textbf{52.67} \\
($2\!\times\!10^{-5}$, 512, 0.03)       & 51.71 \\
($4\!\times\!10^{-5}$, 512, 0.03)       & 51.52 \\
($1\!\times\!10^{-5}$, 256, 0.03)       & 50.86 \\
($1\!\times\!10^{-5}$, 1024, 0.03)      & 51.90 \\
($1\!\times\!10^{-5}$, 512, 0.01)       & 51.81 \\
($1\!\times\!10^{-5}$, 512, 0.10)       & 50.67 \\
\bottomrule
\end{tabular}
\end{table}

\subsection{Effect of Removing Viewpoint Descriptors}
\label{app:view_prior}

\textbf{Motivation.} In the original formulation of \textit{\modelname} questions, we include coarse viewpoint descriptors (e.g., ``back'', ``left'', ``right'') in the textual prompt to indicate that the input images correspond to canonical, orthogonal views rather than arbitrary camera poses. This design choice was initially made for clarity and simplicity, especially given that earlier-generation VLMs exhibited near-random performance and were unable to reliably exploit such directional cues.

However, as model capabilities improve, these textual descriptors may introduce unintended shortcuts: models could partially infer spatial relations from language alone without fully aligning multi-view visual observations. To assess this effect, we construct a variant of our evaluation setting where all viewpoint descriptors are removed from the question text, forcing models to rely purely on visual consistency across views. We evaluate this variant under both frozen inference and supervised fine-tuning (SFT) settings, while keeping all other components unchanged.

\textbf{Results.} As shown in Table~\ref{tab:v2_sft} and Table~\ref{tab:v2_froz}, the overall trends from the main paper are preserved under both frozen and SFT settings. In the frozen setting, the map-then-reason configuration (\texttt{Plain-CGMap-FFR-Out}) achieves the highest overall accuracy (42.67\%), outperforming \texttt{Raw-QA} (38.57\%) by 4.1 points. Notably, this relative gain is more pronounced than in the setting with viewpoint descriptors, suggesting that explicit spatial scaffolding plays a more critical role when textual shortcuts are unavailable.

Under SFT (Table~\ref{tab:v2_sft}), all configurations converge to a narrower performance range, with \texttt{Plain-CGMap-FFR-Out} maintaining a slight edge (best: 54.0\%) over \texttt{Raw-QA} (best: 52.8\%). The reduced gap suggests that without viewpoint descriptors, the task places greater demands on fine-grained visual alignment, making it harder for models to learn robust spatial representations from limited supervised data alone. Nevertheless, the map-then-reason approach remains the top-performing configuration across both evaluation settings, consistent with our main findings.

\begin{table}[htbp]
\centering
\caption{QA accuracy (\%) of \textit{\modelname} on \namesub \textbf{without viewpoint descriptors} under different configurations for frozen VLMs.}
\label{tab:v2_froz}
\begin{tabular}{lcccc}
\toprule
\textbf{Config.} & \textbf{Overall} & \textbf{Rotation} & \textbf{Among} & \textbf{Around} \\
\midrule
\texttt{Raw-QA}              & 38.57          & 34.00          & 36.83          & 46.40          \\
\midrule
\texttt{Aug-CGMap-In}        & 32.00\down & 34.50      & 30.50          & 33.60          \\
\texttt{FFR}                 & 39.14\up & 32.50          & 33.83          & 57.20          \\
\texttt{Aug-CGMap-FFR-Out}   & 38.95\up & 19.50          & \textbf{41.00} & 49.60          \\
\texttt{Plain-CGMap-FFR-Out} & \textbf{42.67}\up & 28.50          & 39.83          & \textbf{60.80} \\
\texttt{CGMap-In-FFR-Out}    & 40.67\up & \textbf{34.50} & \textbf{42.67}          & 40.80          \\
\bottomrule
\end{tabular}
\end{table}

\begin{table}[t]
\centering
\caption{Best QA accuracy (\%) across SFT training steps \textbf{without viewpoint descriptors}. Each value reports the peak accuracy observed during training (up to step 57).}
\label{tab:v2_sft}
\resizebox{\columnwidth}{!}{
\begin{tabular}{lccccccc}
\toprule
& \texttt{Raw-QA} & \texttt{Aug-CGMap-Out} & \texttt{Plain-CGMap-Out} & \texttt{FFR} & \texttt{Aug-CGMap-FFR-Out} & \texttt{Plain-CGMap-FFR-Out} \\
\midrule
\textbf{Best Acc.} & 52.8 & 51.6 & 52.8 & 52.2 & 53.8 & \textbf{54.0} \\
\textbf{@ Step} & 45 & 50 & 45 & 55 & 20 & 25 \\
\bottomrule
\end{tabular}
}
\end{table}

\section{Can Reinforcement Learning Further Refine Spatial Thought Processes?}
As discussed in the main paper, while Supervised Fine-Tuning (SFT) establishes a strong foundation for spatial reasoning, reinforcement learning (RL) presents an avenue for further optimizing spatial thought processes through outcome-driven feedback. The core inquiry is whether guiding VLMs with rewards can lead to the development of more precise spatial mental models and enhanced reasoning capabilities. This section of the appendix provides a more detailed exposition of the experimental setup employed for the RL phase of our research. Additionally, we present case studies to offer qualitative insights into how RL refines the models' spatial representations and reasoning chains.

\subsection{Detailed Experimental Setup} \label{app_sec:rl_exp_setup}
For the reinforcement learning (RL) phase of our research, we employed the VAGEN framework. The core policy optimization algorithm used was Group Relative Policy Optimization (GRPO). To ensure consistency and allow for direct comparison with earlier stages of our work, key components from the Supervised Fine-Tuning (SFT) experiments were retained. Specifically, the base Vision-Language Model (VLM) for all RL configurations was Qwen2.5-VL-3B-Instruct, and evaluations were performed on the MINDCUBE-TINY benchmark. All previously established evaluation metrics were also retained.

In consideration of computational costs, each distinct RL configuration was trained for a duration of 0.5 epoch. The primary hyperparameters governing the RL training process were set as follows:
\begin{itemize}[leftmargin=12pt, topsep=-1pt, itemsep=0pt]
    \item \textbf{Training Batch Size:} 32
    \item \textbf{Maximum Prompt Length:} 1024 tokens
    \item \textbf{Maximum Response Length:} 512 tokens
    \item \textbf{Actor Learning Rate:} $1 \times 10^{-6}$
    \item \textbf{Critic Learning Rate:} $1 \times 10^{-5}$
    \item \textbf{Number of Trajectories per Rollout:} 8
    \item \textbf{Maximum Turns per Trajectory:} 1
\end{itemize}

As detailed in Section 5.1 of the main paper, we investigated three RL task configurations:
\begin{enumerate}[leftmargin=12pt, topsep=-1pt, itemsep=0pt]
    \item \textbf{RL-FFR (from scratch):} The Qwen2.5-VL-3B-Instruct model was trained to generate free-form reasoning chains without prior SFT for this specific task format.
    \item \textbf{RL-Aug-CGMap-FFR-Out (from scratch):} The model was trained to jointly produce augmented cognitive maps and associated free-form reasoning, also starting from the base pre-trained VLM.
    \item \textbf{RL-Aug-CGMap-FFR-Out (from SFT):} For this configuration, the RL training was initialized using the weights from the strongest performing SFT checkpoint, specifically the Aug-CGMap-FFR-Out SFT model.
\end{enumerate}

The reward function was designed to be sparse yet directly indicative of desired behaviors. A reward of +1 was assigned if the model's output was structurally valid (e.g., the generated cognitive map adhered to the predefined schema). A more significant reward of +5 was given if the model provided the correct final answer to the spatial reasoning question. This dual-component reward was intended to steer the learning process towards generating both well-formed intermediate spatial representations and achieving high end-task accuracy.

\subsection{RL Reward Design Ablation}
A comprehensive reward ablation study can provide more convincing evidence for our approach. Inspired by the Deepseek-R1 approach\cite{guo2025deepseek}, our reward design incorporates a dual-component structure: a small score assigned for valid format generation and a larger score for correct answer generation. This design encourages both structural correctness and factual accuracy in the model's output.

To investigate the efficacy of this reward structure, we conducted an ablation study with three different reward configurations, as detailed in Table\ref{tab:rl_reward}. Our baseline, represented by a frozen Visual-Language Model (VLM), achieved an accuracy of 40.57\%. Our proposed setting, which assigns +1 for a valid format and +5 for a correct answer, achieved an accuracy of 49.52\%. The "Answer Only" setting, which assigns +5 for a correct answer but 0 for format, resulted in a similar accuracy of 50.76\%. In contrast, the "Format Only" setting, which provides a score of +5 for a valid format but 0 for the answer, yielded a lower accuracy of 45.05\%.

These results confirm that the reward component for a correct answer is the most critical factor in our RL reward design. The high value assigned to the correct answer score (+5) is highly effective at improving the model's performance, a finding that is consistent with recent work on RL for Large Language Models.

\begin{table}[h]
\centering
\caption{RL Reward Ablation Study}
\label{tab:rl_reward}
\begin{tabular}{@{}lc@{}}
\toprule
\textbf{RL Reward Setting (format score, answer score)} & \textbf{Acc (\%)} \\
\midrule
Frozen VLM (baseline) & 40.57 \\
Our Setting (+1, +5) & 49.52 \\
Answer Only (+0, +5) & 50.76 \\
Format Only (+5, +0) & 45.05 \\
\bottomrule
\end{tabular}
\end{table}

\subsection{VLM Response Examples After RL for Configurations in  Section \ref{app_sec:example_configuration}}

\begin{tcolorbox}[colback=boxbackground, colframe=accentblue, 
title=Output for \texttt{\iclr{FFR}}: Input Prompt Refer to \ref{app_sec:config_ff_rsn}, boxrule=0.5mm, width=\textwidth, arc=2mm, auto outer arc=true, breakable]
\textless think\textgreater  From image 4, the white jar is positioned on a surface with a floral-patterned bed sheet. To the left of the jar, there is a table with a pink tablecloth featuring heart shapes and a Hello Kitty design. There are also cups on the table. Therefore, the object to the left of the white jar is the table with the cups.\textless /think\textgreater  \textless answer\textgreater  \textcolor{red}{\bf A. Table with cups on it}\textless /answer\textgreater  
\end{tcolorbox}

\bigskip

\begin{tcolorbox}[colback=boxbackground, colframe=accentblue, 
title=Output for \texttt{RL-Aug-CGMap-FFR-Out (from scratch)}: Input Prompt Refer to \ref{app_sec:config_ag_cgmap_ffr_out}, boxrule=0.5mm, width=\textwidth, arc=2mm, auto outer arc=true, breakable]
\textless cogmap\textgreater  
\begin{verbatim}
{
  "objects": [
    {"name": "white jar", "position": [265, 436], 
        "facing": "left"}
  ],
  "views": [
    {"name": "View/Image 1", "position": [265, 436], 
      "facing": "left"},
    {"name": "View/Image 2", "position": [265, 436], 
      "facing": "left"},
    {"name": "View/Image 3", "position": [265, 436], 
      "facing": "left"},
    {"name": "View/Image 4", "position": [265, 436], 
      "facing": "left"}
  ]
}
\end{verbatim}
\textless Your cognitive map\textgreater  
\textless /cogmap\textgreater  
\newline
\textless think\textgreater  
The white jar is located in the center of the 10x10 grid, facing to the left. The clothes rack is to the left of the white jar, as seen in the image 2 view. The bed sheet with a floral pattern is located to the right of the white jar, as seen in the image 1 view. The white headboard is not visible in any of the provided images.
\newline
Based on the provided images, the white jar is located in the center of the 10x10 grid, facing to the left. The clothes rack is to the left of the white jar, as seen in the image 2 view. The bed sheet with a floral pattern is located to the right of the white jar, as seen in the image 1 view. The white headboard is not visible in any of the provided images.\textless /think\textgreater  \textless answer\textgreater  \textcolor{red}{\bf A. Table with cups on it}.\textless /answer\textgreater  
\end{tcolorbox}

\bigskip

\begin{tcolorbox}[colback=boxbackground, colframe=accentblue, 
title=Output for \texttt{RL-Aug-CGMap-FFR-Out (from SFT)}: Input Prompt Refer to \ref{app_sec:config_ag_cgmap_ffr_out}, boxrule=0.5mm, width=\textwidth, arc=2mm, auto outer arc=true, breakable]
Based on my observation, the answer is:
\newline
\textless cogmap\textgreater  
\begin{verbatim}
```json
{
  "objects": [
    {"name": "white jar", "position": [5, 5]},
    {"name": "bed sheet with a floral pattern", 
        "position": [5, 8]},
    {"name": "white headboard", "position": [2, 5]},
    {"name": "clothes rack", "position": [5, 2]},
    {"name": "table with cups on it", "position": [8, 5]}
  ],
  "views": [
    {"name": "Image 1", "position": [5, 6], "facing": "up"},
    {"name": "Image 2", "position": [4, 5], "facing": "right"},
    {"name": "Image 3", "position": [5, 4], "facing": "down"},
    {"name": "Image 4", "position": [6, 5], "facing": "left"}
  ]
}
'''
\end{verbatim}
\textless /cogmap\textgreater  
\textless think\textgreater  
In this scene, I observe four images showing different perspectives. All images feature the white jar as the main object. In image 1, I can see white jar in front of the clothes rack. In image 2, I can see white jar in front of the table with cups on it. In image 3, I can see white jar in front of the bed sheet with a floral pattern. In image 4, I can see white jar in front of the white headboard. To identify the position change across views, I focus on the main object's angle variation. Then, I analyze the angles and relative positions of other objects on the platform to back up this observation. I understand that:  Image 1 is the initial view. Image 2 is captured after a 90-degree clockwise rotation from image 1. Image 3 is after another 90-degree clockwise rotation (180 degrees from image 1). Image 4 is after a further 90-degree clockwise rotation (270 degrees from image 1). Through analyzing these perspective changes, I can construct a complete spatial understanding: when I view table with cups on it behind white jar in the second view, it implies that in the first view, table with cups on it is on the right side of white jar. Similarly, when I see white headboard behind white jar in the fourth view, it indicates that in the first view, white headboard is on the left side of white jar. However, I am still uncertain about what lies behind me in the first view. Then, I recognize that I can examine the opposite view to find out. The opposite view of the fist view is the third view. As bed sheet with a floral pattern is observed behind white jar in the third view, it means that in the first view, bed sheet with a floral pattern is positioned behind me. This way, I can fully comprehend the spatial relationships of all objects in the entire scene. So, from the perspective of image 4: clothes rack is to the right of white jar, table with cups on it is to my behind, bed sheet with a floral pattern is to the left of white jar. So the answer is C. Bed sheet with a floral pattern.\textless /think\textgreater  
\textless answer\textgreater  \textcolor{mygreen}{\bf C. Bed sheet with a floral pattern}\textless /answer\textgreater  
\end{tcolorbox}

\end{CJK*}
\end{document}